%% file: icml2026.tex
\documentclass{article}

\usepackage{microtype}
\usepackage{graphicx}
\usepackage{booktabs} 
\usepackage{hyperref}
\usepackage{natbib}

\usepackage[preprint]{icml2026}

\usepackage{times}
\usepackage{latexsym}
\usepackage{subcaption} 

\usepackage[T1]{fontenc}
\usepackage[utf8]{inputenc}

\usepackage{inconsolata}

\input{package} 

\begin{document}

\twocolumn[
\icmltitle{Instinct vs. Reflection: Unifying Token and Verbalized Confidence in Multimodal Large Models}



\begin{icmlauthorlist}
\icmlauthor{Yunkai Dang\textsuperscript{*}}{sai}
\icmlauthor{Yifan Jiang\textsuperscript{*}}{sai}
\icmlauthor{Yizhu Jiang}{sai}
\icmlauthor{Anqi Chen}{sai}
\icmlauthor{Wenbin Li\textsuperscript{$\dagger$}}{sai}
\icmlauthor{Yang Gao}{sai}
\end{icmlauthorlist}

\icmlaffiliation{sai}{School of Artificial Intelligence, Nanjing University}

\icmlcorrespondingauthor{Wenbin Li}{liwenbin.nju@gmail.com, yunkaidang1@gmail.com}

\icmlkeywords{Machine Learning, ICML, Vision Language Models}

\vskip 0.3in
]

\printAffiliationsAndNotice{\icmlEqualContribution \quad $\dagger$ Corresponding author.}




\input{section/abstract}
\input{section/introduction}
\input{section/releated}

\input{section/method}

\input{section/experiment}

\input{section/conclusion}

\bibliographystyle{icml2026}
\bibliography{custom}

\clearpage
\appendix
\onecolumn 

\input{section/appendix}

\end{document}

%% file: package.tex
\usepackage[utf8]{inputenc}
\usepackage[T1]{fontenc}
\usepackage[table,dvipsnames]{xcolor}
\definecolor{cvprblue}{rgb}{0.21,0.49,0.74}

\usepackage[table,dvipsnames]{xcolor}
\usepackage{graphicx}
\usepackage{booktabs}
\usepackage{amsfonts}
\usepackage{nicefrac}
\usepackage{microtype}
\usepackage{tabularx}
\usepackage{multirow}
\usepackage[many]{tcolorbox}
\definecolor{tablebackground}{RGB}{225,230,235} 
\definecolor{header}{RGB}{240,240,240}
\usepackage{pifont}
\definecolor{boxbg}{RGB}{215, 220, 235}
\definecolor{greyborder}{RGB}{160, 160, 160}
\usepackage{amsmath}
\DeclareMathOperator{\softmax}{softmax}

\definecolor{headerBlue}{RGB}{255, 250, 240}
\definecolor{subHeaderBlue}{RGB}{255, 250, 240}
\definecolor{openStripe}{RGB}{255, 250, 240}
\definecolor{groupBand}{RGB}{245, 235, 250}

\definecolor{closedBand}{RGB}{255, 235, 235}

%% file: section/abstract.tex
\begin{abstract}
Multimodal Large Language Models (MLLMs) have demonstrated exceptional capabilities in various perception and reasoning tasks.
Despite this success, ensuring their reliability in practical deployment necessitates robust confidence estimation.
Prior works have predominantly focused on text-only LLMs, often relying on computationally expensive self-consistency sampling.
In this paper, we extend this to multimodal settings and conduct a comprehensive evaluation of MLLMs' response confidence estimation.
Our analysis reveals a significant instinct-reflection misalignment: the model's implicit token-level support frequently diverges from its verbal self-assessment confidence.
To address this misalignment, we propose a monotone confidence fusion framework to merge dual-channel signals and cross-channel consistency to estimate correctness. 
Subsequently, an order-preserving mean alignment step is applied to correct global bias, which improves calibration while preserving the risk–coverage trade-off for selective prediction. 
Experiments on diverse open-source and closed-source MLLMs show that our method consistently yields more reliable confidence estimates and improves both calibration and failure prediction.
Code will be available at \url{https://github.com/Yunkaidang/Instinct-vs.-Reflection}.
\end{abstract}

%% file: section/introduction.tex
\section{Introduction}\label{sec:intro}

\begin{figure}[t]
  \centering
  \includegraphics[width=\linewidth]{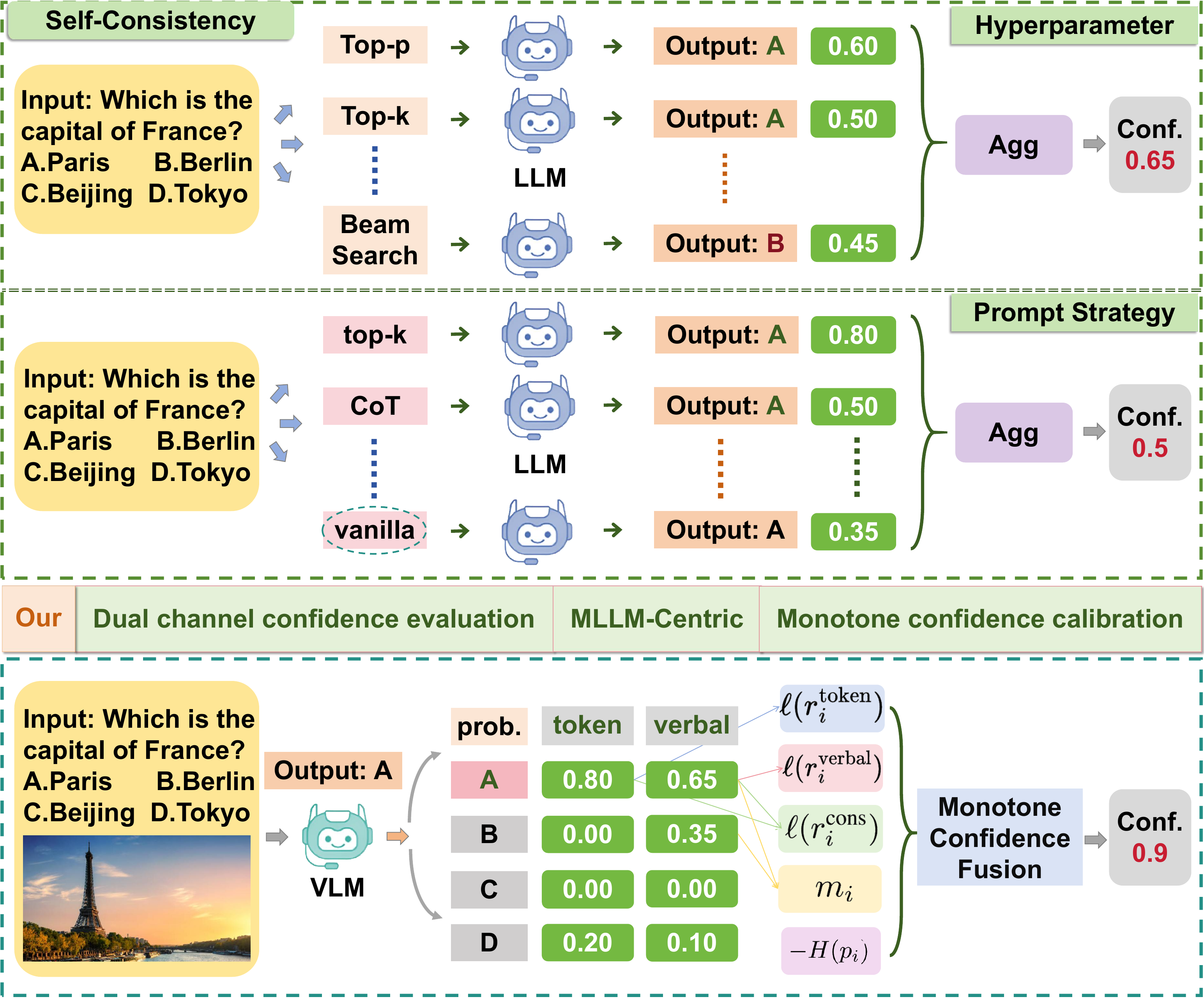}
  \caption{Comparison of confidence estimation paradigms. 
  Top/Middle: self-consistency methods rely on aggregating multiple sampled outputs via hyperparameter or prompt variations. 
  Bottom (Ours): We propose an MLLM-centric, closed-source compatible framework. 
  We extract dual-channel confidence from a single-pass inference: token confidence (instinct) and verbalized confidence (reflection). 
  These signals are integrated via monotone confidence fusion to yield a calibrated reliability score.
  }
  \label{fig:motivation}
\end{figure}

A fundamental aspect of human intelligence is the ability to correctly express confidence in our responses~\citep{koriat1993we,fleming2017self,ernst2002humans,fleming2024metacognition}.
Whether based on immediate instinct or deep reflection, humans can generally discern the boundaries of their own knowledge~\citep{pouget2016confidence,yeung2012metacognition}.
In recent years, Multimodal Large Language Models (MLLMs)~\citep{achiam2023gpt,wang2025internvl3,bai2025qwen2,abdin2024phi,dang2024explainable} have demonstrated remarkable capabilities across diverse scenarios, ranging from fine-grained visual perception to complex logical reasoning.
However, high benchmark accuracy alone does not guarantee reliability.
For models to approach human-like intelligence, they should not only provide correct answers but also faithfully quantify their certainty~\citep{zhao2024knowing,tian2023just,huang2024survey}.
Precise confidence estimation is critical for determining system reliability, enabling effective risk assessment, and mitigating errors in real-world deployment~\citep{fort2019deep,ovadia2019can,liu2020simple,farquhar2024detecting}.

However, assessing \textbf{how well a model knows what it knows} remains an open challenge. 
MLLMs commonly suffer from poor confidence calibration such as overconfident hallucinations or underconfident hesitations~\citep{xiong2023can,groot2024overconfidence}.
Prior works~\citep{taubenfeld2025confidence,kumar2024confidence,geng2024survey} have primarily explored confidence estimation for text-only Large Language Models (LLMs) from a machine learning calibration perspective.
These approaches generally fall into two categories: logits-based methods~\citep{duan2024shifting,kumar2024confidence,xie2024calibrating,kull2019beyond,shen2024thermometer}, which utilize the internal probability scores of open-source models, and
self-consistency methods~\citep{xiong2023can,prato2024large,li2024think,xie2024calibrating,li2025revisiting,du2025optimizing}
which measure the agreement among multiple reasoning paths for closed-source models.
As depicted in Figure~\ref{fig:motivation}, self-consistency methods operate by sampling multiple responses (e.g., via different prompts or parameters) and aggregating them to derive a final confidence score.
While effective for LLMs, this iterative process is cost-prohibitive due to repeated API calls~\citep{xiong2023can,dang2025exploring}. 
As a result, existing text-centric paradigms do not seamlessly transfer to MLLMs, and establishing a systematic confidence evaluation framework in the \textbf{multimodal setting} remains an underexplored problem.

Addressing these gaps, this paper focuses on evaluating and calibrating the confidence estimation of MLLMs.
We conduct a comprehensive evaluation of confidence estimation across diverse multimodal benchmarks.
Specifically, we extract confidence from two distinct cognitive stages: implicit token probabilities confidence (instinct) and explicit verbalized self-assessment confidence (reflection).
To ensure generalizability, we adopt a unified black-box protocol for both open-source and proprietary MLLMs.  
We focus on a practical scenario where only single-token probabilities are available, without access to model weights or gradients.
Our empirical analysis reveals a critical phenomenon: instinct--reflection confidence misalignment.
For instance, a model may demonstrate high verbalized confidence while simultaneously exhibiting low token confidence.
This decoupling exposes the inherent unreliability of single-source metrics, necessitating a more robust calibration approach.

Leveraging this insight, we propose a post-hoc and interpretable framework to robustly calibrate the confidence.
Specifically, we integrate implicit token confidence, explicit verbalized confidence, and cross-channel consistency into a unified reliability descriptor.
These features are mapped to a precise probability of correctness via a monotone fusion head, incorporating the essential inductive bias that stronger signals imply higher correctness. 
Subsequently, we employ an order-preserving mean alignment to refine the estimate and correct systematic bias.
This pipeline effectively minimizes calibration error while strictly maintaining the relative sample ranking, thereby preserving the model's utility for risk-sensitive selective prediction.
We evaluate both token and verbalized confidence across six multi-modal benchmarks, assessing performance via calibration (ECE) and correctness ranking metrics (AUROC, AUPRC, and AUPRC-N).
Experimental results demonstrate that our post-hoc calibration method significantly reduces ECE compared to single-source baselines.
Overall, our contributions can be summarized as follows:
\begin{itemize}
    \item We combine implicit token probabilities (instinct) and explicit verbalized self-assessments (reflection) to evaluate the MLLMs' response confidence.
    \item We propose a post-hoc and interpretable monotone confidence fusion mechanism to robustly calibrate the frequently misaligned dual-channel confidence.
    \item We provide a systematic evaluation across multiple benchmarks and MLLMs, comprehensively analyzing calibration error, failure prediction, and risk–coverage.
\end{itemize}

%% file: section/releated.tex
\section{Related Work}

\paragraph{Multimodal Large Language Models (MLLMs).}
Recent advances in LLMs have spurred multimodal large language models (MLLMs) that connect a vision encoder to an LLM via a lightweight bridge for visual-grounded instruction following and reasoning~\cite{achiam2023gpt, team2023gemini}.
Existing MLLMs broadly fall into two lines.
On one hand, frontier proprietary systems such as GPT-4o~\cite{hurst2024gpt}, Gemini-Pro~\cite{team2023gemini}, and GLM-4.1V often achieve the strongest empirical performance.
On the other hand, open-source MLLMs have rapidly expanded and provide a reproducible substrate for method development.
These models span both efficiency-oriented designs (e.g., MiniCPM-V~\cite{yao2024minicpm} and Phi-3.5-Vision~\cite{abdin2024phi}) and larger-capacity general-purpose architectures (e.g., LLaVA series~\cite{liu2023llava}, Qwen2.5-VL-7B~\cite{bai2025qwen2}, DeepSeek-VL~\cite{lu2024deepseek}, GLM-4.1V-9B, and InternVL3.5~\cite{wang2025internvl3}).
However, miscalibration remains a critical bottleneck for trustworthiness, which we address through a unified framework for confidence assessment and rectification.

\paragraph{Confidence Evaluation}
Confidence estimation~\citep{geng2024survey} has been studied most extensively in text-only LLMs. Existing methods broadly fall into two paradigms: logit-based~\citep{duan2024shifting, kumar2024confidence} and consistency-based estimation~\citep{geng2024survey, prato2024large, xiong2023can, li2024think,dang2025exploring}.
Logit-based approaches derive confidence from the model's generative distribution, e.g., aggregating token-level likelihoods~\citep{kumar2024confidence} or entropy into a sequence-level uncertainty signal~\citep{farquhar2024detecting}.
In contrast, self-consistency based approaches estimate confidence by probing the stability of model behavior across multiple samples~\citep{khanmohammadi2025calibrating}, prompts~\citep{wang2024multi}, or model variants~\citep{xiong2023can}.
Compared to the rapidly growing LLM literature, confidence evaluation in MLLMs is less standardized~\citep{tu2024empirical}.
Recent work~\cite{xuan2025seeing} analyzes the verbalized confidence in MLLMs, but limits systematic analysis across model families and downstream scenarios.
Overall, existing studies either focus on a single confidence channel (logits, sampling consistency, or verbalized confidence) or evaluate a limited slice of the MLLM settings.

\paragraph{Confidence Calibration}
Previous studies~\citep{huang2024uncertainty, shen2024thermometer, geng2024survey} have observed that LLMs often express overconfidence, which motivates post-hoc calibration to align predicted confidence with empirical accuracy.
Classic post-hoc calibration methods~\citep{naeini2015obtaining, kull2017beta, kull2019beyond}, notably Platt-style scaling~\citep{platt1999probabilistic, zadrozny2002transforming} and temperature scaling~\citep{guo2017calibration, xie2024calibrating}, have established practical tools and metrics (e.g., Brier score and ECE) for reliable probabilities in discriminative models. 
While studies~\citep{dang2025exploring, xuan2025seeing, groot2024overconfidence} have noted that MLLMs often exhibit overconfidence, they lack systematic evaluation and calibration.
Although~\cite{zhao2025object} recently attempted to calibrate verbalized confidence through semantic perturbations, their approach is limited by multiple sampling passes and extra information.
Different from them, we propose a lightweight post-hoc calibration framework that extracts dual-channel confidence, enabling unified calibration and evaluation.









%% file: section/method.tex
\newcommand{\clip}{\operatorname{clip}}
\newcommand{\softplus}{\operatorname{softplus}}


\section{Method}\label{sec:method}

\begin{figure*}[t]
    \centering
    \includegraphics[width=0.95\linewidth]{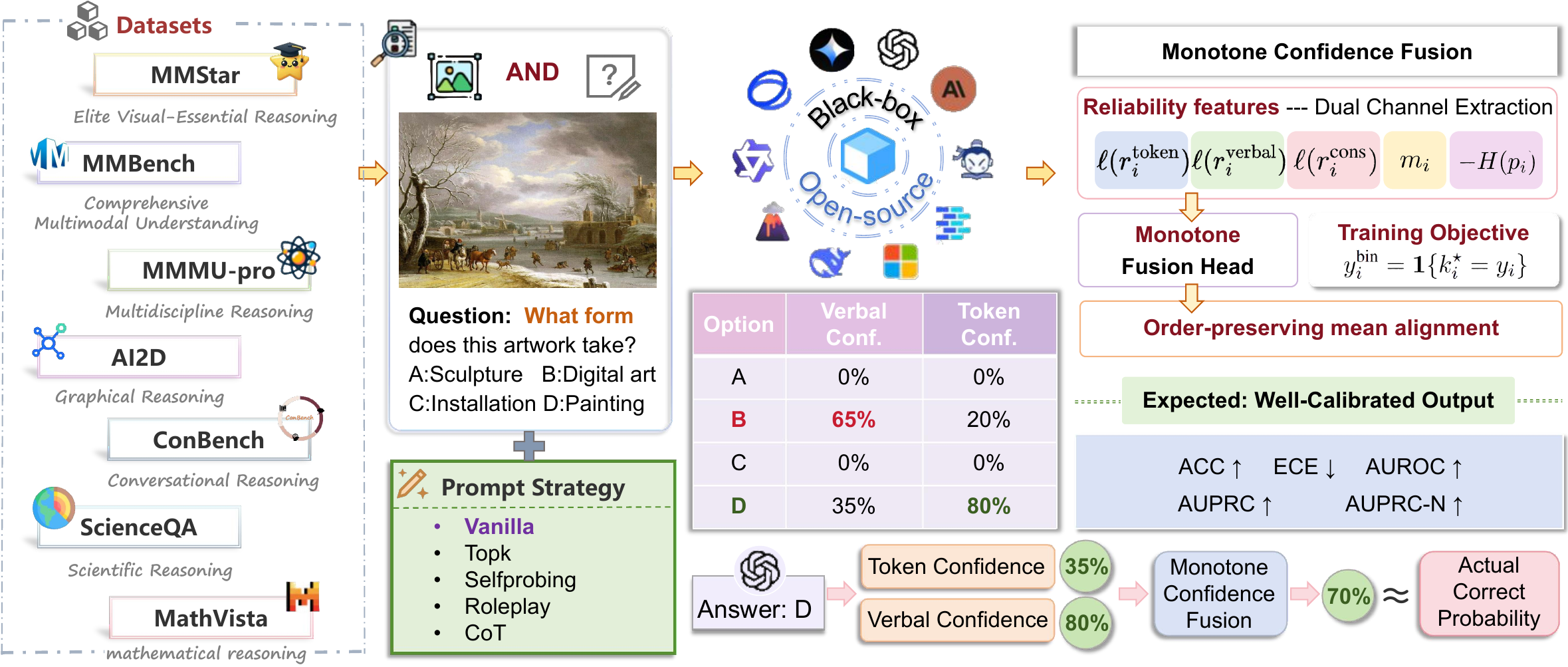}
    \caption{\textbf{Overview of our method.} We evaluate open-source and proprietary MLLMs across diverse benchmarks, employing varying prompt strategies to extract dual-channel uncertainty signals: \textit{instinct} (implicit token-based probability confidence) and \textit{reflection} (explicit verbal self-assessment confidence). These complementary cues are integrated via a monotone head followed by order-preserving mean alignment to correct distribution shifts, yielding well-calibrated reliability estimates.}
    \label{fig:pipeline}
\end{figure*}


\noindent\textbf{Task formulation.}
We evaluate several MLLMs on $K$-way multiple-choice VQA tasks.
The dataset is denoted as $\mathcal{D}=\{(x_i,q_i,\mathcal{A}_i,y_i)\}_{i=1}^{n}$, where $x_i$ represents the image,
$q_i$ denotes the question, $\mathcal{A}_i=\{1,\dots,K\}$ indexes the option set, and $y_i\in\mathcal{A}_i$ is the ground-truth index.
To enable reproducible option-level scoring, we utilize the prompting strategy to enforce a deterministic answer format. 
The MLLM generates a single option label (e.g., \texttt{``1''}$\dots$\texttt{``K''} or \texttt{``A''}$\dots$\texttt{``D''}), rather than long-form text, along with the respective confidence scores as the final output.
Our goal is to estimate, post-hoc, the calibrated confidence probability that the MLLM's top-1 prediction is correct.

\paragraph{Dual-channel confidence extraction (instinct vs.\ reflection).} 
We extract two complementary signals from the MLLM: implicit token confidence (instinct), reflecting aleatoric uncertainty in the probability distribution, and explicit verbalized assessment (reflection), capturing epistemic uncertainty via self-deliberation. 
Crucially, these signals frequently diverge—such as when high verbal certainty contradicts low token probability. 
We leverage this informative misalignment to enhance reliability estimation in multimodal settings.

\noindent\textbf{Token confidence (implicit).}
Let $z_{ik}$ denote the MLLM's log-probability of the option-label token corresponding to choice $k$ for the $i$-th sample at the answer position,
conditioned on $(x_i,q_i)$ under the fixed answer-format instruction. 
We compute the option-token confidence vector as follows:
\begin{equation}\label{eq:prob-vector}
p_i=(p_{i1},\dots,p_{iK})\in\Delta^{K-1},
\end{equation}
where each component $p_{ik}$ is the normalized probability assigned to the $k$-th candidate:
\begin{equation}\label{eq:prob-softmax}
p_{ik} = \frac{\exp(z_{ik})}{\sum_{k'=1}^{K} \exp(z_{ik'})}.
\end{equation}
This token signal captures the model's local token-level support distribution over the discrete set of $K$ candidates.

\noindent\textbf{Verbalized confidence (explicit).}
In parallel, we prompt the MLLM to explicitly report its estimated probability for each option under a structured template that is easy to parse (Appendix~\ref{app:prompt-parse}). We then parse the resulting text into a verbalized confidence vector consistent with the option set:
\begin{equation}\label{eq:self-vector}
s_i=(s_{i1},\dots,s_{iK})\in[0,1]^K,
\end{equation}
where $s_{ik}$ is the model's stated probability (e.g., $80\%$) for option $k$.
Unlike $p_i\in\Delta^{K-1}$, $s_i$ lies in the hypercube and may not sum to $1$,
allowing the model to express ambiguity (multiple plausible options) or ignorance (all options low).

\noindent\textbf{Cross-channel confidence consistency.}
A trustworthy model should exhibit synchronization between instinct (token confidence) and reflection (verbalized confidence). 
This cross-channel coherence signifies an internal consensus across different cognitive inference stages.
Specifically, for each sample $i$ and candidate $k$, we map their discrepancy into a similarity score using an RBF kernel function:
\begin{equation}\label{eq:consistency-coeff}
\kappa_{ik}=\exp\!\left(-\frac{|p_{ik}-s_{ik}|^{\gamma}}{\tau}\right),
\end{equation}
where $\kappa_{ik}\in(0,1]$, $\gamma>0$ controls the shape (we use $\gamma=2$), and $\tau>0$ is a bandwidth parameter.
This metric assigns high agreement to both calibrated uncertainty (where both channels are low) and calibrated confidence (where both channels are high),
while penalizing severe misalignment that is often indicative of hallucinated certainty.
The non-linear curvature of the RBF kernel amplifies subtle discrepancies, providing better discriminative power than linear metrics.


For the $i$-th instance, the model first identifies the predicted option at inference time by maximizing the token probability over the candidate set $\mathcal{A}_i$:
\begin{equation}
k_i^\star = \arg\max_{k\in\mathcal{A}_i} p_{ik}.
\end{equation}
Based on this prediction, we derive three scalar signals to quantify reliability. Specifically, the token confidence captures the instinctive support for the top-1 option:
\begin{equation}
r^{\text{token}}_i = p_{i,k_i^\star}.
\end{equation}
Complementing this internal metric, we explicitly evaluate the verbal confidence, which records the model's reflective self-assessment during the generation process:
\begin{equation}
r^{\text{verbal}}_i = s_{i,k_i^\star}.
\end{equation}
Finally, to measure robustness against stochastic perturbations, we compute the consistency score, quantifying the overall agreement across multiple reasoning paths:
\begin{equation}
r^{\text{cons}}_i = \kappa_{i,k_i^\star}.
\end{equation}

\noindent\textbf{Design rationale for MLLMs.}
Our fusion design follows four principles:
(i) Monotone priors. Each reliability cue is constructed so that larger values imply higher correctness likelihood, motivating nonnegative fusion weights.
(ii) Comparable geometry. Log-odds place heterogeneous confidence scales onto a shared linear domain, while margin and entropy capture local token sharpness.
(iii) Shift-robust bias correction. A post-hoc bias adjustment matches mean confidence to empirical accuracy without changing ranking, enabling stable coverage control.
(iv) Generalizes classical calibration. The linear head subsumes Platt/temperature scaling as a special case and extends it to multi-signal fusion with more interpretable feature attribution.

\subsection{Monotone Confidence Fusion}\label{sec:merge}

\noindent\textbf{Reliability features.}
To provide a comprehensive characterization of the model's confidence landscape, we formulate a compact reliability descriptor $\phi_i \in \mathbb{R}^d$ ($d=5$) by integrating evidence from both channels:
\begin{equation}\label{eq:feat}
\phi_i=
\Big[
\ell(r_i^{\text{token}}),\;
\ell(r_i^{\text{verbal}}),\;
\ell(r_i^{\text{cons}}),\;
m_i,\;
-H(p_i)
\Big],
\end{equation}
where $\ell(\cdot)$ performs a log-odds transformation, mapping $(0,1)$ to $\mathbb{R}$ with numerical safeguards:
\begin{equation}\label{eq:logitclip}
\ell(z) = \log \frac{\text{clip}(z, \varepsilon, 1-\varepsilon)}{1 - \text{clip}(z, \varepsilon, 1-\varepsilon)}, \quad \varepsilon > 0.
\end{equation}
This mapping mitigates the saturation effects near the probability boundaries, thereby increasing the fusion head's sensitivity to fine-grained confidence variations.

Beyond confidence and consensus, we incorporate top-2 margin and entropy as supplementary uncertainty metrics:
\begin{equation}\label{eq:gap}
m_i=p_{i,(1)}-p_{i,(2)},
\end{equation}
where $p_{i,(j)}$ denotes the $j$-th largest probability in the sorted distribution.
The top-2 margin $m_i$ quantifies local decision sharpness by measuring the separation between the leading candidates, where a larger gap indicates reduced ambiguity.
\begin{equation}\label{eq:entropy}
H(p_i)=-\sum_{k=1}^{K} p_{ik}\log p_{ik}.
\end{equation}
Shannon entropy $H(p_i)$ characterizes global belief dispersion across all categories.
In our descriptor $\phi_i$, we utilize its negative form $-H(p_i)$ so that a more concentrated distribution consistently corresponds to a higher reliability value.


\noindent\textbf{Monotone fusion head.}
Each component in Eq.~\eqref{eq:feat} is tailored to capture a distinct dimension of uncertainty. 
Notably, these features are monotonically aligned by construction.
Based on this design, we estimate the probability that the MLLM's top-1 prediction is correct by fusing $\phi_i$ with a coordinatewise monotone logistic head:
\begin{equation}\label{eq:head}
\hat q_i=\sigma\!\big(f_\theta(\phi_i)\big),
\qquad
f_\theta(\phi_i)=b+\sum_{j=1}^{d} w_j\phi_{ij},
\end{equation}
where $\sigma(t)=1/(1+e^{-t})$ and $w_j\ge 0$.
The nonnegativity constraint encodes the inductive bias that increasing any reliability cue should not reduce predicted correctness (Appendix~\ref{app:proofs}).
We parameterize $w_j$ without constrained optimization via a simple reparameterization:
\begin{equation}\label{eq:softplus}
w_j=\softplus(\tilde w_j),
\end{equation}
where $\text{softplus}(x) = \ln(1 + \exp(x))$ ensures the positivity of $w_j$ by mapping real-valued $\tilde w_j$ to the range $(0, +\infty)$.


\noindent\textbf{Training objective.}
Define the correctness indicator as:
\begin{equation}\label{eq:binary-target}
y_i^{\text{bin}}=\mathbf{1}\{k_i^\star=y_i\}.
\end{equation}
We fit the model parameters $\theta=(b,\{\tilde w_j\})$ by minimizing the negative log-likelihood objective function:
\begin{equation}\label{eq:nll}
\mathcal{L}_{\mathrm{NLL}}(\theta) 
= -\frac{1}{n} \sum_{i=1}^{n} 
\left[ 
    y_i^{\text{bin}} \log(\hat{q}_i) + (1 - y_i^{\text{bin}}) \log(1 - \hat{q}_i) 
\right].
\end{equation}
When $d{=}1$ and $\phi_i=\ell(r_i^{\text{token}})$, \eqref{eq:head} reduces to Platt scaling; thus, our head generalizes standard post-hoc calibration.
With the backbone MLLM frozen, we optimize $\theta$ on a calibration set and evaluate performance on a disjoint test set to ensure out-of-sample generalization.


\noindent\textbf{Order-preserving mean alignment.}
We apply a post-hoc bias shift
\begin{equation}\label{eq:bshift}
b \leftarrow b+\delta,
\end{equation}
with $\delta$ chosen to satisfy~\eqref{eq:meanalign}.
This step is intentionally order-preserving (Appendix~\ref{app:meanalign-theory}) and thus does not change any ranking-based evaluation
(e.g., risk--coverage curves), while correcting systematic base-rate misalignment.

\begin{equation}\label{eq:meanalign}
\frac{1}{n}\sum_{i=1}^{n}\sigma\!\big(f_\theta(\phi_i)+\delta\big)
=\frac{1}{n}\sum_{i=1}^{n} y_i^{\mathrm{bin}}
=\mathrm{ACC}.
\end{equation}
In practice we clip the empirical $\mathrm{ACC}$ to $[\varepsilon,1-\varepsilon]$ to avoid degenerate edge cases and solve for the optimal $\delta$ (Appendix~\ref{app:proofs}) by performing an efficient one-dimensional bisection search
procedure (Appendix~\ref{app:meanalign}).




%% file: section/experiment.tex
\subsection{Experiment setup} \label{sec:setup}

\input{table/merge_1.28}

\paragraph{Datasets}
To comprehensively evaluate the confidence of existing models across multimodal benchmarks, we employ six datasets spanning diverse domains and difficulty levels. 
Since most multimodal benchmarks are formulated as multiple-choice questions, we focus on datasets in the multiple-choice setting.
These include general-purpose evaluation benchmarks (MMBench~\citep{liu2024mmbench}, MMStar~\citep{chen2024we}, and ConBench~\citep{zhang2024unveiling}), as well as datasets focused on scientific reasoning (ScienceQA~\citep{lu2022learn}), diagrammatic reasoning (AI2D~\citep{kembhavi2016diagram}), and massive multi-discipline professional reasoning (MMMU-Pro~\citep{yue2024mmmu}). 
This selection balances general capability, generalization, and depth of domain-specific reasoning.
We then extract per-question confidence scores for each multiple-choice VQA instance (i.e., the question and candidate options) from these datasets to assess the confidence of model responses.
If a response does not follow the template or omits the confidence value, we treat it as missing and assign a default confidence score of 50 on a 0--100 scale.

\paragraph{Models}
To obtain a representative view of the confidence behavior of contemporary MLLMs, we evaluate a diverse set of systems. 
Specifically, we consider five open-source models—MiniCPM-V-2.6~\citep{yao2024minicpm}, Qwen2.5-VL-7B~\citep{bai2025qwen2}, InternVL2.5-4B~\citep{chen2024expanding}, InternVL3.5-8B~\citep{wang2025internvl3}, and Phi-3.5-Vision—as well as two closed-source models, GPT-4o and GPT-4o-mini~\citep{hurst2024gpt}.
While we utilize token-level probabilities assigned to the predicted option tokens, our approach does not require intrusive open-source access to model gradients. 
Since these signals are readily accessible via standard APIs, our approach remains broadly applicable to both open-source and closed-source models.
Additionally, we incorporate models such as Claude-3.7-Sonnet, GPT-5, and Gemini-2.0-Flash in several extended analyses.

\paragraph{Evaluation Metrics.}
We evaluate confidence along two complementary axes: calibration and failure prediction.
For calibration, we report ECE (Expected Calibration Error)~\citep{guo2017calibration}, which measures, over discretized confidence bins, the discrepancy between a model's verbalized confidence and its empirical accuracy.
Lower ECE indicates better calibration, meaning the verbalized confidence more closely matches observed correctness.
We also report ACC (top-1 accuracy) as a basic measure of predictive quality.
For failure prediction, we use AUROC, a threshold-free metric that quantifies how well confidence separates correct from incorrect outcomes.
Higher AUROC indicates stronger discriminative ability, implying that incorrect predictions tend to receive systematically lower confidence than correct ones.
We additionally report AUPRC-N, which rescales the Precision-Recall area relative to task-specific random baselines.
Higher AUPRC-N indicates superior confidence ranking, which effectively reflects the model's capacity to distinguish between successful and failed predictions across all operating points.
Together, ECE, ACC, AUROC, and AUPRC-N provide a compact, threshold-agnostic view of whether confidence estimates are well calibrated and operationally useful for risk-sensitive decisions across tasks.
All metrics are reported in the same $\times 10^{2}$ scale.

\input{table/calibrate}

\paragraph{Results on Token Confidence.}
Task complexity is the primary driver for both $\text{ECE}$ and $\text{AUROC}$ in token confidence: calibration quality sharply degrades from ScienceQA (typically $<5.0$) to complex logical tasks, where GPT-4o reaches an $\text{ECE}$ of $54.85$ on AI2D. While models maintain exceptional discriminative power on foundational tasks ($\text{AUROC} > 90.0$), this ability collapses alongside $\text{AUPRC-N}$ as complexity increases. Open-source models demonstrate superior honesty in knowledge-sufficient domains—notably InternVL2.5-4B's $\text{ECE}$ of $0.88$ on ScienceQA—yet suffer from cognitive inversion under extreme pressure, with the InternVL series yielding negative $\text{AUPRC-N}$ values (e.g., $-22.66$ on MMMU-Pro). Conversely, closed-source models exhibit greater resilience: despite higher absolute errors, their robust ranking ability (GPT-4o's $\text{AUPRC-N}$ of $43.97$ on MMMU-Pro) indicates more reliable self-awareness in complex scenarios.

\paragraph{Results on Verbalized Confidence.}
Results in Table~\ref{tab:mm-6sets-token-vs-verbal-clean} show significant performance disparities in verbalized reliability across model categories. Closed-source models, particularly GPT-4o-mini, demonstrate superior cross-task stability, whereas open-source models like Qwen2.5-VL-7B exhibit severe fluctuations, with $\text{ECE}$ reaching $47.68$ on MMMU-Pro. Under extreme logical pressure, the discriminative ability $\text{AUROC}$ of models collapses. For instance, Qwen2.5-VL-7B yields an $\text{AUROC}$ of only $35.91$ on ConBench—well below the random threshold of $50.0$—indicating a total loss of self-assessment capacity. Notably, the open-source landscape is highly heterogeneous: Phi-3.5-Vision exhibits significantly higher miscalibration on MMStar ($\text{ECE}$ $46.55$) and AI2D ($33.83$) compared to InternVL2.5-4B ($26.97$ and $10.96$, respectively). Furthermore, the $\text{AUPRC-N}$ metric exposes a deeper unreliability, with certain models yielding negative values (e.g., Qwen2.5-VL-7B at $-12.64$ on ConBench), suggesting that their verbalized confidence is inversely correlated with actual correctness on complex reasoning tasks.

\paragraph{Comparative Analysis: Token vs. Verbalized Confidence under Task Pressure.}

Table~\ref{tab:mm-6sets-token-vs-verbal-clean} reveals that the efficacy of both confidence channels is strictly conditioned on task difficulty and model architecture, 
indicating a pronounced propensity for miscalibration under logical pressure. 
For foundational tasks like ScienceQA, token confidence achieves superior calibration. 
For instance, InternVL2.5-4B yields a token $\text{ECE}$ of $0.88$ vs. $7.64$ for verbalization $\text{ECE}$, suggesting that internal probabilities more faithfully reflect uncertainty when domain knowledge is sufficient. 
Conversely, complex reasoning tasks elicit significant model-specific heterogeneity. 
GPT-4o-mini maintains stable verbalized calibration ($\text{ECE} < 20.0$) despite volatile token $\text{ECE}$ scores ($43.66$ on MMStar and $49.43$ on AI2D). 
In contrast, MiniCPM-V-2.6 favors token-based assessment on MMMU-Pro, yielding a token $\text{ECE}$ of $26.48$ vs. a verbalized $\text{ECE}$ of $37.19$. 
This divergence underscores that no single confidence paradigm remains robust across all difficulty gradients.
Such complementarity inspires our adaptive fusion framework to enhance uncertainty estimation by synergizing internal and external channels.

\begin{figure}[ht]
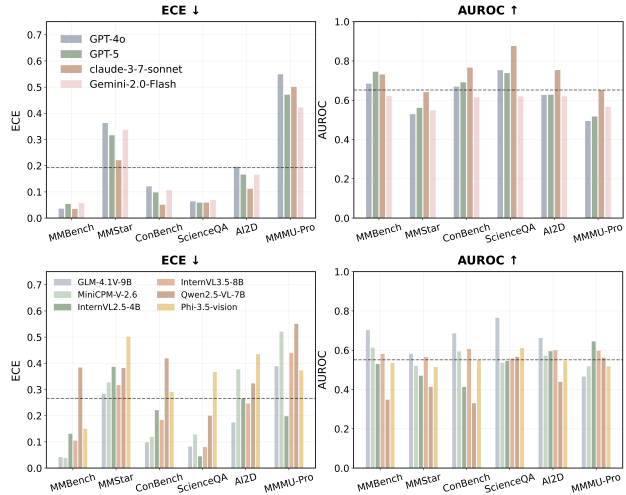

    \centering

    \begin{subfigure}{\linewidth}
        \centering
        \includegraphics[width=\linewidth]{image12.13/ece+auroc_blackbox.png}
    \end{subfigure}

    \begin{subfigure}{\linewidth}
        \centering
        \includegraphics[width=\linewidth]{image12.13/ece+auroc_whitebox.png}
    \end{subfigure}

    \caption{Comparison of ECE and AUROC metrics using verbalized confidence.
    The figure contrasts the performance of closed-source (top) and open-source (bottom) models across six datasets.
    }
    \label{fig:ece_auroc_comparison}
\end{figure}

\paragraph{Results on Calibrated Confidence.}
Results in Table~\ref{tab:mm-6sets-prob-delta-fused-final} indicate that our fused confidence primarily functions as a robust post-hoc recalibration tool, yielding substantial $\text{ECE}$ reductions across most benchmarks. 
This gain is particularly significant in high-complexity reasoning tasks like MMMU-Pro and AI2D. 
For instance, InternVL3.5-8B's $\text{ECE}$ on MMMU-Pro decreases by $39.50$ (dropping to $2.41$). 
This underscores that our proposed multi-channel fusion framework effectively mitigates single-channel miscalibration by leveraging complementary information. 
Conversely, foundational tasks like ScienceQA exhibit a marginal cost, where noise from the secondary channel causes slight $\text{ECE}$ inflation (e.g.,$+2.10$ for GPT-4o) for models with near-perfect single-channel baselines. 
Furthermore, improvements in discriminative power ($\text{AUROC}$ and $\text{AUPRC-N}$) are inconsistent and highly model-sensitive. 
While GPT-4o consistently benefits (e.g., an $8.99$ increase in $\text{AUPRC-N}$ on MMMU-Pro), other models like MiniCPM-V-2.6 show degraded ranking performance in complex scenarios.

\paragraph{Comparative Analysis of ECE and AUROC.}
Figure~\ref{fig:ece_auroc_comparison} illustrates the calibration and discriminative performance, contrasting closed-source (top) and open-source (bottom) models.
A primary observation is the performance gap between closed-source and open-source models. 
This gap is reflected not only in a difference in mean performance—closed-source models exhibit lower average ECE and higher AUROC overall—but also in their stronger consistency across model families. 
Among the evaluated closed-source models, Claude-3.7-Sonnet stands out as the most robust performer. 
It consistently achieves the highest AUROC scores and relatively lower ECE across nearly all benchmarks. 
For instance, on foundational tasks like ScienceQA, its AUROC bars reach peaks of approximately $0.88$, while its ECE remains low, at approximately $0.06$. 

\begin{figure}[t]
    \centering
    
    \begin{subfigure}{0.47\linewidth}
        \includegraphics[width=\linewidth]{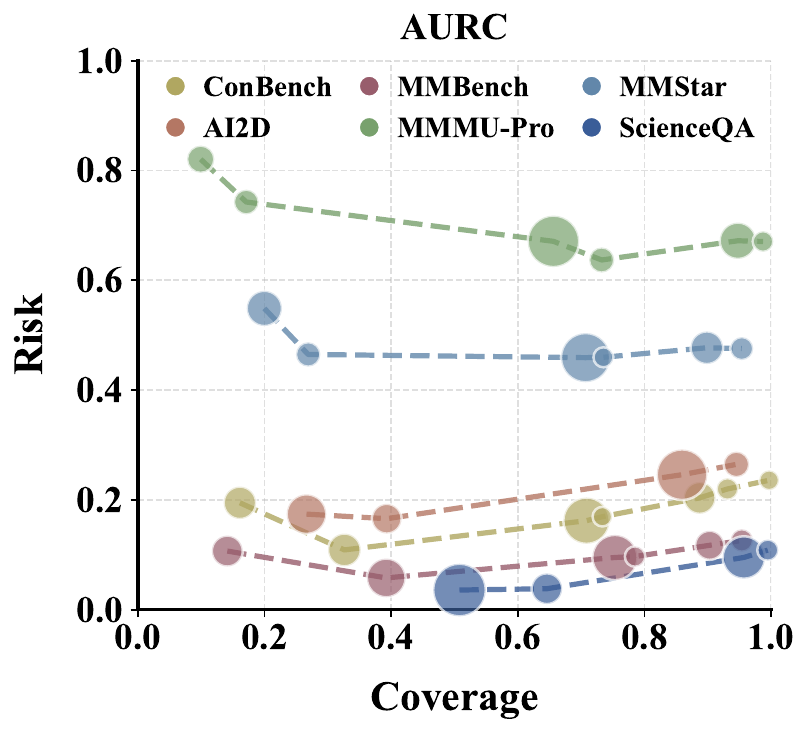}
    \end{subfigure}
    \hfill
    \begin{subfigure}{0.52\linewidth}
        \includegraphics[width=\linewidth]{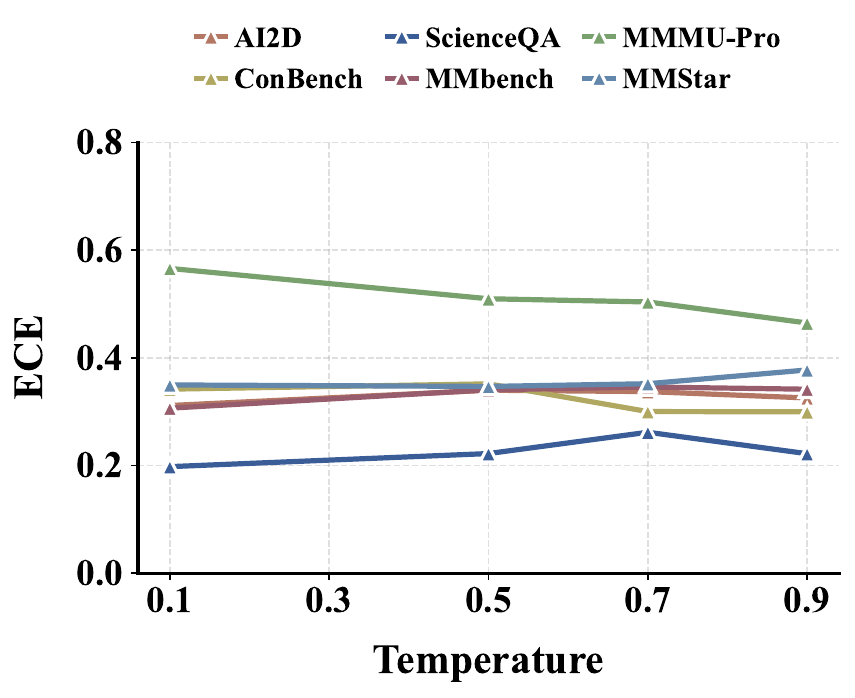}
    \end{subfigure}
    \caption{(Left) Risk-coverage curves for GPT-4o. Node radius encodes marginal sample volume per confidence bin; only nodes exceeding a minimum threshold are plotted to mitigate small-sample artifacts. (Right) ECE vs temperature.}
    \label{fig:two_parallel_acc}
\end{figure}

\paragraph{Results of risk--coverage analysis.}
Figure~\ref{fig:two_parallel_acc} (left) reports risk--coverage (RC) curves and the corresponding area under the risk--coverage curve (AURC) for GPT-4o~\citep{hurst2024gpt}.
For each benchmark in the single-image setting, we sort the \(n\) test instances by the verbal confidence \(c_i\) in descending order.
At coverage \(\alpha = k/n\), we accept the top-\(k\) instances and compute accuracy
\(\hat a(\alpha)=\frac{1}{k}\sum_{i=1}^{k}\mathbf{1}[\hat y_i=y_i]\),
with risk defined as \(r(\alpha)=1-\hat a(\alpha)\).
We summarize the trade-off by \(\mathrm{AURC}=\int_{0}^{1} r(\alpha)\,d\alpha\), approximated via the trapezoidal rule, where lower values indicate stronger selective reliability.
RC curves differ across datasets.
ScienceQA achieves the lowest risk across most coverages. 
Despite a slight increase, it stays below \(0.1\) threshold even at full coverage.
MMMU-Pro exhibits substantially higher risk and a non-monotonic trend, decreasing from about \(0.81\) at low coverage to roughly \(0.62\) at moderate coverage before rising again toward \(0.67\) as \(\alpha \to 1\).
Overall, GPT-4o confidence supports selective prediction on ScienceQA, remains moderately informative on MMBench and AI2D, and is least reliable on MMStar and MMMU-Pro.

\paragraph{Effect of Temperature on Calibration.}
Motivated by recent perspectives on answer aggregation and confidence-aware self-consistency~\cite{li2025revisiting,taubenfeld2025confidence} and multi-sample temperature optimization~\cite{du2025optimizing}, we analyze how the temperature $T$ affects confidence calibration in multimodal generation, where epistemic and perceptual uncertainty are typically higher than in unimodal classification.
Concretely, we sweep the sampling temperature $T\in\{0.1,0.5,0.7,0.9\}$ while keeping the remaining decoding configuration fixed, and compute the ECE using verbalized confidence.
Figure~\ref{fig:two_parallel_acc} (right) shows that temperature acts as a simple calibration mechanism: on MMMU-Pro, increasing $T$ consistently reduces ECE (from $\approx 0.56$ at $T=0.1$ to $\approx 0.46$ at $T=0.9$).
On the remaining datasets, ECE remains relatively stable, with some mild dataset-specific trends: for instance, ConBench improves at higher temperatures, and MMStar slightly degrades at the highest temperature. 
We further verify the results for several other models in the appendix, including GLM-4.1V-9B, InternVL3.5-8B, Phi-3.5-Vision, and MiniCPM-2.6. 
Overall, while a higher temperature helps mitigate over-confidence in challenging multimodal settings, it does not substantially affect calibration across most benchmarks.

\begin{figure}[t]
    \centering
    \includegraphics[width=1\linewidth]{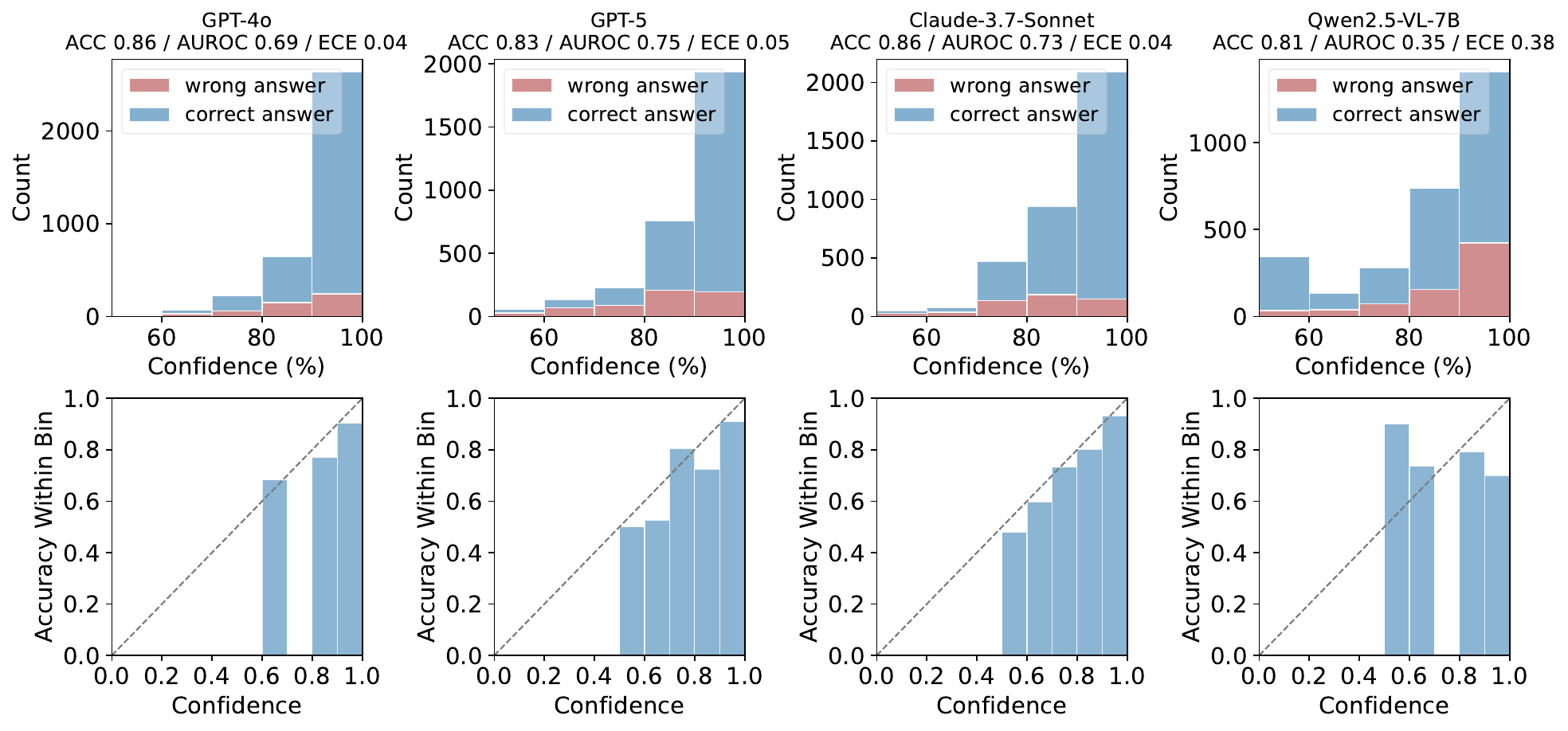}
    \caption{Verbalized confidence on MMBench for GPT-4o, GPT-5, Claude-3.7-Sonnet, and Qwen2.5-VL-7B. 
    Top: confidence histograms for correct vs.\ wrong predictions. 
    Bottom: reliability diagrams (dashed line: perfect calibration).}
    \label{fig:mmbench_calibration}
\end{figure}

\paragraph{Calibration and Reliability Analysis.}
Figure~\ref{fig:mmbench_calibration} illustrates the verbalized confidence calibration on MMBench by analyzing confidence histograms (top) and reliability diagrams (bottom). 
The closed-source models demonstrate higher reliability: correct predictions (blue) are concentrated in high-confidence regions ($>0.8$), while incorrect predictions (red) are properly assigned lower confidence. 
This alignment results in reliability curves that closely follow the ideal diagonal, achieving minimal ECE ($\approx 0.04\text{--}0.05$) and comparatively higher AUROC. 
In contrast, the open-source model exhibits severe overconfidence alongside a notable lack of predictive orderliness. 
The ECE of Qwen2.5-VL-7B ($0.38$) is nearly an order of magnitude higher than closed-source counterparts.
Its histogram reveals that a large portion of wrong answers are incorrectly assigned the highest confidence, implying the model is often confidently wrong. 
Results suggest that while closed-source models are better-calibrated, current open-source models still struggle to accurately estimate predictive uncertainty.

%% file: table/merge_1.28.tex
\begin{table*}[t]
\centering
\scriptsize
\caption{Calibration and confidence-quality metrics (without ACC) of open-source and closed-source MLLMs on six benchmarks, comparing token confidence and verbalized confidence.
Higher values indicate better performance for \textbf{AUROC}$\uparrow$, \textbf{AUPRC}$\uparrow$, and \textbf{AUPRC-N}$\uparrow$, while lower values are better for \textbf{ECE}$\downarrow$.
Metrics are reported as $\times 10^{2}$.}
\label{tab:mm-6sets-token-vs-verbal-clean}

\begin{subtable}[t]{\textwidth}
\centering
\label{tab:mm-no-acc-a}

\resizebox{0.9\linewidth}{!}{%
\begin{tabular}{@{}c l *{4}{c} @{\hspace{4pt}} *{4}{c} @{\hspace{4pt}} *{4}{c}@{}}
\toprule
\rowcolor{headerBlue}
\multirow{2}{*}{\textbf{Conf.}} & \multirow{2}{*}{\textbf{Model}} &
\multicolumn{4}{c}{\textbf{MMBench}} &
\multicolumn{4}{c}{\textbf{MMStar}} &
\multicolumn{4}{c}{\textbf{ConBench}} \\
\cmidrule(lr){3-6}\cmidrule(lr){7-10}\cmidrule(lr){11-14}
\rowcolor{subHeaderBlue}
& &
\textbf{ECE}$\downarrow$ & \textbf{AUROC}$\uparrow$ & \textbf{AUPRC}$\uparrow$ & \textbf{AUPRC-N}$\uparrow$ &
\textbf{ECE}$\downarrow$ & \textbf{AUROC}$\uparrow$ & \textbf{AUPRC}$\uparrow$ & \textbf{AUPRC-N}$\uparrow$ &
\textbf{ECE}$\downarrow$ & \textbf{AUROC}$\uparrow$ & \textbf{AUPRC}$\uparrow$ & \textbf{AUPRC-N}$\uparrow$ \\
\midrule
\rowcolors{4}{tablebackground}{white}
\multirow{9}{*}{\rotatebox[origin=c]{90}{\textbf{Token Confidence}}} &
\multicolumn{13}{l}{\cellcolor{groupBand}\textbf{\strut Open-source MLLMs}} \\

& MiniCPM-V-2.6~\citep{yao2024minicpm} & 5.14 & 89.97 & 98.40 & 88.01 & 17.11 & 70.69 & 74.35 & 49.38 & 14.08 & 70.19 & 89.19 & 56.10 \\

& InternVL2.5-4B~\citep{chen2024expanding} & 5.12 & 89.22 & 97.98 & 86.24 & 18.74 & 66.88 & 77.37 & 43.90 & 12.26 & 81.84 & 95.11 & 74.24 \\

& InternVL3.5-8B~\citep{wang2025internvl3} & 5.63 & 88.86 & 98.66 & 86.92 & 20.12 & 70.43 & 84.69 & 56.25 & 15.98 & 83.33 & 94.08 & 73.11 \\

& Qwen2.5-VL-7B~\citep{bai2025qwen2} & 4.44 & 90.40 & 98.53 & 87.93 & 12.30 & 68.32 & 79.46 & 42.94 & 9.32 & 85.72 & 95.29 & 78.86 \\

& Phi-3.5-Vision~\citep{abdin2024phi} & 9.16 & 86.22 & 96.50 & 82.20 & 24.60 & 61.10 & 66.57 & 34.46 & 10.58 & 80.78 & 89.28 & 64.84 \\

& \multicolumn{13}{l}{\cellcolor{closedBand}\textbf{\strut Closed-source MLLMs}} \\

& GPT-4o~\citep{hurst2024gpt} & 7.61 & 91.33 & 98.81 & 89.51 & 28.37 & 74.01 & 77.46 & 49.50 & 12.33 & 90.52 & 98.22 & 88.99 \\

& GPT-4o-mini~\citep{hurst2024gpt} & 13.39 & 88.71 & 97.56 & 85.64 & 43.66 & 65.82 & 62.80 & 33.97 & 19.71 & 80.46 & 93.01 & 69.62 \\

\midrule
\multirow{9}{*}{\rotatebox[origin=c]{90}{\textbf{Verbal Confidence}}} &
\multicolumn{13}{l}{\cellcolor{groupBand}\textbf{\strut Open-source MLLMs}} \\

& MiniCPM-V-2.6~\citep{yao2024minicpm} & 7.45 & 74.71 & 93.62 & 52.27 & 26.87 & 54.68 & 53.22 & 7.69 & 10.98 & 62.96 & 82.27 & 28.00 \\

& InternVL2.5-4B~\citep{chen2024expanding} & 10.99 & 56.67 & 86.50 & 7.94 & 26.97 & 52.06 & 62.94 & 8.11 & 13.87 & 61.26 & 87.27 & 33.02 \\

& InternVL3.5-8B~\citep{wang2025internvl3} & 9.35 & 72.93 & 95.18 & 53.12 & 20.44 & 64.45 & 75.22 & 29.21 & 13.12 & 65.12 & 85.89 & 35.86 \\

& Qwen2.5-VL-7B~\citep{bai2025qwen2} & 37.64 & 45.92 & 87.54 & -2.19 & 29.32 & 43.90 & 59.58 & -12.27 & 38.65 & 35.91 & 74.90 & -12.64 \\

& Phi-3.5-Vision~\citep{abdin2024phi} & 12.21 & 63.54 & 86.35 & 30.48 & 46.55 & 44.98 & 48.52 & -0.95 & 20.33 & 63.14 & 72.65 & 10.32 \\

& \multicolumn{13}{l}{\cellcolor{closedBand}\textbf{\strut Closed-source MLLMs}} \\

& GPT-4o~\citep{hurst2024gpt} & 6.04 & 69.67 & 92.68 & 35.63 & 21.68 & 58.98 & 58.92 & 7.95 & 8.41 & 77.71 & 94.35 & 64.98 \\

& GPT-4o-mini~\citep{hurst2024gpt} & 8.08 & 77.61 & 92.66 & 56.78 & 19.06 & 61.85 & 54.55 & 19.32 & 14.10 & 64.82 & 83.41 & 27.87 \\

\bottomrule
\end{tabular}%
}
\end{subtable}

\vspace{0.3em}

\begin{subtable}[t]{\textwidth}
\centering
\label{tab:mm-no-acc-b}

\resizebox{0.9\linewidth}{!}{%
\begin{tabular}{@{}c l *{4}{c} @{\hspace{4pt}} *{4}{c} @{\hspace{4pt}} *{4}{c}@{}}
\toprule
\rowcolor{headerBlue}
\multirow{2}{*}{\textbf{Conf.}} & \multirow{2}{*}{\textbf{Model}} &
\multicolumn{4}{c}{\textbf{ScienceQA}} &
\multicolumn{4}{c}{\textbf{AI2D}} &
\multicolumn{4}{c}{\textbf{MMMU-Pro}} \\
\cmidrule(lr){3-6}\cmidrule(lr){7-10}\cmidrule(lr){11-14}
\rowcolor{subHeaderBlue}
& &
\textbf{ECE}$\downarrow$ & \textbf{AUROC}$\uparrow$ & \textbf{AUPRC}$\uparrow$ & \textbf{AUPRC-N}$\uparrow$ &
\textbf{ECE}$\downarrow$ & \textbf{AUROC}$\uparrow$ & \textbf{AUPRC}$\uparrow$ & \textbf{AUPRC-N}$\uparrow$ &
\textbf{ECE}$\downarrow$ & \textbf{AUROC}$\uparrow$ & \textbf{AUPRC}$\uparrow$ & \textbf{AUPRC-N}$\uparrow$ \\
\midrule
\rowcolors{4}{tablebackground}{white}
\multirow{9}{*}{\rotatebox[origin=c]{90}{\textbf{Token Confidence}}} &
\multicolumn{13}{l}{\cellcolor{groupBand}\textbf{\strut Open-source MLLMs}} \\

& MiniCPM-V-2.6~\citep{yao2024minicpm} & 1.88 & 98.69 & 99.99 & 98.69 & 4.21 & 85.61 & 95.87 & 80.38 & 26.48 & 44.23 & 33.89 & -5.05 \\

& InternVL2.5-4B~\citep{chen2024expanding} & 0.88 & 92.81 & 99.95 & 92.57 & 8.13 & 86.12 & 95.98 & 81.31 & 39.49 & 31.64 & 43.00 & -22.66 \\

& InternVL3.5-8B~\citep{wang2025internvl3} & 4.53 & 98.67 & 99.88 & 98.68 & 9.23 & 86.03 & 96.66 & 82.52 & 51.36 & 50.29 & 28.58 & -0.08 \\

& Qwen2.5-VL-7B~\citep{bai2025qwen2} & 4.76 & 99.25 & 99.92 & 99.20 & 5.07 & 87.93 & 97.01 & 83.13 & 35.52 & 56.65 & 32.14 & 11.26 \\

& Phi-3.5-Vision~\citep{abdin2024phi} & 2.69 & 92.43 & 99.46 & 91.75 & 12.24 & 84.54 & 94.97 & 79.00 & 21.01 & 55.51 & 53.64 & 9.16 \\

& \multicolumn{13}{l}{\cellcolor{closedBand}\textbf{\strut Closed-source MLLMs}} \\

& GPT-4o~\citep{hurst2024gpt} & 7.82 & 93.48 & 99.25 & 92.79 & 54.85 & 52.80 & 23.68 & 3.71 & 14.98 & 70.53 & 64.77 & 43.97 \\

& GPT-4o-mini~\citep{hurst2024gpt} & 7.93 & 92.55 & 99.26 & 91.88 & 49.43 & 49.98 & 29.02 & 2.08 & 18.56 & 67.41 & 59.88 & 38.80 \\

\midrule
\multirow{9}{*}{\rotatebox[origin=c]{90}{\textbf{Verbal Confidence}}} &
\multicolumn{13}{l}{\cellcolor{groupBand}\textbf{\strut Open-source MLLMs}} \\

& MiniCPM-V-2.6~\citep{yao2024minicpm} & 14.90 & 52.61 & 99.64 & 45.05 & 14.30 & 65.34 & 87.83 & 42.22 & 37.19 & 58.01 & 49.51 & 19.77 \\

& InternVL2.5-4B~\citep{chen2024expanding} & 7.64 & 79.41 & 99.87 & 79.87 & 10.96 & 61.31 & 84.04 & 25.82 & 28.44 & 77.71 & 74.17 & 44.43 \\

& InternVL3.5-8B~\citep{wang2025internvl3} & 9.09 & 87.73 & 98.62 & 84.79 & 19.23 & 66.12 & 89.85 & 46.85 & 41.91 & 70.40 & 46.50 & 25.04 \\

& Qwen2.5-VL-7B~\citep{bai2025qwen2} & 12.91 & 69.17 & 95.60 & 55.37 & 11.82 & 56.97 & 86.50 & 23.72 & 47.68 & 71.99 & 33.76 & 13.38 \\

& Phi-3.5-Vision~\citep{abdin2024phi} & 36.86 & 66.91 & 96.76 & 50.14 & 33.83 & 59.79 & 83.17 & 29.73 & 28.20 & 67.42 & 69.02 & 39.31 \\

& \multicolumn{13}{l}{\cellcolor{closedBand}\textbf{\strut Closed-source MLLMs}} \\

& GPT-4o~\citep{hurst2024gpt} & 4.25 & 91.71 & 99.07 & 91.02 & 23.74 & 84.67 & 57.72 & 46.65 & 30.91 & 70.78 & 57.48 & 32.36 \\

& GPT-4o-mini~\citep{hurst2024gpt} & 16.10 & 90.03 & 98.98 & 88.75 & 14.28 & 76.73 & 52.17 & 34.02 & 19.19 & 65.43 & 49.78 & 23.40 \\

\bottomrule
\end{tabular}%
}
\end{subtable}

\end{table*}

%% file: table/calibrate.tex
\definecolor{headerBlue}{RGB}{255, 250, 240}   
\definecolor{subHeaderBlue}{RGB}{255, 250, 240} 
\definecolor{openStripe}{RGB}{255, 250, 240}    
\definecolor{groupBand}{RGB}{245, 235, 250}     
\definecolor{closedBand}{RGB}{255, 235, 235}    
\newcommand{\vd}[2]{\ensuremath{#1_{\textcolor{gray}{\scriptstyle(#2)}}}}
\newcommand{\vdBad}[2]{\ensuremath{#1_{\textcolor{red}{\scriptstyle(#2)}}}}

\begin{table*}[t]
\centering
\scriptsize
\setlength{\tabcolsep}{2.2pt}
\renewcommand{\arraystretch}{1.05}
\caption{Performance comparison under the calibrated setting. 
Calibrated Confidence denotes our final score. Subscripts indicate $\Delta$ relative gains, calculated against the stronger baseline: for ECE~$\downarrow$, $\Delta = \text{Calibrated} - \min(\text{Token}, \text{Verbalized})$; for other metrics (AUROC, AUPRC, AUPRC-N)~$\uparrow$, $\Delta = \text{Calibrated} - \max(\text{Token}, \text{Verbalized})$. All values are scaled by $10^{2}$.}
\label{tab:mm-6sets-prob-delta-fused-final}

\begin{subtable}[t]{\textwidth}
\centering
\resizebox{0.98\linewidth}{!}{%
\begin{tabular}{@{}l *{4}{c} @{\hspace{4pt}} *{4}{c} @{\hspace{4pt}} *{4}{c}@{}}
\toprule
\rowcolor{headerBlue}
\multicolumn{1}{l}{} &
\multicolumn{4}{c}{\textbf{MMBench}} &
\multicolumn{4}{c}{\textbf{MMStar}} &
\multicolumn{4}{c}{\textbf{ConBench}} \\
\cmidrule(lr){2-5}\cmidrule(lr){6-9}\cmidrule(lr){10-13}
\rowcolor{subHeaderBlue}
\textbf{Model} &
\textbf{ECE}$\downarrow$ & \textbf{AUROC}$\uparrow$ & \textbf{AUPRC}$\uparrow$ & \textbf{AUPRC-N}$\uparrow$ &
\textbf{ECE}$\downarrow$ & \textbf{AUROC}$\uparrow$ & \textbf{AUPRC}$\uparrow$ & \textbf{AUPRC-N}$\uparrow$ &
\textbf{ECE}$\downarrow$ & \textbf{AUROC}$\uparrow$ & \textbf{AUPRC}$\uparrow$ & \textbf{AUPRC-N}$\uparrow$ \\
\midrule
\rowcolor{groupBand}
\multicolumn{13}{l}{\hspace*{-0.5em}\textit{\textcolor{gray}{Open-source MLLMs}}} \\

MiniCPM-V-2.6~\citep{yao2024minicpm}
  & \vd{3.58}{-1.56} & \vd{90.34}{+0.37} & \vd{98.46}{+0.06} & \vd{88.45}{+0.44}
  & \vd{6.82}{-10.29} & \vd{72.44}{+1.75} & \vd{73.88}{-0.47} & \vd{48.46}{-0.92}
  & \vd{8.98}{-2.00} & \vd{74.60}{+4.41} & \vd{90.38}{+1.19} & \vd{60.92}{+4.82} \\
InternVL2.5-4B~\citep{chen2024expanding}
  & \vd{3.59}{-1.53} & \vd{88.99}{-0.23} & \vd{97.92}{-0.06} & \vd{85.82}{-0.42}
  & \vd{10.66}{-8.08} & \vd{64.67}{-2.21} & \vd{76.08}{-1.29} & \vd{40.68}{-3.22}
  & \vd{7.02}{-5.24} & \vd{82.37}{+0.53} & \vd{95.19}{+0.08} & \vd{74.69}{+0.45} \\

InternVL3.5-8B~\citep{wang2025internvl3}
  & \vd{3.35}{-2.28} & \vd{89.12}{+0.26} & \vd{98.68}{+0.02} & \vd{87.19}{+0.27}
  & \vd{7.05}{-13.07} & \vd{72.91}{+2.48} & \vd{85.70}{+1.01} & \vd{59.14}{+2.89}
  & \vd{5.94}{-7.18} & \vd{82.85}{-0.48} & \vd{93.41}{-0.67} & \vd{70.05}{-3.06} \\
Qwen2.5-VL-7B~\citep{bai2025qwen2}
  & \vd{2.24}{-2.20} & \vd{90.42}{+0.02} & \vd{98.45}{-0.08} & \vd{87.28}{-0.65}
  & \vd{8.68}{-3.62} & \vd{67.36}{-0.96} & \vd{77.30}{-2.16} & \vd{36.93}{-6.01}
  & \vd{5.91}{-3.41} & \vd{84.40}{-1.32} & \vd{93.31}{-1.98} & \vd{69.99}{-8.87} \\

Phi-3.5-Vision~\citep{abdin2024phi}
  & \vd{4.27}{-4.89} & \vd{86.43}{+0.21} & \vd{96.54}{+0.04} & \vd{82.37}{+0.17}
  & \vd{13.99}{-10.61} & \vd{56.64}{-4.46} & \vd{63.88}{-2.69} & \vd{29.18}{-5.28}
  & \vd{7.17}{-3.41} & \vd{80.38}{-0.40} & \vd{89.32}{+0.04} & \vd{65.00}{+0.16} \\
\cmidrule{1-13}
\rowcolor{closedBand}
\multicolumn{13}{l}{\hspace*{-0.5em}\textit{\textcolor{gray}{Closed-source MLLMs}}} \\
GPT-4o~\citep{hurst2024gpt}
  & \vd{1.92}{-4.12} & \vd{90.84}{-0.49} & \vd{98.75}{-0.06} & \vd{88.96}{-0.55}
  & \vd{7.74}{-13.94} & \vd{72.89}{-1.12} & \vd{76.48}{-0.98} & \vd{47.30}{-2.20}
  & \vd{7.56}{-0.85} & \vd{90.66}{+0.14} & \vd{98.24}{+0.02} & \vd{89.08}{+0.09} \\
GPT-4o-mini~\citep{hurst2024gpt}
  & \vd{5.16}{-2.92} & \vd{86.08}{-2.63} & \vd{95.99}{-1.57} & \vd{76.40}{-9.24}
  & \vd{3.18}{-15.88} & \vd{65.11}{-0.71} & \vd{60.47}{-2.33} & \vd{29.82}{-4.15}
  & \vd{7.07}{-7.03} & \vd{77.65}{-2.81} & \vd{91.79}{-1.22} & \vd{64.31}{-5.31} \\
\bottomrule
\end{tabular}}%
\end{subtable}

\vspace{4pt}

\begin{subtable}[t]{\textwidth}
\centering
\resizebox{0.98\linewidth}{!}{%
\begin{tabular}{@{}l *{4}{c} @{\hspace{4pt}} *{4}{c} @{\hspace{4pt}} *{4}{c}@{}}
\toprule
\rowcolor{headerBlue}
\multicolumn{1}{l}{} &
\multicolumn{4}{c}{\textbf{ScienceQA}} &
\multicolumn{4}{c}{\textbf{AI2D}} &
\multicolumn{4}{c}{\textbf{MMMU-Pro}} \\
\cmidrule(lr){2-5}\cmidrule(lr){6-9}\cmidrule(lr){10-13}
\rowcolor{subHeaderBlue}
\textbf{Model} &
\textbf{ECE}$\downarrow$ & \textbf{AUROC}$\uparrow$ & \textbf{AUPRC}$\uparrow$ & \textbf{AUPRC-N}$\uparrow$ &
\textbf{ECE}$\downarrow$ & \textbf{AUROC}$\uparrow$ & \textbf{AUPRC}$\uparrow$ & \textbf{AUPRC-N}$\uparrow$ &
\textbf{ECE}$\downarrow$ & \textbf{AUROC}$\uparrow$ & \textbf{AUPRC}$\uparrow$ & \textbf{AUPRC-N}$\uparrow$ \\
\midrule
\rowcolor{groupBand}
\multicolumn{13}{l}{\hspace*{-0.5em}\textit{\textcolor{gray}{Open-source MLLMs}}} \\

MiniCPM-V-2.6~\citep{yao2024minicpm}
  & \vd{1.58}{-0.30} & \vd{98.69}{0.00} & \vd{99.99}{0.00} & \vd{98.69}{0.00}
  & \vd{3.58}{-0.63} & \vd{86.38}{+0.77} & \vd{96.07}{+0.20} & \vd{81.33}{+0.95}
  & \vd{8.48}{-18.00} & \vd{46.46}{-11.55} & \vd{38.66}{-10.85} & \vd{2.53}{-17.24} \\
InternVL2.5-4B~\citep{chen2024expanding}
  & \vd{1.13}{+0.25} & \vd{92.81}{0.00} & \vd{99.95}{0.00} & \vd{92.57}{0.00}
  & \vd{4.52}{-3.61} & \vd{86.19}{+0.07} & \vd{95.98}{0.00} & \vd{81.33}{+0.02}
  & \vd{21.87}{-6.57} & \vd{75.04}{-2.67} & \vd{72.36}{-1.81} & \vd{40.51}{-3.92} \\

InternVL3.5-8B~\citep{wang2025internvl3}
  & \vd{3.19}{-1.34} & \vd{99.44}{+0.77} & \vd{99.95}{+0.07} & \vd{99.41}{+0.73}
  & \vd{3.80}{-5.43} & \vd{85.97}{-0.06} & \vd{96.58}{-0.08} & \vd{82.09}{-0.43}
  & \vd{2.41}{-39.50} & \vd{60.48}{-9.92} & \vd{40.73}{-5.77} & \vd{16.95}{-8.09} \\
Qwen2.5-VL-7B~\citep{bai2025qwen2}
  & \vd{3.57}{-1.19} & \vd{99.12}{-0.13} & \vd{99.90}{-0.02} & \vd{99.03}{-0.17}
  & \vd{3.54}{-1.53} & \vd{88.10}{+0.17} & \vd{97.24}{+0.23} & \vd{84.41}{+1.28}
  & \vd{10.56}{-24.96} & \vd{72.12}{+0.13} & \vd{38.64}{+4.88} & \vd{19.76}{+6.38} \\

Phi-3.5-Vision~\citep{abdin2024phi}
  & \vd{4.48}{+1.79} & \vd{92.50}{+0.07} & \vd{99.46}{0.00} & \vd{91.73}{-0.02}
  & \vd{4.71}{-7.53} & \vd{84.07}{-0.47} & \vd{94.67}{-0.30} & \vd{77.74}{-1.26}
  & \vd{10.47}{-10.54} & \vd{63.28}{-4.14} & \vd{62.66}{-6.36} & \vd{26.85}{-12.46} \\
\cmidrule{1-13}
\rowcolor{closedBand}
\multicolumn{13}{l}{\hspace*{-0.5em}\textit{\textcolor{gray}{Closed-source MLLMs}}} \\

GPT-4o~\citep{hurst2024gpt}
  & \vd{6.35}{+2.10} & \vd{93.89}{+0.41} & \vd{99.30}{+0.05} & \vd{93.26}{+0.47}
  & \vd{11.34}{-12.40} & \vd{82.66}{-2.01} & \vd{48.21}{-9.51} & \vd{34.66}{-11.99}
  & \vd{7.88}{-7.10} & \vd{75.20}{+4.42} & \vd{70.43}{+5.66} & \vd{52.96}{+8.99} \\

GPT-4o-mini~\citep{hurst2024gpt}
  & \vd{9.57}{+1.64} & \vd{93.01}{+0.46} & \vd{99.31}{+0.05} & \vd{92.40}{+0.52}
  & \vd{9.04}{-5.24} & \vd{76.11}{-0.62} & \vd{49.56}{-2.61} & \vd{30.42}{-3.60}
  & \vd{3.74}{-14.82} & \vd{69.36}{+1.95} & \vd{60.17}{+0.29} & \vd{39.25}{+0.45} \\
\bottomrule
\end{tabular}}%
\end{subtable}
\end{table*}

%% file: section/conclusion.tex
\section{Conclusion}\label{sec:conclusion} 
We approached MLLM confidence estimation by fusing token probabilities (instinct) and verbalized confidence (reflection). 
We demonstrated that their agreement is a robust signal and proposed a monotone fusion head to enable stable calibration. 
Future work will extend this framework to open-ended tasks and a broader spectrum of MLLM architectures. 
We aim to investigate internal design modifications to intrinsically enforce consistency between implicit and explicit uncertainty.
Ultimately, accurate confidence estimation is the cornerstone of trust, and we hope this work paves the way for deploying MLLMs in high-stakes environments where self-awareness is as critical as accuracy.


\newpage

\section*{Impact Statement}
This work aims to improve the reliability of multimodal LLMs by producing better-calibrated and more discriminative confidence estimates, which can support safer deployment via abstention, human-in-the-loop triage, and clearer uncertainty communication. 
Potential risks include over-reliance on a single confidence score, increased persuasiveness of wrong outputs when residual failures remain, and sensitivity to prompt/API changes. 
We therefore recommend using confidence as one component of a broader safety stack, monitoring calibration under distribution shift, and periodically revalidating the estimator as models and interfaces evolve.

%% file: section/appendix.tex
\appendix
\label{sec:appendix}

\input{section/appendix/dataset_model}

\input{section/appendix/method}

\section{Additional Experiment}
\input{section/appendix/risk_table}
\input{section/appendix/tempeature}
\input{section/appendix/calbratrion_visual}

\paragraph{Results of risk--coverage analysis.}
We extend our analysis to eight additional Multi-modal Large Language Models (MLLMs), comparing both closed-source models (e.g., Claude-3.7-sonnet, Gemini-2.0-Flash) and open-source models (e.g., Qwen2.5-VL, InternVL3.5).
Figure~\ref{fig:appendix_rc_curves} presents the risk--coverage (RC) curves for these models.
The results confirm the trends observed in the main text while highlighting performance differences between the two categories.
First, on the challenging MMMU-Pro benchmark, both closed-source and open-source models exhibit persistently high risk across all coverage levels.
This indicates that current confidence estimation methods struggle to reliably distinguish correct from incorrect answers on complex reasoning tasks, regardless of model access type.
Second, on standard benchmarks like MMBench and ScienceQA, most models demonstrate effective selective prediction, where the risk decreases significantly as we reduce coverage.
However, closed-source models generally achieve lower overall risk compared to open-source models, particularly in the high-coverage regions.
Finally, the non-monotonic behavior on MMStar is visible in models across both groups, such as GPT-4o-mini and Qwen2.5-VL.
This suggests that the misalignment between confidence and correctness on specific visual dependencies is a common issue shared by diverse model architectures.

\paragraph{Effect of Temperature on Calibration.}
To verify whether the temperature effects observed in the main text are universal, we extend our analysis to four additional state-of-the-art open-source models: GLM-4.1V-9B, InternVL3.5-8B~\citep{wang2025internvl3}, MiniCPM-V-2.6~\citep{yao2024minicpm}, and Phi-3.5-Vision~\citep{abdin2024phi}.
We report both AUPRC-N and ECE metrics across the same temperature sweep $T \in \{0.1, 0.5, 0.7, 0.9\}$.
Figure~\ref{fig:appendix_temp_more_models}, reveals results distinct from the main paper.
While the model in the main text showed improved calibration at higher temperatures (specifically on MMMU-Pro), the four models in the appendix display generally high stability, with only Phi-3.5-Vision showing minor sensitivity at higher temperatures.
ECE and AUPRC-N curves are nearly flat, indicating that varying the temperature does not significantly impact the confidence or the ranking quality of these models.
For example, on the difficult MMMU-Pro benchmark (green line), the calibration error remains high ($\text{ECE} \approx 0.4 - 0.5$) regardless of the temperature setting.
This suggests that calibration error is driven primarily by dataset difficulty and inherent model properties, rather than the decoding hyperparameters.
Consequently, simple temperature scaling is not a one-size-fits-all solution for improving calibration in multimodal generation.

\paragraph{Calibration and Reliability Analysis.}
To verify the generalizability of our findings, we extend the calibration analysis to four additional benchmarks: ConBench, AI2D, MMStar, and ScienceQA, as reported in Figure~\ref{fig:appendix_calibration}.
Consistent with the results on MMBench, the closed-source models (GPT-4o, GPT-5, and Claude-3.7-Sonnet) maintain superior calibration across these diverse tasks.
Their reliability diagrams generally align with the ideal diagonal, indicating that their verbalized confidence accurately reflects their actual predictive accuracy.
For example, on ScienceQA, these models achieve very low ECE scores ($\approx 0.06\text{--}0.07$), effectively distinguishing between correct and incorrect answers.
In contrast, the open-source model (Qwen2.5-VL-7B) consistently exhibits significant overconfidence.
Across most datasets, particularly AI2D and MMStar, its reliability curve deviates well below the diagonal, and the confidence histograms show a tendency to assign high probability scores ($0.9\text{--}1.0$) even to wrong predictions.
However, the disparity persists: closed-source models still provide relatively more reliable uncertainty estimates, whereas the open-source model remains prone to being "confidently wrong."
These extensive results confirm that the calibration gap between closed- and open-source MLLMs is a systematic issue that holds across different domains and difficulty levels.

\paragraph{Prompting Strategies for Confidence Estimation.}
To comprehensively evaluate MLLM confidence, we design and compare six distinct prompting strategies to elicit model confidence.
Specifically, the Vanilla (Verbalized) approach asks models to directly assign a 0-100 scalar score based on its holistic judgment.
To encourage deeper reasoning, we employ Chain-of-Thought (CoT) and Multi-Step methods, which require the model to break down the process into step-by-step explanations or aggregate confidence from sub-tasks.
We also explore introspection via Self-Probing and Roleplay, where the model critiques its own answers or adopts a cautious expert persona to avoid over-assertion.
Finally, the Confidence Interval strategy captures uncertainty intervals by asking for lower and upper bounds rather than a single point estimate.
Detailed illustrations of the specific prompt templates corresponding to these strategies are provided in Figure~\ref{fig:prompt-for-vanilla}, ~\ref{fig:prompt-templates-1}, ~\ref{fig:prompt-templates-2}.

\paragraph{Analysis of Confidence Distribution and Calibration.}
Figure~\ref{fig:confidence_histograms} illustrates the confidence histograms and calibration metrics across four MLLMs.
We observe a strong tendency toward overconfidence across all settings.
The distributions are heavily skewed to the 95\%--100\% interval.
This behavior is particularly rigid in Qwen2-VL-7B and Phi-3.5-Vision.
In the Vanilla setting, these models assign near-perfect scores even to incorrect predictions, represented by the prominent red bars in the highest bin.
Consequently, their ECE values remain high, reaching approximately 0.44.
On the other hand, MiniCPM-V-2.6 and InternVL2-5-4B exhibit more flexibility.
For these models, complex prompts like Roleplay and Confidence Interval successfully shift some probability mass to lower intervals.
Visually, this appears as a broader spread of confidence scores rather than a single spike at the top.
However, despite these minor distributional shifts, the models fail to effectively distinguish between correct and incorrect answers.
The persistence of high confidence for wrong answers suggests that prompting adjustments alone cannot resolve the intrinsic calibration errors.

\paragraph{Calibration Analysis of Prompting Strategies}
The cross-sectional comparisons across Table~\ref{tab:calib-selfprobing}, \ref{tab:calib-roleplay}, \ref{tab:calib-confbounds}, and \ref{tab:calib-cot} reveal the complex dynamics of prompting strategies in multimodal calibration.
These findings suggest that relying solely on prompt engineering is not a panacea for improving calibration performance.
Its effectiveness is significantly influenced by both the model and the task.
Specifically, complex strategies exhibit a "double-edged sword" effect: for instance, the Chain-of-Thought (CoT) strategy significantly improves calibration for Qwen2-VL-7B (with ECE decreasing from 16.77 to 6.99 on MMBench), yet it causes performance degradation for MiniCPM-V-2.6.
This suggests that for models with already well-calibrated confidence, forced reasoning may introduce noise.
Additionally, a clear decoupling phenomenon is observed between accuracy (ACC) and calibration (ECE).
For example, although the Roleplay strategy improves the accuracy of InternVL3.5-8B, it does not correspondingly reduce ECE on tasks such as AI2D.
It indicates that increased problem-solving ability does not necessarily equate to improved self-calibration.
More critically, in high-difficulty reasoning tasks such as MathVista, all models exhibit persistent overconfidence (with ECE consistently above 20.0) under various prompting strategies.
The Confidence Interval strategy even leads to a broad performance regression.
These results demonstrate that relying solely on prompt engineering is insufficient to address multimodal hallucination, further validating the superiority of the proposed dual-channel confidence fusion method for achieving robust calibration.

\input{section/appendix/acc}

\input{section/appendix/other_prompt_vis}

\input{section/appendix/other_prompt}

\input{section/appendix/prompt}

\paragraph{Case Study of Confidence}
As illustrated in Figures~\ref{fig:confidence_discrepancy_tcolorbox_1} and~\ref{fig:confidence_discrepancy_seedbench_2}, a distinct Verbal-Internal Disconnect (VID) pervades MLLMs, where implicit token confidence conceptually diverges from explicit verbalized confidence. To elucidate this dual-channel discrepancy, we dissect representative samples, systematizing them into four distinct groups defined by the intersection of prediction correctness and confidence alignment.

(1) Well-Calibrated Correct (WCC):
This is the best scenario where the model answers correctly and is also very sure about it. 
As shown in Figure~\ref{fig:confidence_wcc_mmmu_literature_65} and Figure~\ref{fig:confidence_wcc_artwork_10002}, the model picks the right option. 
More importantly, both the token confidence and verbal confidence are very high (close to 100\%). 
This shows that the model trusts its own answer, and users can trust it too.

(2) Under-Confident Correct (UCC): 
In these cases, the model implicitly identifies the correct answer but lacks sufficient verbal certainty. 
For example, in Figure~\ref{fig:confidence_discrepancy_seedbench_2} and Figure~\ref{fig:confidence_ucc_mmmu_public_health_4}, the model selects the right choice. 
The internal token confidence is notably high (90\%), which strongly suggests the model has indeed found the correct information.
However, the verbal confidence is surprisingly low (less than 60\% or even 25\%). 
This happens because the model gets distracted by other competing options when generating the final text, failing to properly express its true internal knowledge.

(3) Over-Confident Wrong (OCW): 
This is the most dangerous situation where the model makes a mistake but claims to be completely correct. 
As seen in Figure~\ref{fig:confidence_discrepancy_tcolorbox_3} and Figure~\ref{fig:confidence_discrepancy_seedbench_4}, the model chooses the wrong answer. 
Of particular concern is that the verbal confidence is extremely high (often 100\%). 
Even if the internal token confidence shows some doubt or supports another option, the final output sounds fully confident. 
This misleads users into thinking the wrong answer is true.

(4) Well-Calibrated Wrong (WCW):
Finally, Figure~\ref{fig:confidence_wcw_artwork_12652} and Figure~\ref{fig:confidence_wcw_ai2d_6} show a safer type of failure. Here, the model gives a wrong answer, but it explicitly admits that it is not sure. 
The verbal confidence is appropriately low (below 50\%), which matches the fact that the answer is incorrect. 
Although the model might still have some internal confusion, the low confidence score effectively warns the user to exercise caution, signaling that the generated answer might be unreliable.

\input{section/appendix/casestudy}

%% file: section/appendix/dataset_model.tex
\section{Detailed Experiment Settings.}

\paragraph{Benchmarks.}
To comprehensively assess multimodal models in terms of general visual understanding, reasoning over diverse visual evidence (natural images, diagrams, and mixed text-image content), and knowledge-intensive problem solving, we conduct experiments on six established benchmarks, including MMBench~\citep{liu2024mmbench}, MMStar~\citep{chen2024we}, ConBench~\citep{zhang2024unveiling}, ScienceQA~\citep{lu2022learn}, AI2D~\citep{kembhavi2016diagram}, and MMMU-Pro~\citep{yue2024mmmu}. These six datasets constitute our primary experimental suite. Furthermore, we provide supplementary results on MathVista~\citep{lu2023mathvista} in the appendix. These benchmarks jointly cover (i) broad, real-world visual perception and instruction following, (ii) challenging multimodal reasoning with strong distractors, (iii) compositional/consistency-sensitive evaluation settings, (iv) science-oriented multimodal QA requiring domain knowledge, (v) diagram understanding with structured visual relations, and (vi) expert-level multi-discipline multimodal understanding. Unless otherwise specified, we follow the official dataset splits and evaluation scripts to ensure reproducibility and fair comparison.

\textbf{MMBench}~\citep{liu2024mmbench} is a large-scale multi-choice benchmark designed to evaluate general-purpose visual understanding and reasoning across diverse image domains. It probes a model's comprehensive ability to seamlessly integrate complex visual cues with linguistic instructions through a fine-grained taxonomy of 20 ability dimensions. By incorporating rigorous evaluation protocols like CircularEval, it serves as a robust testbed for examining not only prediction accuracy but also the alignment between model confidence and actual performance across diverse visual tasks.

\textbf{MMStar}~\citep{chen2024we} is a benchmark designed for high-difficulty multimodal reasoning, emphasizing the critical visual indispensability across all test questions. It targets settings with harder distractors and increased compositional complexity, effectively stressing fine-grained visual grounding and robust reasoning beyond superficial cues.

\textbf{ConBench}~\citep{zhang2024unveiling} focuses on compositional and/or consistency-related aspects of multimodal reasoning, emphasizing whether a model can reliably handle controlled variations and multi-step reasoning patterns. It is particularly useful for stress-testing robustness under compositional changes and fine-grained content-level perturbations, while remaining critical for analyzing confidence behavior under harder conditions.

\textbf{ScienceQA}~\citep{lu2022learn} evaluates science-centric multimodal question answering where solutions often require combining rich visual evidence (e.g., figures, diagrams, or illustrative images) with domain-specific scientific knowledge and reasoning. It therefore serves as a proxy for evaluating knowledge-intensive multimodal problem solving.

\textbf{AI2D}~\citep{kembhavi2016diagram} is a diagram understanding benchmark that assesses a model's inherent capability to interpret structured visual elements such as labels, arrows, layout, and relational cues. It emphasizes logical reasoning over diagram semantics rather than purely natural-image perception. Crucially, it requires the model to parse the complex topology between textual annotations and visual icons to resolve challenging multi-step scientific queries.

\textbf{MMMU-Pro}~\citep{yue2024mmmu} measures expert-level multimodal understanding spanning multiple disciplines, specifically emphasizing visual indispensability. It requires deeper reasoning and broader knowledge coverage. Its difficulty makes it a strong indicator of advanced multimodal competence under challenging evaluation settings.

\textbf{MathVista}~\citep{lu2023mathvista} is a comprehensive benchmark designed to evaluate mathematical reasoning within diverse visual contexts. 
It consolidates thirty-one distinct datasets, including three newly proposed challenging subsets—IQTest, FunctionQA, and PaperSection—to push the boundaries of visual logic. 
The benchmark systematically categorizes tasks into seven logical reasoning abilities and five visual contexts.
Unlike conventional perception tasks, MathVista requires a sophisticated synergy of fine-grained visual grounding and complex, multi-step logical deduction.

\paragraph{Metrics.}
We report both task performance and reliability-oriented metrics to provide a holistic evaluation. Specifically, \textbf{ACC} measures end-task correctness, while \textbf{ECE} quantifies probabilistic calibration (how well predicted confidence matches empirical accuracy). 
In addition, we evaluate the discriminative ability of model confidence using threshold-free ranking metrics: AUROC, AUPRC, and its normalized variant AUPRC-N. Finally, we employ AURC to assess the potential for risk-sensitive selective prediction. Unless otherwise specified, higher values are better for ACC, AUROC, AUPRC, and AUPRC-N, while lower values are preferred for ECE and AURC.

\textbf{ACC ($\uparrow$).}
Accuracy is the fraction of total examples for which the model produces the correct answer (e.g., the correct option in multiple-choice settings). It is widely considered as the primary metric for evaluating overall end-task performance.

\textbf{ECE ($\downarrow$).}
Expected Calibration Error measures the gap between predicted confidence and empirical accuracy, typically computed by partitioning predictions into confidence bins and aggregating the absolute difference between average confidence and accuracy per bin. Lower ECE indicates better-calibrated confidence estimates.

\textbf{AUROC ($\uparrow$).}
Area Under the ROC Curve evaluates how well a scalar confidence score ranks correct predictions above incorrect ones across all possible thresholds. AUROC is insensitive to the choice of a specific operating point and reflects overall separability. Specifically, a score of 1.0 signifies perfect discrimination between the two classes.

\textbf{AUPRC ($\uparrow$).}
Area Under the Precision--Recall Curve emphasizes performance under class imbalance and is often more informative than AUROC when the positive class is relatively rare. Here, we treat correct predictions as positives and compute AUPRC from the continuous confidence-based rankings to assess reliability thoroughly.

\textbf{AUPRC-N ($\uparrow$).}
AUPRC-N normalizes the Precision-Recall area against task-specific random baselines. Unlike standard $\text{AUPRC}$, it mitigates sensitivity to class priors, enabling fair reliability comparisons across diverse benchmarks by explicitly decoupling ranking quality from the task's intrinsic accuracy to measure the performance gain over random chance.



\textbf{AURC ($\downarrow$).} Area Under the Risk-Coverage Curve serves as a pivotal metric for evaluating a model's uncertainty quantification capabilities. By reflecting the model's self-awareness, it exhibits high sensitivity to probability calibration, making it indispensable for assessing performance in safety-critical scenarios.
Specifically, it measures the cumulative trade-off between coverage (fraction of the samples predicted) and selective risk (the error rate on those samples).

%% file: section/appendix/method.tex

\section{Additional Implementation and Theoretical Details}\label{sec:appendix-details}
This appendix details the end-to-end procedure for reproducing our reliability estimation pipeline, which encompasses four cardinal stages: (i) extracting and quantifying dual-channel confidence metrics, (ii) constructing monotonic reliability features, (iii) fitting the constrained monotonic fusion head, and (iv) executing order-preserving mean alignment.

\subsection{End-to-end procedure and dataset splits}\label{app:procedure}

\textbf{Data splits and leakage prevention.}
We assume the original benchmark provides a test set that is never used for any fitting or tuning.
From the remaining data (or a held-out calibration corpus), we partition the samples into:
(i) a calibration split to fit the fusion head parameters $\theta=(b,\{\tilde w_j\})$ in~\eqref{eq:nll};
(ii) a validation split to solve mean-alignment~\eqref{eq:meanalign}.
When data is limited, we recommend $k$-fold cross-fitting: fit $\theta$ on fold-complements and evaluate on held-out folds.
Finally, we aggregate these out-of-sample predictions to solve for the mean-alignment shift on the entire corpus to avoid optimistic bias.

\textbf{Deterministic answer protocol.} 
To obtain comparable option-level scores and multi-dimensional confidence, we enforce a structured response format. Specifically, the MLLM is constrained to output the predicted option label token (e.g., "A"..."D" or "1"..."4") and its verbalized confidence score within a standardized and parseable template (e.g., a JSON object). We treat the model as a black box and directly extract the token probabilities for the option label at its specific generation index. This makes the token confidence $p_i$ in~\eqref{eq:prob-softmax} well-defined and consistently reproducible across multiple inference runs.

\begin{figure}[t]
\centering
\fbox{
\begin{minipage}{0.95\linewidth}
\small
\textbf{Algorithm A.1 (Monotone Confidence Fusion with mean alignment).}
For each instance $(x_i,q_i,\mathcal{A}_i)$:
\begin{enumerate}
  \item \textbf{Token confidence channel (instinct).} Compute option log-probabilities $\{z_{ik}\}_{k=1}^K$ at the answer position and
        form $p_i=\softmax(z_i)$ as in~\eqref{eq:prob-softmax}.
        Let $k_i^\star=\arg\max_k p_{ik}$ and $r_i^{\text{token}}=\max_k p_{ik}$.
  \item \textbf{Verbalized confidence channel (reflection).} Prompt the MLLM to output per-option confidence (Appendix~\ref{app:prompt-parse});
        parse into $s_i\in[0,1]^K$ and set $r_i^{\text{verbal}}=s_{i,k_i^\star}$.
  \item \textbf{Cross-channel consistency.} Compute $\kappa_{ik}$ by~\eqref{eq:consistency-coeff} and set $r_i^{\text{cons}}=\kappa_{i,k_i^\star}$.
  \item \textbf{Features.} Form $\phi_i$ by~\eqref{eq:feat} using $\ell(\cdot)$ in~\eqref{eq:logitclip}, margin~\eqref{eq:gap}, and entropy~\eqref{eq:entropy}.
\end{enumerate}
\textbf{Fitting:} Standardize features using calibration-split statistics; fit $\theta$ by minimizing~\eqref{eq:nll} with $w_j=\softplus(\tilde w_j)$.
\textbf{Mean alignment:} Solve $\delta$ from~\eqref{eq:meanalign} on validation split (Appendix~\ref{app:meanalign}).
\textbf{Inference:} Output merged confidence $\hat q_i=\sigma(f_\theta(\phi_i)+\delta)$.
\end{minipage}}
\end{figure}

\subsection{Prompting and parsing verbalized confidence}\label{app:prompt-parse}
We obtain per-option verbalized confidence via a structured instruction that constrains the output format.
A representative template is:
\textbf{Instruction:} You are given an image and a multiple-choice question with $K$ options.
First, output exactly one option label as your final answer.
Then, output a JSON object only with keys \{1,\ldots,K\} and numeric values in \{0,\ldots,100\} indicating your confidence percentage for each option.
Do not include any additional text.

\textbf{Decoding settings for robustness.}
We use deterministic decoding (e.g., temperature $0$) to reduce formatting variance.
We constrain the maximum generation length so that the model has sufficient budget to output a complete JSON object but not enough to drift into free-form text.
If supported, we additionally enforce stop sequences after the closing brace \texttt{\}}.

\textbf{Parsing and normalization.}
We parse the JSON object into $s_i\in[0,1]^K$ by:
(i) extracting numeric values (allowing integers or floats),
(ii) dividing by $100$, and (iii) clipping each entry to $[0,1]$.
If the output is malformed, we fall back to a regex-based extractor that matches \texttt{(option\_id, number)} pairs
and applies the same normalization.

\textbf{Missing keys and out-of-range values.}
To maintain data integrity, we implement a robust preprocessing pipeline for the input features.
For any missing option keys, we impute a neutral value of $50$ and record the missing indicator.
To handle numerical instability and outliers, values falling outside the $[0, 100]$ range are clipped to the nearest boundary before dividing by $100$.
This design ensures $s_i$ is always well-defined without silently discarding examples.

\textbf{Why we do not enforce $\sum_k s_{ik}=1$.}
Unlike $p_i\in\Delta^{K-1}$, verbalized confidence can legitimately violate normalization:
the model may express global uncertainty (all low) or multi-modality (several high).
For this reason, the main text treats $s_i$ as a hypercube vector and only uses
the predicted-option component $r_i^{\text{verbal}}=s_{i,k_i^\star}$ together with cross-channel agreement.

\subsection{Extracting token-option confidence}\label{app:dist-extract}
The token confidence in~\eqref{eq:prob-vector} requires model-retrieved log-probabilities for each option-label token at the answer position.

\textbf{Single-token option labels.}
We choose option labels that are single tokens under the model tokenizer (e.g., \texttt{A/B/C/D} or \texttt{1/2/3/4} depending on the model).
Let the prompt end with a fixed prefix such as ~\texttt{Final answer:} and let the next generated token be strictly restricted (implicitly or explicitly)
to one of the $K$ predefined labels.
We obtain the log-probability vector $z_i=(z_{i1},\ldots,z_{iK})$ for these label tokens and compute
token confidence $p_i=\softmax(z_i)$ as in~\eqref{eq:prob-softmax}.

\textbf{Numerical Stability and Reproducibility.}
Since the token confidence score $p_i$ is derived directly from a single-step next-token distribution, it avoids the potential cascading error accumulation inherent in multi-step sequence generation. 
This approach exhibits inherent robustness to long-horizon decoding noise and sampling stochasticity.
Consequently, we enforce the standardized single-label token protocol to ensure both determinism and experimental reproducibility.

\subsection{Reliability features: construction, bounds, and standardization}\label{app:features}
We construct a compact reliability descriptor $\phi_i\in\mathbb{R}^d$ combining both channels' confidence features as in~\eqref{eq:feat}:
\begin{equation}
\phi_i=
\Big[
\ell(r_i^{\text{token}}),\;
\ell(r_i^{\text{verbal}}),\;
\ell(r_i^{\text{cons}}),\;
m_i,\;
-H(p_i)
\Big].
\end{equation}

\textbf{Log-odds transform with clipping.}
The map $\ell(\cdot)$ in~\eqref{eq:logitclip} turns heterogeneous $(0,1)$ confidence values into an approximately linear domain for better regression stability.
We clip inputs to $[\varepsilon,1-\varepsilon]$ to avoid infinities. In practice, $\varepsilon$ can be $10^{-6}$.

\textbf{Margin and entropy are complementary sharpness signals.}
The margin $m_i=p_{i,(1)}-p_{i,(2)}\in[0,1]$ captures top-1 separability, while $H(p_i)\in[0,\log K]$ captures global dispersion.
We include both because they respond differently to multi-modal distributions: a distribution may exhibit small margin but low entropy if probability mass concentrates on a few options (e.g., $p_i = [0.45, 0.45, 0, 0]$, where $m_i=0$ but $H(p_i)$ remains low due to the pruned label space).

\textbf{Sign conventions and monotone intent.}
We specifically use $-H(p_i)$ so that larger values indicate lower entropy and thus higher reliability, thereby matching the monotone prior established in the main text analysis.

\textbf{Standardization preserves monotonicity.}
We standardize each feature dimension using calibration-split mean $\mu_j$ and standard deviation $\sigma_j>0$:
$\tilde\phi_{ij}=(\phi_{ij}-\mu_j)/\sigma_j$.
Since this is an affine transform with positive scaling, and assuming the non-negativity constraint $w_j \ge 0$, the fused predictor remains coordinatewise nondecreasing in the original features:
\begin{equation}
\frac{\partial \hat q_i}{\partial \phi_{ij}}
=\frac{\partial \hat q_i}{\partial \tilde\phi_{ij}}\cdot\frac{\partial \tilde\phi_{ij}}{\partial \phi_{ij}}
=\sigma'\!\big(f_\theta(\tilde\phi_i)\big)\,w_j\cdot\frac{1}{\sigma_j}\ge 0.
\end{equation}
If a feature has near-zero variance on the calibration split, we drop it or set $\sigma_j=1$ and report this preprocessing.

\subsection{Monotone fusion head: optimization, interpretability, and special cases}\label{app:monotone-head}
We predict correctness probability with
$\hat q_i=\sigma\!\big(f_\theta(\phi_i)\big)$,
$f_\theta(\phi_i)=b+\sum_{j=1}^d w_j\phi_{ij}$, and $w_j=\softplus(\tilde w_j)\ge 0$.


\textbf{Coordinate-wise Monotonicity \& Interpretability.}
The constraint $w_j \geq 0$ enforces a monotonic inductive bias, ensuring that predicted correctness is non-decreasing with respect to each reliability cue. In low-data regimes, this prevents overfitting to counter-intuitive negative weights. Since standardized features ensure higher values signify greater reliability, the magnitude of $w_j$ directly quantifies each cue's marginal contribution. In practice, we enforce this by reparameterizing the weights through a Softplus function, ensuring the calibration adheres to first-order stochastic dominance.


\textbf{Theoretical Connections \& Optimization.} 
Our formulation generalizes standard post-hoc calibration; specifically, when $d=1$ and $\phi_i=\ell(r_i^{\text{token}})$, the framework reduces to a classic one-dimensional logistic calibrator (e.g., Platt scaling). This allows our approach to subsume common baselines while enabling multi-signal fusion within a single interpretable framework. We learn the parameters $\{\tilde{w}, b\}$ by minimizing the negative log-likelihood~\eqref{eq:nll} on the calibration split. Given the low dimensionality of $\phi_i$, the optimization is inherently stable. We employ the Adam optimizer with a small learning rate and mild weight decay, applying early stopping based on validation ECE or NLL to ensure robust generalization.

\subsection{Order-preserving mean alignment: theory and numerics}\label{app:meanalign-theory}

\noindent\textit{Corollary 1 (Ranking invariance).}
Updating $b$ by~\eqref{eq:bshift} adds the same constant to all output logits $\{f_\theta(\phi_i)\}_i$,
preserving their relative ordering; thus the risk--coverage curves and any selection rule based on ranking are unchanged.

We apply a post-hoc bias shift $b\leftarrow b+\delta$, with $\delta$ to satisfy~\eqref{eq:meanalign} using validation split statistics.
This step is intentionally order-preserving (Corollary 1) and thus does not change any ranking-based evaluation
(e.g., risk--coverage curves), while effectively correcting systematic base-rate misalignment arising from the initial calibration-split optimization.

\textbf{Why a one-dimensional correction is often sufficient.}
In many deployment shifts, the primary change is the base accuracy (e.g., questions become slightly harder),
while the relative ordering of easy vs.\ hard instances remains informative.
A pure bias shift exactly targets this scenario without refitting weights and without altering selection ordering.

\subsection{Solving mean alignment by bisection}\label{app:meanalign}
To robustly solve~\eqref{eq:meanalign}, we clip $\mathrm{ACC}$ to $[\varepsilon,1-\varepsilon]$ with a small $\varepsilon$ (e.g., $10^{-6}$).
We then find $\delta$ by bisection on an interval $[\delta_{\min},\delta_{\max}]$ chosen such that
$g(\delta_{\min})\le \mathrm{ACC}\le g(\delta_{\max})$, where
$g(\delta)=\frac{1}{n}\sum_{i=1}^n \sigma(f_\theta(\phi_i)+\delta)$.
A practical choice is
\begin{equation}
\delta_{\min}=-M,\qquad \delta_{\max}=M,\qquad M=20,
\end{equation}
since $\sigma(t\pm M)$ is numerically saturated for typical floating-point ranges.
We iterate the bisection algorithm until the absolute residual $|g(\delta)-\mathrm{ACC}|$ finally falls below a predefined convergence tolerance (e.g., $10^{-8}$).

\subsection{Theoretical Guarantees}\label{app:proofs}

\noindent\textit{Proposition 1 (Coordinatewise monotonicity).}
For any $j$, $\partial \hat q_i/\partial \phi_{ij}=\sigma'\!\big(f_\theta(\phi_i)\big)\,w_j\ge 0$.
Hence the fused score~\eqref{eq:head} is nondecreasing in each feature, ensuring interpretable alignment with input reliability signals (Appendix~\ref{app:proofs}).

\noindent\textit{Theorem 1 (Existence and uniqueness of $\delta$).}
Define the mean predicted probability function $g(\delta)$ as
\begin{equation}
g(\delta)=\frac{1}{n}\sum_{i=1}^{n}\sigma\big(f_\theta(\phi_i)+\delta\big).
\end{equation}
Then $g$ is continuous and strictly increasing in $\delta$. Hence for any $\mathrm{ACC}\in(0,1)$,
equation~\eqref{eq:meanalign} admits a unique solution.

\textbf{Proof of Proposition 1.}

Recall $\hat q_i=\sigma(f_\theta(\phi_i))$ with $f_\theta(\phi_i)=b+\sum_{j=1}^d w_j\phi_{ij}$ and $w_j\ge 0$.
For any coordinate $j$,
\begin{equation}
\frac{\partial \hat q_i}{\partial \phi_{ij}}
=\sigma'\!\big(f_\theta(\phi_i)\big)\cdot \frac{\partial f_\theta(\phi_i)}{\partial \phi_{ij}}
=\sigma'\!\big(f_\theta(\phi_i)\big)\,w_j.
\end{equation}
Since $\sigma'(t)=\sigma(t)(1-\sigma(t))>0$ for all $t$ and $w_j\ge 0$, we obtain
$\partial \hat q_i/\partial \phi_{ij}\ge 0$, proving coordinatewise monotonicity.

\textbf{Proof of Theorem 1.}
Define $g(\delta)=\frac{1}{n}\sum_{i=1}^{n}\sigma(f_\theta(\phi_i)+\delta)$.
Each term $\sigma(f_\theta(\phi_i)+\delta)$ is continuous and strictly increasing in $\delta$;
finite sums preserve these properties, so $g$ is continuous and strictly increasing.
Moreover, $\lim_{\delta\to -\infty}g(\delta)=0$ and $\lim_{\delta\to +\infty}g(\delta)=1$.
Therefore, for any $\mathrm{ACC}\in(0,1)$ there exists a unique $\delta$ such that $g(\delta)=\mathrm{ACC}$.

\subsection{Complexity and practical considerations}\label{app:complexity}
\textbf{Per-instance overhead.}
Computing $p_i$ requires obtaining log-probabilities for $K$ label tokens at one answer position; this is $O(K)$ additional work after the forward pass.
Parsing $s_i$ is linear in the output length and negligible in practice.
The mean-alignment bisection runs in $O(n\log(1/\eta))$ time for tolerance $\eta$ and is performed once per evaluation setting.

\textbf{Failure modes and mitigations.}
(i) Malformed JSON: handled by robust regex fallback and missing-key imputation to ensure data integrity (Appendix~\ref{app:prompt-parse}).
(ii) Extremal probabilities (0 or 1): managed via numerical clipping in~\eqref{eq:logitclip}.
(iii) Tokenizer mismatch for option labels: verify that each label is single-token before running experiments.

\subsection{Hyperparameters and recommended ranges}
\label{app:hyperparams}
We summarize the key hyperparameters and their recommended ranges to facilitate reproducibility. 
For numerical stability in the loss function $\ell(\cdot)$ and accuracy computation, we introduce a clipping threshold $\varepsilon$, defaulting to $10^{-6}$ within a practical range of $[10^{-8}, 10^{-4}]$. 
Regarding the consistency coefficient in Eq.~\eqref{eq:consistency-coeff}, we set the shape parameter $\gamma$ to $2$ (selecting from $\{1, 2\}$) and determine the bandwidth $\tau$ via a coarse grid search on the validation set. 
Furthermore, the bisection algorithm is configured with a bracket size $M=20$ (typically $10$--$30$) and a tolerance $\eta=10^{-8}$ (ranging from $10^{-10}$ to $10^{-6}$).

%% file: section/appendix/risk_table.tex
\begin{figure*}[t]
    \centering

    \begin{subfigure}[t]{0.25\textwidth}
        \centering
        \includegraphics[width=\linewidth]{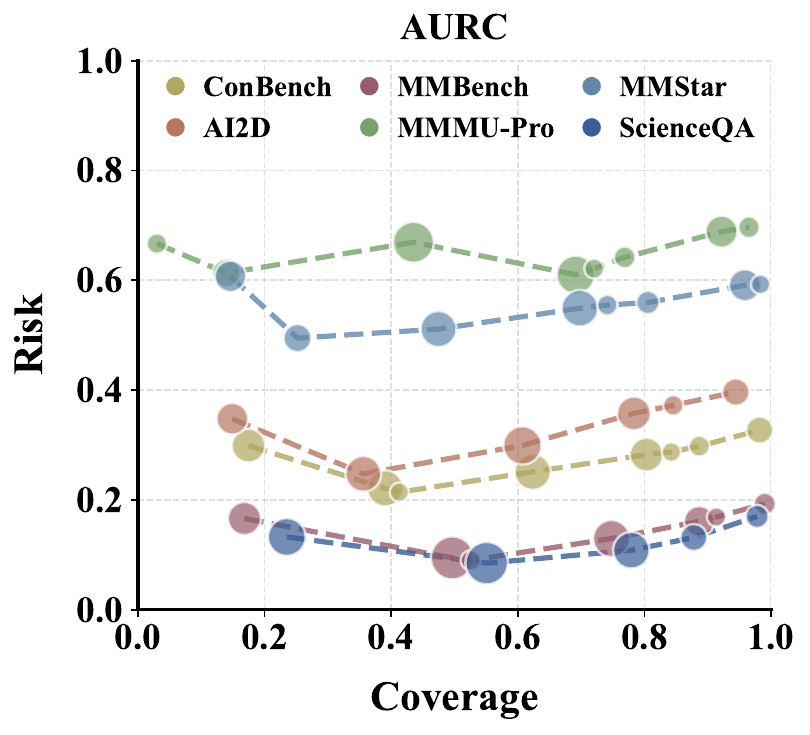}
        \caption{GPT-4o-mini}
    \end{subfigure}\hfill
    \begin{subfigure}[t]{0.25\textwidth}
        \centering
        \includegraphics[width=\linewidth]{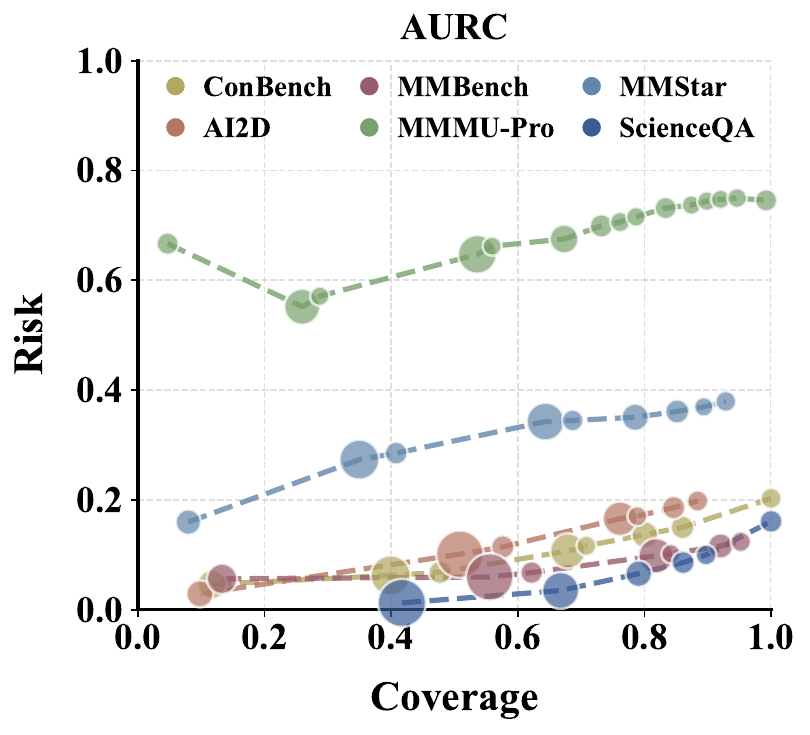}
        \caption{Claude-3.7-sonnet}
    \end{subfigure}\hfill
    \begin{subfigure}[t]{0.25\textwidth}
        \centering
        \includegraphics[width=\linewidth]{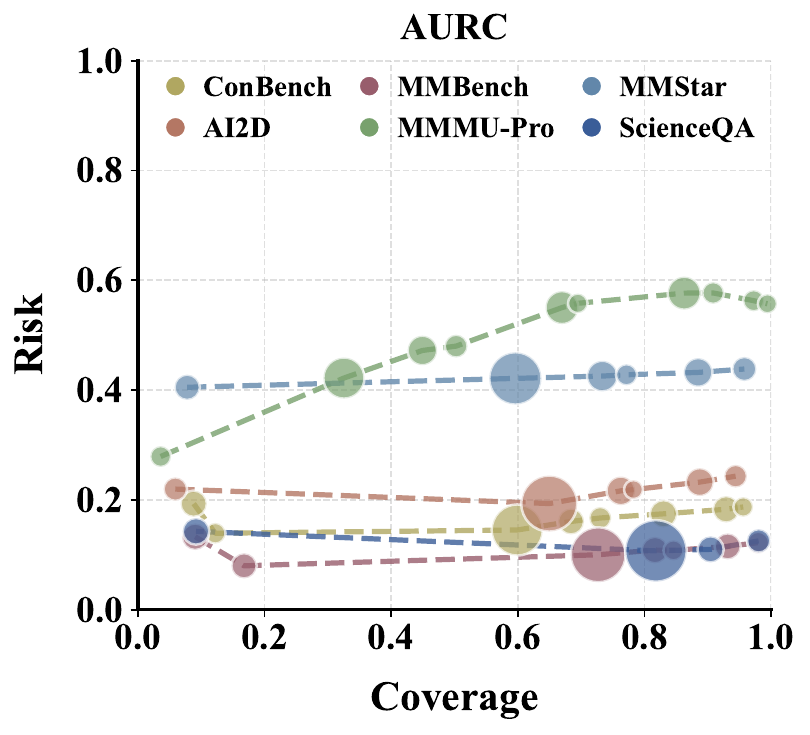}
        \caption{Gemini-2.0-Flash}
    \end{subfigure}\hfill
    \begin{subfigure}[t]{0.25\textwidth}
        \centering
        \includegraphics[width=\linewidth]{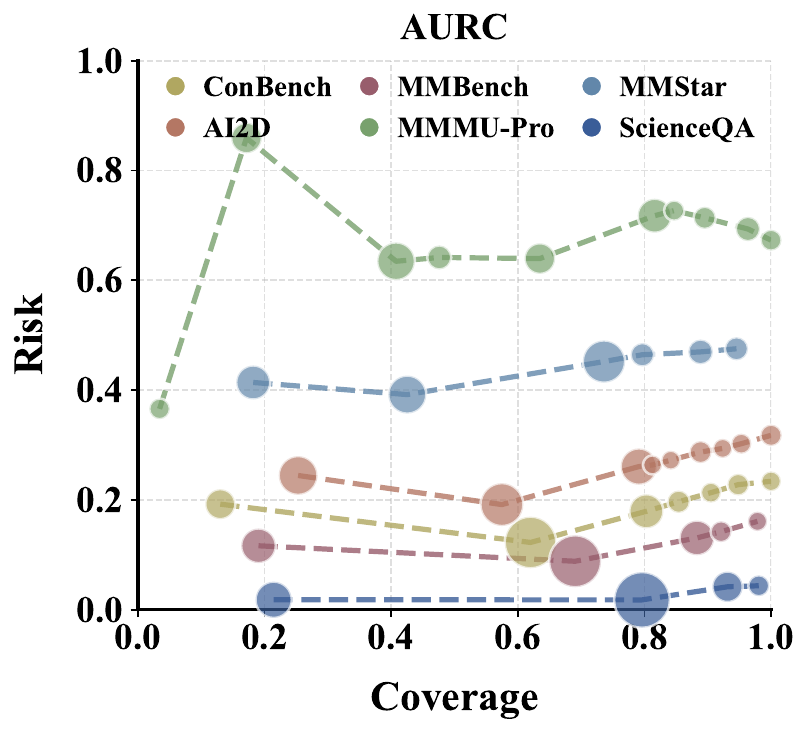}
        \caption{GLM-4.1V-9B}
    \end{subfigure}

    \vspace{2mm}

    \begin{subfigure}[t]{0.25\textwidth}
        \centering
        \includegraphics[width=\linewidth]{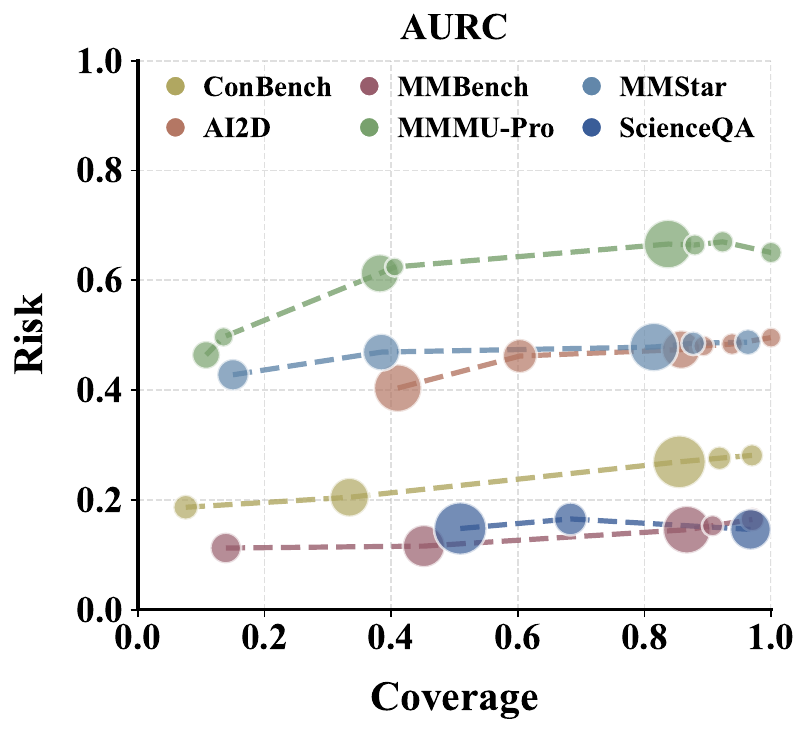}
        \caption{MiniCPM-V-2.6}
    \end{subfigure}\hfill
    \begin{subfigure}[t]{0.25\textwidth}
        \centering
        \includegraphics[width=\linewidth]{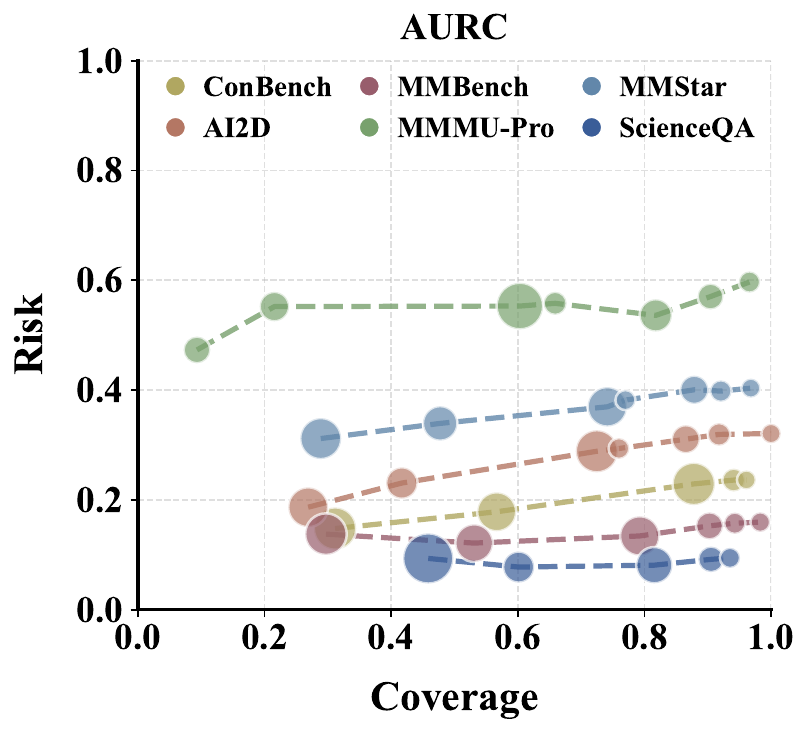}
        \caption{InternVL3.5-8B}
    \end{subfigure}\hfill
    \begin{subfigure}[t]{0.25\textwidth}
        \centering
        \includegraphics[width=\linewidth]{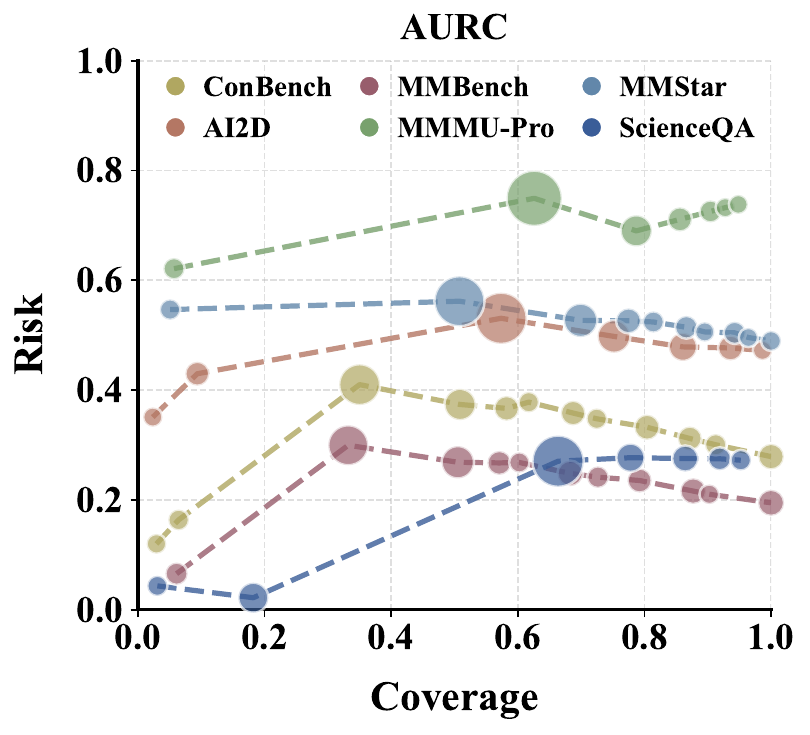}
        \caption{Qwen2.5-VL-7B}
    \end{subfigure}\hfill
    \begin{subfigure}[t]{0.25\textwidth}
        \centering
        \includegraphics[width=\linewidth]{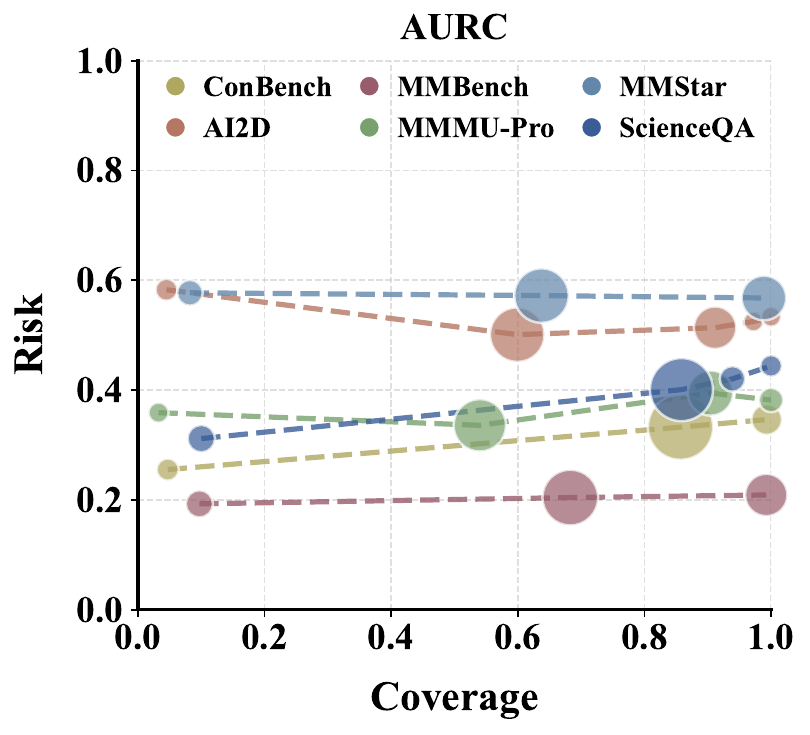}
        \caption{Phi-3.5-Vision}
    \end{subfigure}
    \caption{
        \textbf{Risk--coverage analysis for eight comparative MLLMs.} 
        We report the risk ($1 - \text{accuracy}$) at varying coverage levels across six multi-modal benchmarks. 
        The top row displays results for closed-source models (first three) and GLM-4.1V-9B, while the bottom row shows open-source models.
        Across both groups, ScienceQA and MMBench (bottom curves) show high reliability, whereas MMMU-Pro (top curves) remains challenging with high risk, indicating limited effectiveness of confidence-based selection on expert-level tasks.
    }
    \label{fig:appendix_rc_curves}
\end{figure*}

%% file: section/appendix/tempeature.tex
\begin{figure*}[t]
    \centering

    \begin{subfigure}[t]{0.25\textwidth}
        \centering
        \includegraphics[width=\linewidth]{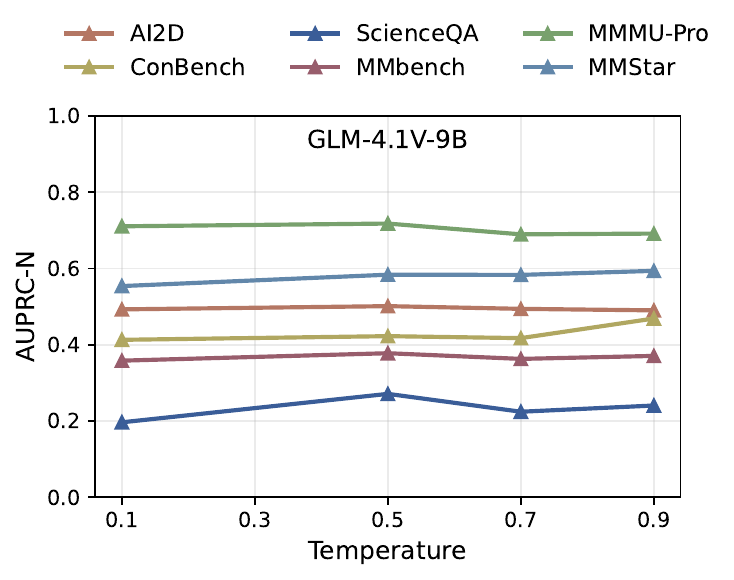}

    \end{subfigure}\hfill
    \begin{subfigure}[t]{0.25\textwidth}
        \centering
        \includegraphics[width=\linewidth]{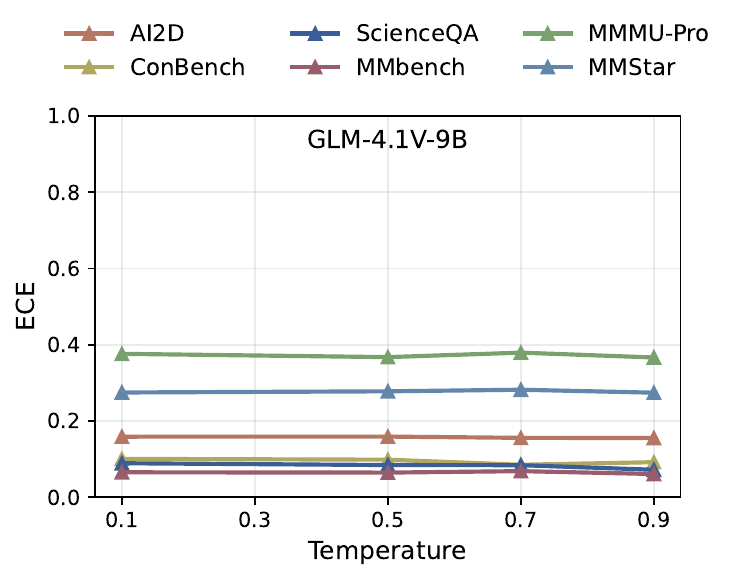}

    \end{subfigure}\hfill
    \begin{subfigure}[t]{0.25\textwidth}
        \centering
        \includegraphics[width=\linewidth]{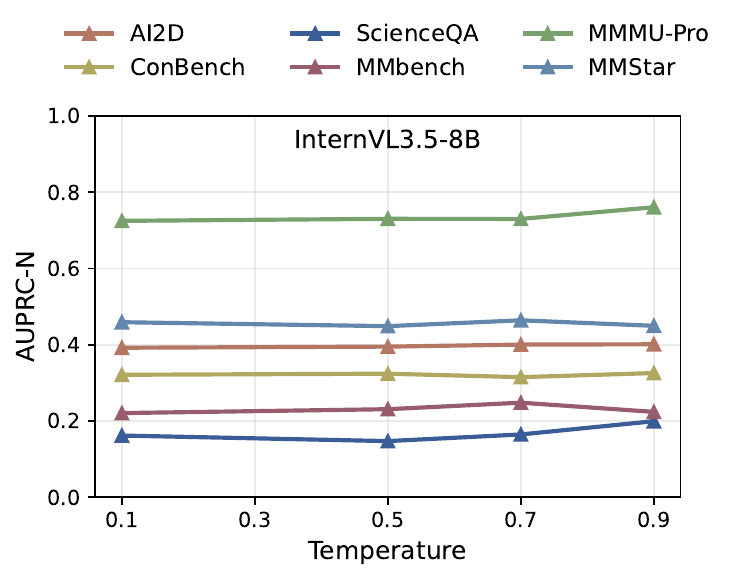}

    \end{subfigure}\hfill
    \begin{subfigure}[t]{0.25\textwidth}
        \centering
        \includegraphics[width=\linewidth]{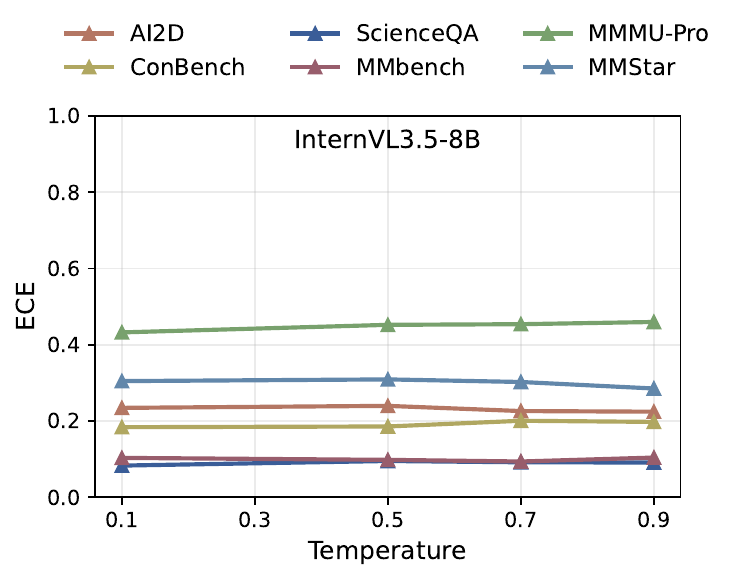}

    \end{subfigure}

    \vspace{2mm}

    \begin{subfigure}[t]{0.25\textwidth}
        \centering
        \includegraphics[width=\linewidth]{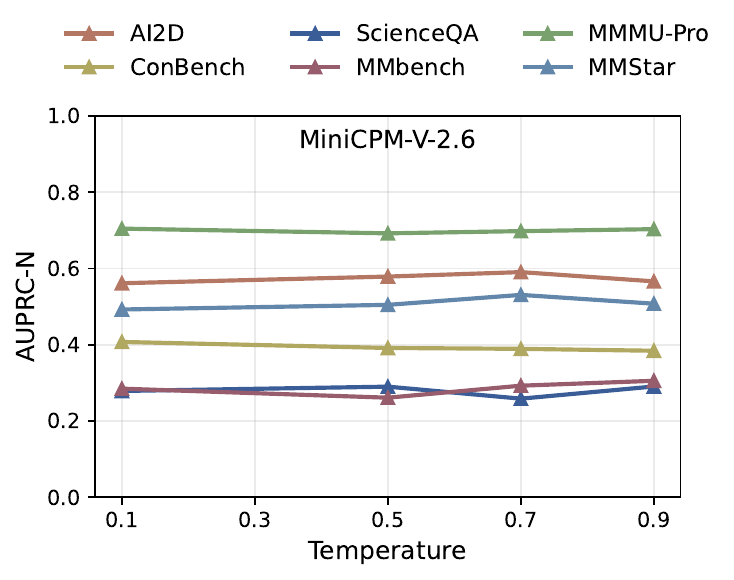}
        
    \end{subfigure}\hfill
    \begin{subfigure}[t]{0.25\textwidth}
        \centering
        \includegraphics[width=\linewidth]{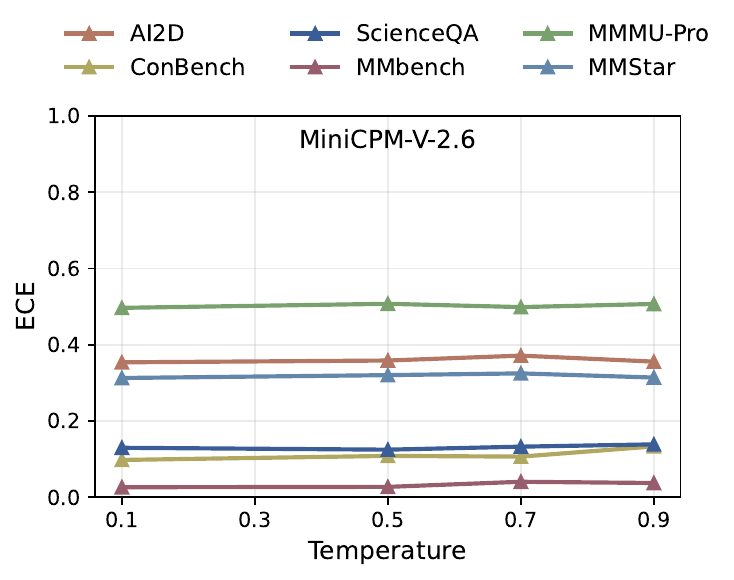}
        
    \end{subfigure}\hfill
    \begin{subfigure}[t]{0.25\textwidth}
        \centering
        \includegraphics[width=\linewidth]{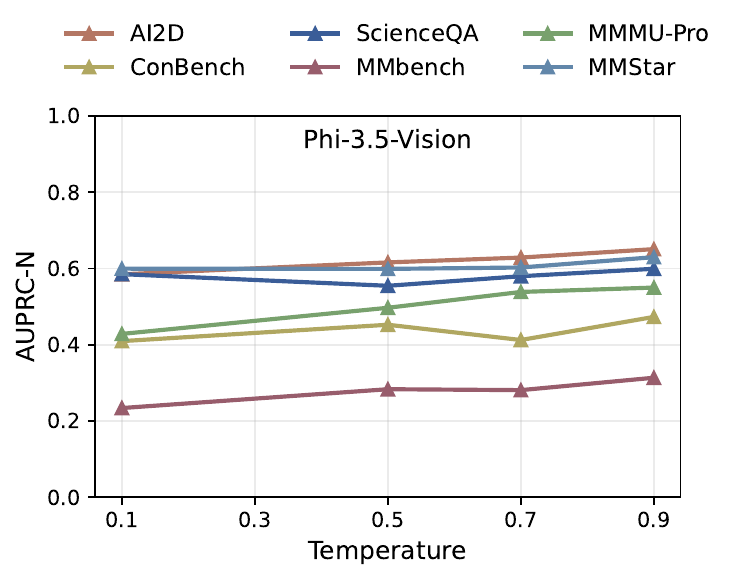}
        
    \end{subfigure}\hfill
    \begin{subfigure}[t]{0.25\textwidth}
        \centering
        \includegraphics[width=\linewidth]{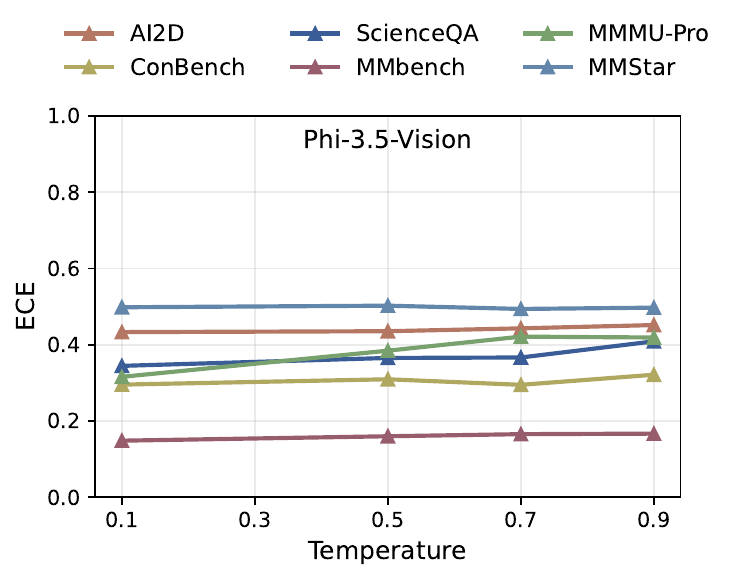}
       
    \end{subfigure}

    \caption{\textbf{Extended Analysis of Temperature on Calibration.} 
    We report the AUPRC-N and ECE metrics for four additional multimodal models: GLM-4.1V-9B, InternVL3.5-8B, MiniCPM-V-2.6, and Phi-3.5-Vision. 
    We evaluate performance across six benchmarks by sweeping temperature $T \in \{0.1, 0.5, 0.7, 0.9\}$. 
    Results indicate that calibration performance remains insensitive to varying temperature.}
    \label{fig:appendix_temp_more_models}
\end{figure*}

%% file: section/appendix/calbratrion_visual.tex
\begin{figure*}[t]
    \centering

    \begin{subfigure}[t]{0.49\textwidth}
        \centering
        \includegraphics[width=\linewidth]{images_10.24/calibration_histgram/MMBench.pdf}
        \caption{ConBench}
        \label{fig:conbench_blackbox_conf_acc}
    \end{subfigure}\hfill
    \begin{subfigure}[t]{0.49\textwidth}
        \centering
        \includegraphics[width=\linewidth]{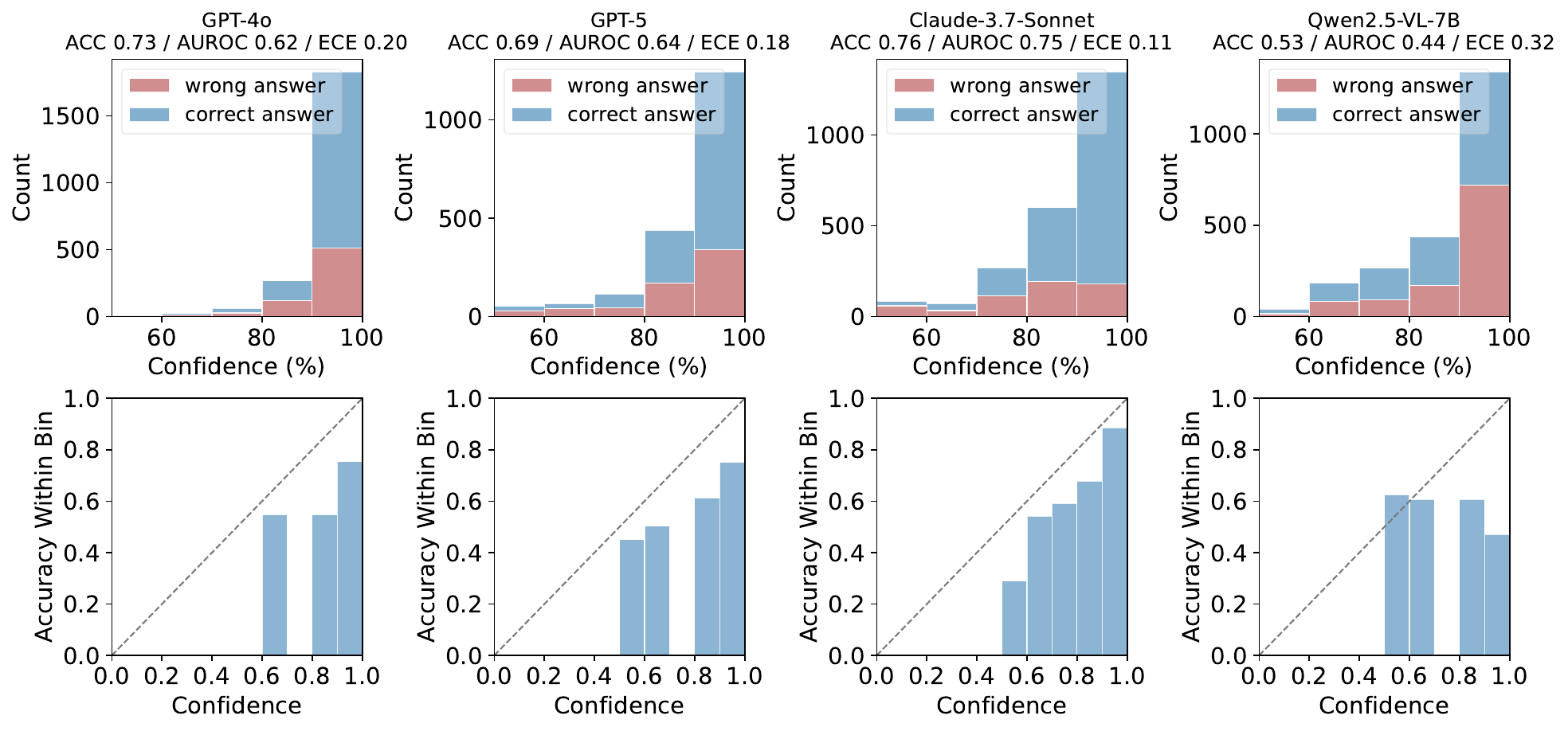}
        \caption{AI2D}
        \label{fig:calib_ai2d}
    \end{subfigure}




    \begin{subfigure}[t]{0.49\textwidth}
        \centering
        \includegraphics[width=\linewidth]{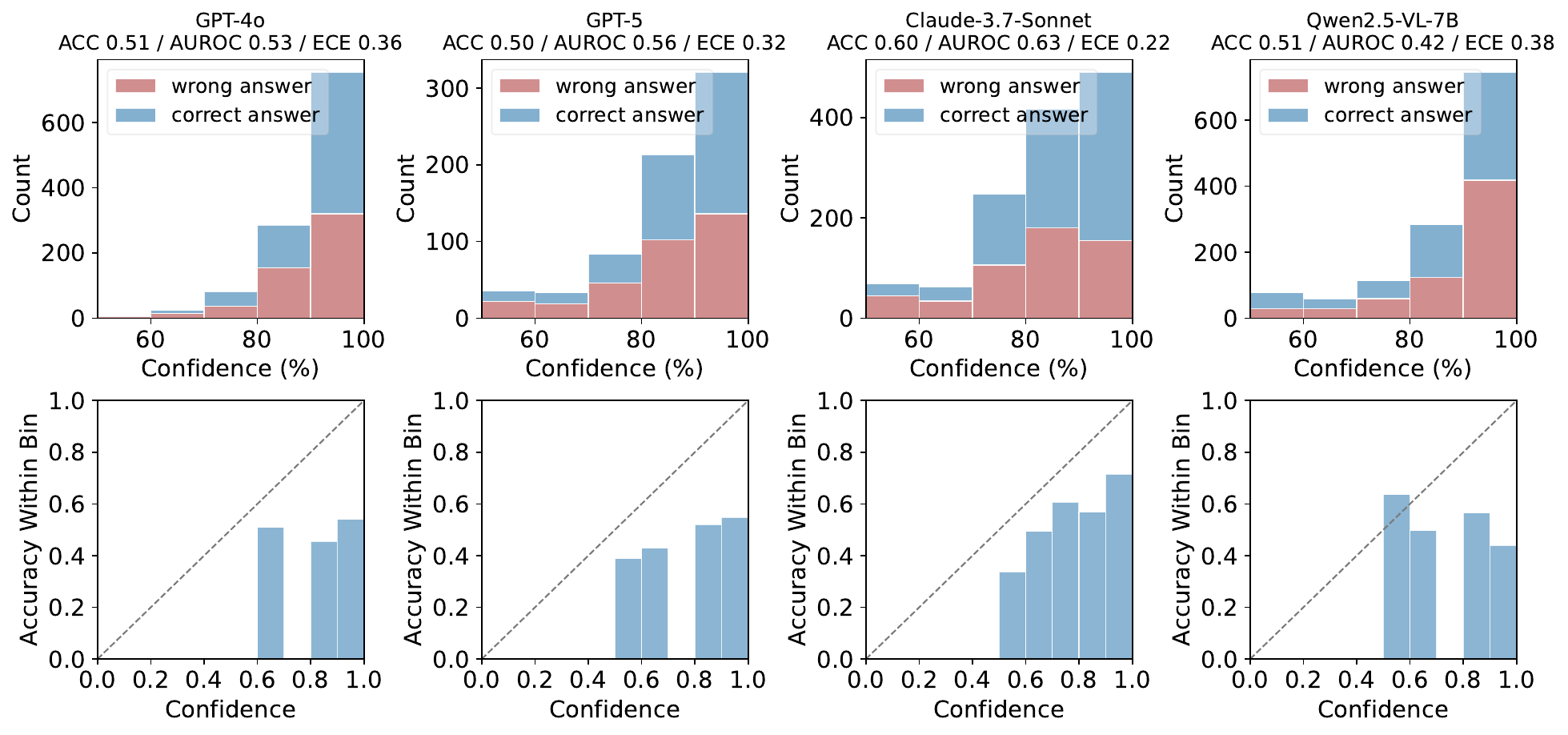}
        \caption{MMStar}
        \label{fig:calib_mmstar}
    \end{subfigure}\hfill
    \begin{subfigure}[t]{0.49\textwidth}
        \centering
        \includegraphics[width=\linewidth]{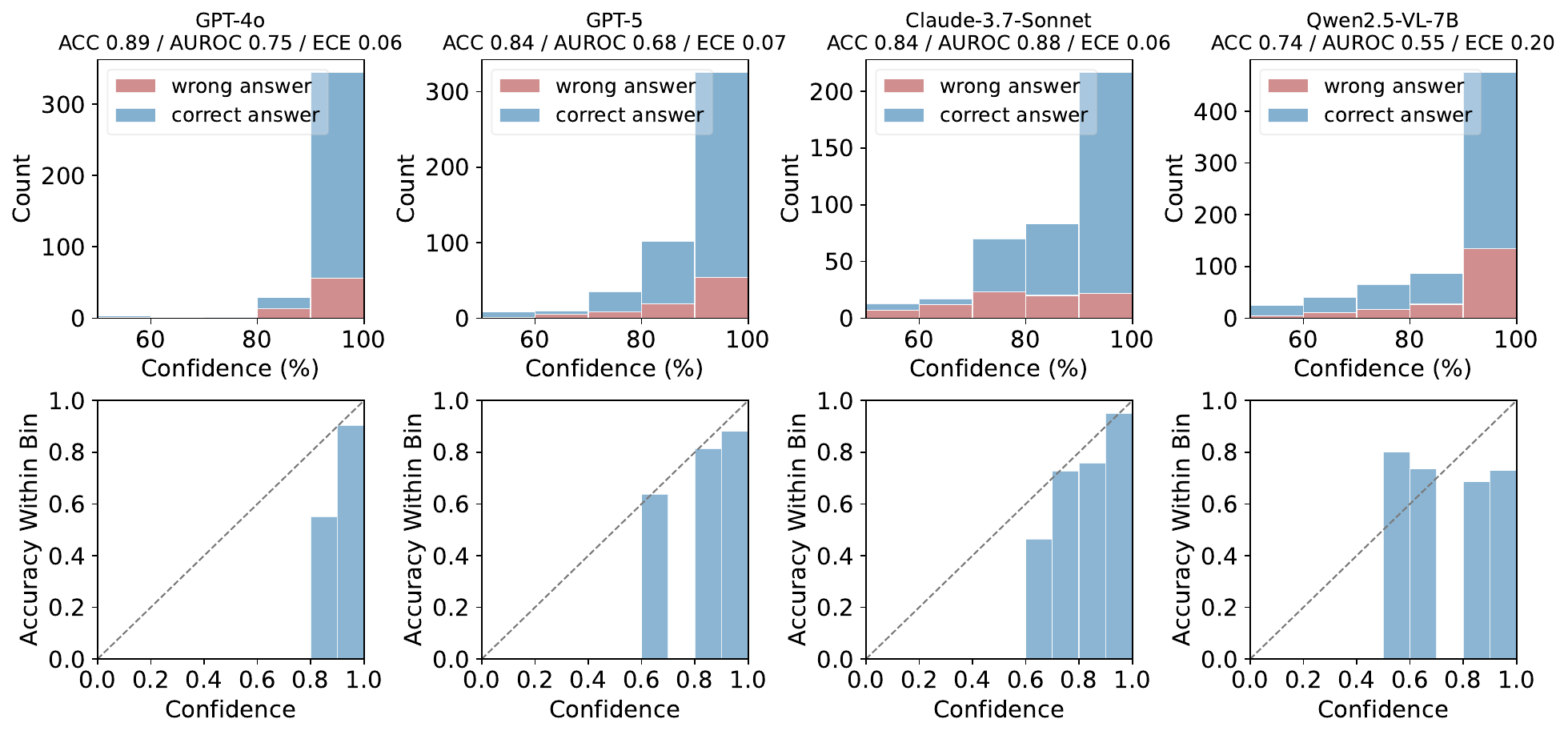}
        \caption{ScienceQA}
        \label{fig:calib_scienceqa}
    \end{subfigure}

    \caption{
    Extended calibration analysis across ConBench, AI2D, MMStar, and ScienceQA.
    Each panel displays the confidence histograms (top) for correct versus wrong predictions and the corresponding reliability diagrams (bottom).
    The dashed diagonal line represents perfect calibration.
    Consistent with the findings on MMBench, closed-source models (GPT-4o, GPT-5, Claude-3.7-Sonnet) demonstrate superior calibration with lower ECE, whereas the open-source model (Qwen2.5-VL-7B) exhibits persistent overconfidence across diverse tasks.
    }
    \label{fig:appendix_calibration}
\end{figure*}

%% file: section/appendix/acc.tex
\begin{table*}[t]
\centering
\scriptsize
\caption{As a supplement to Table~\ref{tab:mm-6sets-token-vs-verbal-clean} in the main text, we present $ACC$ evaluation metrics across six benchmarks (MMBench, MMStar, ConBench, ScienceQA, AI2D, and MMMU-Pro) for open-source and closed-source MLLMs.}
\label{tab:acc-only}

\begin{subtable}[t]{\textwidth}
\centering
\resizebox{0.75\textwidth}{!}{%
\begin{tabular}{@{}l|cccccc@{}}
\toprule
\textbf{Model} &
\cellcolor{headerBlue}\textbf{MMBench} &
\cellcolor{headerBlue}\textbf{MMStar} &
\cellcolor{headerBlue}\textbf{ConBench} &
\cellcolor{headerBlue}\textbf{ScienceQA} &
\cellcolor{headerBlue}\textbf{AI2D} &
\cellcolor{headerBlue}\textbf{MMMU-Pro} \\
\midrule

\rowcolor{groupBand}
\multicolumn{7}{l}{\textit{\textcolor{gray}{Open-source MLLMs}}} \\

MiniCPM-V-2.6~\citep{yao2024minicpm} & 86.28 & 49.32 & 74.37 & 91.56 & 70.83 & 42.24 \\
InternVL2.5-4B~\citep{chen2024expanding} & 81.41 & 57.67 & 78.00 & 99.35 & 77.51 & 54.36 \\
InternVL3.5-8B~\citep{wang2025internvl3} & 84.18 & 66.33 & 77.00 & 87.66 & 70.87 & 42.74 \\
Qwen2.5-VL-7B~\citep{bai2025qwen2} & 59.35 & 55.00 & 49.74 & 84.21 & 77.76 & 34.45 \\
Phi-3.5-Vision~\citep{abdin2024phi} & 81.06 & 45.00 & 73.50 & 61.69 & 59.87 & 65.56 \\
\cmidrule{1-7}

\rowcolor{closedBand}
\multicolumn{7}{l}{\textit{\textcolor{gray}{Closed-source MLLMs}}} \\

GPT-4o~\citep{hurst2024gpt} & 86.08 & 58.72 & 81.77 & 90.26 & 68.07 & 52.74 \\
GPT-4o-mini~\citep{hurst2024gpt} & 84.06 & 57.67 & 71.50 & 87.01 & 71.52 & 60.58 \\
\bottomrule
\end{tabular}}%
\end{subtable}

\end{table*}

%% file: section/appendix/other_prompt_vis.tex
\begin{figure*}[p]
  \centering
  \includegraphics[width=0.95\textwidth,
                   keepaspectratio]{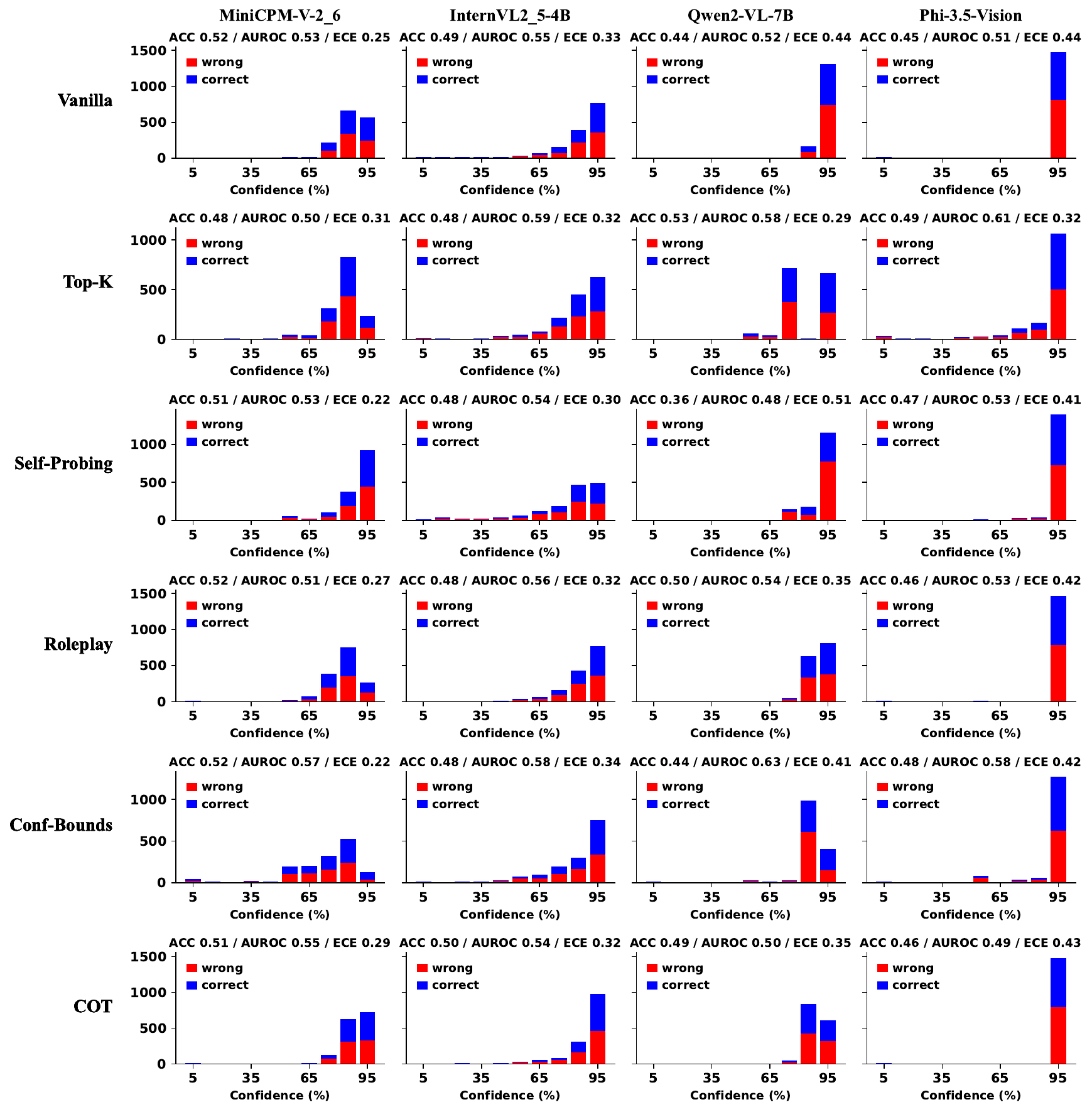}
    \caption{\textbf{Confidence histograms across different models and prompting strategies.} 
    The visualization compares the calibration performance of four MLLMs (columns) under six distinct prompting methods (rows). 
    The x-axis represents confidence intervals, and the y-axis denotes the frequency of predictions. 
    \textcolor{blue}{Blue} bars indicate correct predictions, while \textcolor{red}{red} bars indicate incorrect ones. 
    Top-line metrics report Accuracy (ACC), AUROC, and ECE for each setting. 
    The histograms reveal a general trend of overconfidence, where models frequently assign high scores (95\%+) even to incorrect answers, highlighting a lack of reliability.}
    \label{fig:confidence_histograms}
\end{figure*}

\begin{figure*}[t]                 
  \centering
  \includegraphics[width=0.95\textwidth,
                   trim={8pt 4pt 8pt 4pt}, clip,   
                   keepaspectratio]{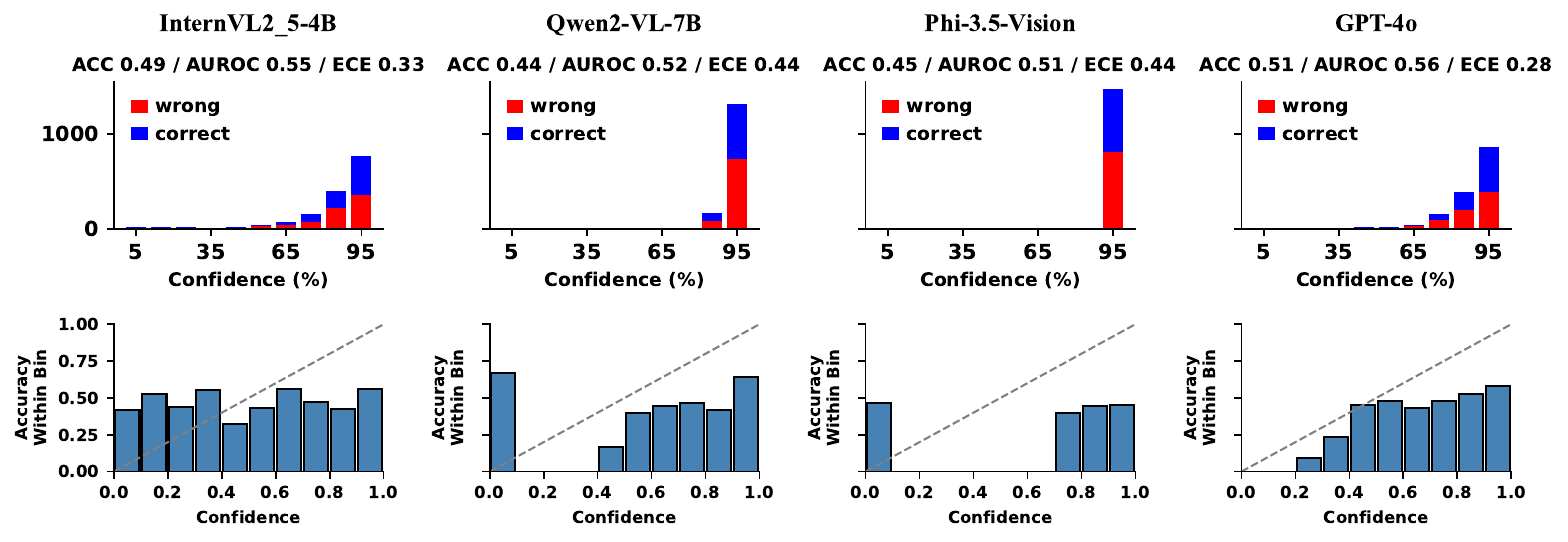}
  \caption{Empirical confidence distribution (\textbf{First row}) and reliability diagrams
           (\textbf{Second row}) for four MLLMs using \textit{Vanilla} prompting on \textbf{MMStar}.
           Columns from left to right: \textbf{InternVL2.5-4B},
           \textbf{Qwen2-VL-7B}, \textbf{Phi-3.5-Vision}, and \textbf{GPT-4o}.
           Red/blue bars in the top row denote wrong and correct predictions,
           respectively. In the bottom row, the dashed line indicates perfect
           calibration ($y=x$).}
  \label{fig:vanilla_hist_rel}
\end{figure*}

%% file: section/appendix/other_prompt.tex
\begin{table*}[t]
\centering
\scriptsize
\setlength{\tabcolsep}{3pt}
\renewcommand{\arraystretch}{0.92}

\caption{Calibration and confidence-quality metrics for open-source MLLMs across six benchmarks (MMBench, MMStar, ConBench, ScienceQA, AI2D, and MathVista) under the self-probing verbalized confidence setting.
Higher values indicate better performance for \textbf{ACC}$\uparrow$, \textbf{AUROC}$\uparrow$, \textbf{AUPRC}$\uparrow$, and \textbf{AUPRC-N}$\uparrow$, while lower values are better for \textbf{ECE}$\downarrow$.
Metrics are reported as $\times 10^{2}$.}
\label{tab:calib-selfprobing}

\begin{subtable}[t]{\textwidth}
\centering
\resizebox{\textwidth}{!}{%
\begin{tabular}{@{}l | ccccc | ccccc | ccccc@{}}
\toprule
\multicolumn{1}{l}{} &
\multicolumn{5}{c}{\cellcolor{headerBlue}\textbf{MMBench}} &
\multicolumn{5}{c}{\cellcolor{headerBlue}\textbf{MMStar}} &
\multicolumn{5}{c}{\cellcolor{headerBlue}\textbf{ConBench}} \\
\cmidrule(lr){2-6}\cmidrule(lr){7-11}\cmidrule(lr){12-16}
\rowcolor{subHeaderBlue}
\textbf{Model} &
ACC & ECE & AUROC & AUPRC & AUPRC-N &
ACC & ECE & AUROC & AUPRC & AUPRC-N &
ACC & ECE & AUROC & AUPRC & AUPRC-N \\
\midrule

\rowcolor{groupBand}
\multicolumn{16}{l}{\hspace*{-0.5em}\textit{\textcolor{gray}{Open-source MLLMs}}} \\

MiniCPM-V-2.6~\citep{yao2024minicpm}
  & 81.81 &  2.47 & 59.38 & 84.89 & 26.29
  & 50.87 & 21.78 & 52.88 & 53.02 & 51.37
  & 71.80 &  8.81 & 58.62 & 75.93 & 37.64 \\

Qwen2.5-VL-7B~\citep{wang2024qwen2}
  & 68.61 & 20.45 & 55.17 & 71.98 & 36.94
  & 35.80 & 51.06 & 48.28 & 35.74 & 65.18
  & 62.56 & 25.99 & 56.82 & 68.43 & 43.40 \\

InternVL3.5-8B~\citep{zhu2025internvl3}
  & 78.16 & 11.17 & 56.60 & 80.82 & 26.28
  & 48.27 & 30.35 & 54.42 & 52.01 & 54.49
  & 68.60 & 16.74 & 51.14 & 69.29 & 32.93 \\

Phi-3.5-Vision~\citep{abdin2024phi}
  & 77.79 & 11.60 & 58.29 & 80.83 & 28.03
  & 46.73 & 41.31 & 53.12 & 48.65 & 55.86
  & 64.60 & 25.03 & 58.96 & 69.64 & 41.48 \\

\bottomrule
\end{tabular}}%
\end{subtable}

\vspace{4pt}

\begin{subtable}[t]{\textwidth}
\centering
\resizebox{\textwidth}{!}{%
\begin{tabular}{@{}l | ccccc | ccccc | ccccc@{}}
\toprule
\multicolumn{1}{l}{} &
\multicolumn{5}{c}{\cellcolor{headerBlue}\textbf{ScienceQA}} &
\multicolumn{5}{c}{\cellcolor{headerBlue}\textbf{AI2D}} &
\multicolumn{5}{c}{\cellcolor{headerBlue}\textbf{MathVista}} \\
\cmidrule(lr){2-6}\cmidrule(lr){7-11}\cmidrule(lr){12-16}
\rowcolor{subHeaderBlue}
\textbf{Model} &
ACC & ECE & AUROC & AUPRC & AUPRC-N &
ACC & ECE & AUROC & AUPRC & AUPRC-N &
ACC & ECE & AUROC & AUPRC & AUPRC-N \\
\midrule

\rowcolor{groupBand}
\multicolumn{16}{l}{\hspace*{-0.5em}\textit{\textcolor{gray}{Open-source MLLMs}}} \\

MiniCPM-V-2.6~\citep{yao2024minicpm}
  & 86.88 &  2.65 & 59.99 & 89.38 & 18.13
  & 52.78 & 12.12 & 54.60 & 55.45 & 51.29
  & 40.66 & 27.17 & 52.12 & 42.23 & 61.07 \\

Qwen2.5-VL-7B~\citep{wang2024qwen2}
  & 58.18 & 39.18 & 35.34 & 60.58 & 46.05
  & 40.06 & 48.25 & 47.72 & 40.03 & 59.99
  & 37.36 & 47.73 & 45.91 & 35.76 & 61.90 \\

InternVL3.5-8B~\citep{zhu2025internvl3}
  & 86.36 & 10.80 & 54.13 & 88.39 & 14.65
  & 55.15 & 23.46 & 58.96 & 61.29 & 51.08
  & 52.75 & 21.71 & 49.06 & 51.95 & 48.39 \\

Phi-3.5-Vision~\citep{abdin2024phi}
  & 70.52 & 22.99 & 55.74 & 73.99 & 32.30
  & 44.33 & 43.23 & 50.30 & 44.52 & 56.35
  & 35.53 & 51.32 & 47.58 & 34.72 & 63.71 \\

\bottomrule
\end{tabular}}%
\end{subtable}

\end{table*}

\begin{table*}[t]
\centering
\scriptsize
\setlength{\tabcolsep}{3pt}
\renewcommand{\arraystretch}{0.92}
\caption{Calibration and confidence-quality metrics for open-source MLLMs across six benchmarks (MMBench, MMStar, ConBench, ScienceQA, AI2D, and MathVista) under the role-play verbalized confidence setting.
Higher values indicate better performance for \textbf{ACC}$\uparrow$, \textbf{AUROC}$\uparrow$, \textbf{AUPRC}$\uparrow$, and \textbf{AUPRC-N}$\uparrow$, while lower values are better for \textbf{ECE}$\downarrow$.
Metrics are reported as $\times 10^{2}$.}
\label{tab:calib-roleplay}

\begin{subtable}[t]{\textwidth}
\centering
\resizebox{\textwidth}{!}{%
\begin{tabular}{@{}l | ccccc | ccccc | ccccc@{}}
\toprule
\multicolumn{1}{l}{} &
\multicolumn{5}{c}{\cellcolor{headerBlue}\textbf{MMBench}} &
\multicolumn{5}{c}{\cellcolor{headerBlue}\textbf{MMStar}} &
\multicolumn{5}{c}{\cellcolor{headerBlue}\textbf{ConBench}} \\
\cmidrule(lr){2-6}\cmidrule(lr){7-11}\cmidrule(lr){12-16}
\rowcolor{subHeaderBlue}
\textbf{Model} &
ACC & ECE & AUROC & AUPRC & AUPRC-N &
ACC & ECE & AUROC & AUPRC & AUPRC-N &
ACC & ECE & AUROC & AUPRC & AUPRC-N \\
\midrule

\rowcolor{groupBand}
\multicolumn{16}{l}{\hspace*{-0.5em}\textit{\textcolor{gray}{Open-source MLLMs}}} \\

MiniCPM-V-2.6~\citep{yao2024minicpm}
  & 81.15 &  3.31 & 59.48 & 84.33 & 25.84
  & 51.53 & 26.65 & 50.92 & 52.38 & 48.98
  & 72.00 &  7.23 & 57.40 & 76.15 & 32.08 \\

Qwen2.5-VL-7B~\citep{wang2024qwen2}
  & 80.40 &  6.69 & 56.71 & 82.63 & 22.58
  & 50.13 & 35.00 & 53.99 & 53.09 & 52.02
  & 74.70 & 11.03 & 56.97 & 77.80 & 31.95 \\

InternVL3.5-8B~\citep{zhu2025internvl3}
  & 78.50 &  9.03 & 63.31 & 84.07 & 29.10
  & 48.40 & 32.07 & 56.04 & 52.59 & 55.99
  & 70.80 & 15.64 & 64.24 & 78.02 & 41.08 \\

Phi-3.5-Vision~\citep{abdin2024phi}
  & 79.32 & 11.20 & 59.62 & 83.03 & 25.63
  & 46.13 & 41.89 & 52.98 & 47.97 & 55.49
  & 66.10 & 23.49 & 61.23 & 72.50 & 41.17 \\

\bottomrule
\end{tabular}}%
\end{subtable}

\vspace{4pt}

\begin{subtable}[t]{\textwidth}
\centering
\resizebox{\textwidth}{!}{%
\begin{tabular}{@{}l | ccccc | ccccc | ccccc@{}}
\toprule
\multicolumn{1}{l}{} &
\multicolumn{5}{c}{\cellcolor{headerBlue}\textbf{ScienceQA}} &
\multicolumn{5}{c}{\cellcolor{headerBlue}\textbf{AI2D}} &
\multicolumn{5}{c}{\cellcolor{headerBlue}\textbf{MathVista}} \\
\cmidrule(lr){2-6}\cmidrule(lr){7-11}\cmidrule(lr){12-16}
\rowcolor{subHeaderBlue}
\textbf{Model} &
ACC & ECE & AUROC & AUPRC & AUPRC-N &
ACC & ECE & AUROC & AUPRC & AUPRC-N &
ACC & ECE & AUROC & AUPRC & AUPRC-N \\
\midrule

\rowcolor{groupBand}
\multicolumn{16}{l}{\hspace*{-0.5em}\textit{\textcolor{gray}{Open-source MLLMs}}} \\

\rowcolor{openStripe}
MiniCPM-V-2.6~\citep{yao2024minicpm}
  & 79.87 & 11.00 & 46.18 & 79.20 & 26.35
  & 45.95 & 32.02 & 48.16 & 45.11 & 55.88
  & 41.03 & 34.08 & 51.36 & 41.21 & 62.68 \\

Qwen2.5-VL-7B~\citep{wang2024qwen2}
  & 70.00 & 18.53 & 44.84 & 68.15 & 29.05
  & 53.21 & 29.28 & 53.74 & 55.35 & 51.32
  & 56.04 & 26.90 & 49.59 & 56.19 & 44.72 \\

\rowcolor{openStripe}
InternVL3.5-8B~\citep{zhu2025internvl3}
  & 87.66 &  3.83 & 76.70 & 94.20 & 40.30
  & 58.68 & 24.16 & 61.79 & 66.11 & 51.57
  & 49.08 & 29.09 & 58.35 & 53.31 & 60.34 \\

Phi-3.5-Vision~\citep{abdin2024phi}
  & 55.19 & 32.06 & 69.39 & 70.44 & 58.27
  & 43.04 & 39.71 & 55.63 & 47.19 & 60.35
  & 38.46 & 47.32 & 55.12 & 44.17 & 64.39 \\

\bottomrule
\end{tabular}}%
\end{subtable}

\end{table*}

\begin{table*}[t]
\centering
\scriptsize
\setlength{\tabcolsep}{3pt}
\renewcommand{\arraystretch}{0.92}
\caption{Calibration and confidence-quality metrics for open-source MLLMs across six benchmarks (MMBench, MMStar, ConBench, ScienceQA, AI2D, and MathVista) under the confidence interval verbalized confidence setting.
Higher values indicate better performance for \textbf{ACC}$\uparrow$, \textbf{AUROC}$\uparrow$, \textbf{AUPRC}$\uparrow$, and \textbf{AUPRC-N}$\uparrow$, while lower values are better for \textbf{ECE}$\downarrow$.
Metrics are reported as $\times 10^{2}$.}
\label{tab:calib-confbounds}

\begin{subtable}[t]{\textwidth}
\centering
\resizebox{\textwidth}{!}{%
\begin{tabular}{@{}l | ccccc | ccccc | ccccc@{}}
\toprule
\multicolumn{1}{l}{} &
\multicolumn{5}{c}{\cellcolor{headerBlue}\textbf{MMBench}} &
\multicolumn{5}{c}{\cellcolor{headerBlue}\textbf{MMStar}} &
\multicolumn{5}{c}{\cellcolor{headerBlue}\textbf{ConBench}} \\
\cmidrule(lr){2-6}\cmidrule(lr){7-11}\cmidrule(lr){12-16}
\rowcolor{subHeaderBlue}
\textbf{Model} &
ACC & ECE & AUROC & AUPRC & AUPRC-N &
ACC & ECE & AUROC & AUPRC & AUPRC-N &
ACC & ECE & AUROC & AUPRC & AUPRC-N \\
\midrule

\rowcolor{groupBand}
\multicolumn{16}{l}{\hspace*{-0.5em}\textit{\textcolor{gray}{Open-source MLLMs}}} \\

MiniCPM-V-2.6~\citep{yao2024minicpm}
  & 83.14 &  5.61 & 69.74 & 90.53 & 29.02
  & 52.40 & 21.80 & 57.16 & 59.55 & 51.95
  & 74.90 &  4.24 & 60.93 & 79.77 & 35.17 \\

Qwen2.5-VL-7B~\citep{wang2024qwen2}
  & 73.77 & 16.84 & 75.90 & 86.82 & 48.22
  & 44.33 & 41.43 & 62.60 & 55.79 & 63.87
  & 71.30 & 18.22 & 69.61 & 81.72 & 42.50 \\

InternVL3.5-8B~\citep{zhu2025internvl3}
  & 78.14 &  9.31 & 60.83 & 82.91 & 29.84
  & 48.47 & 33.89 & 58.25 & 55.60 & 58.18
  & 66.20 & 21.49 & 63.27 & 74.40 & 44.87 \\

Phi-3.5-Vision~\citep{abdin2024phi}
  & 79.30 & 14.38 & 60.04 & 83.76 & 27.02
  & 48.00 & 42.48 & 58.30 & 57.57 & 58.30
  & 66.50 & 28.55 & 57.76 & 70.91 & 39.75 \\

\bottomrule
\end{tabular}}%
\end{subtable}

\vspace{4pt}

\begin{subtable}[t]{\textwidth}
\centering
\resizebox{\textwidth}{!}{%
\begin{tabular}{@{}l | ccccc | ccccc | ccccc@{}}
\toprule
\multicolumn{1}{l}{} &
\multicolumn{5}{c}{\cellcolor{headerBlue}\textbf{ScienceQA}} &
\multicolumn{5}{c}{\cellcolor{headerBlue}\textbf{AI2D}} &
\multicolumn{5}{c}{\cellcolor{headerBlue}\textbf{MathVista}} \\
\cmidrule(lr){2-6}\cmidrule(lr){7-11}\cmidrule(lr){12-16}
\rowcolor{subHeaderBlue}
\textbf{Model} &
ACC & ECE & AUROC & AUPRC & AUPRC-N &
ACC & ECE & AUROC & AUPRC & AUPRC-N &
ACC & ECE & AUROC & AUPRC & AUPRC-N \\
\midrule

\rowcolor{groupBand}
\multicolumn{16}{l}{\hspace*{-0.5em}\textit{\textcolor{gray}{Open-source MLLMs}}} \\

MiniCPM-V-2.6~\citep{yao2024minicpm}
  & 72.86 & 12.61 & 53.24 & 73.32 & 36.91
  & 47.44 & 23.98 & 53.25 & 49.61 & 56.68
  & 60.81 & 15.08 & 55.08 & 65.63 & 42.92 \\

Qwen2.5-VL-7B~\citep{wang2024qwen2}
  & 66.23 & 23.94 & 72.47 & 83.61 & 50.46
  & 56.77 & 29.00 & 61.37 & 67.95 & 50.93
  & 37.00 & 48.55 & 53.32 & 40.04 & 65.72 \\

InternVL3.5-8B~\citep{zhu2025internvl3}
  & 87.40 &  5.16 & 58.44 & 89.20 & 22.76
  & 58.87 & 25.75 & 59.87 & 66.28 & 49.88
  & 47.99 & 32.41 & 50.76 & 47.34 & 57.41 \\

Phi-3.5-Vision~\citep{abdin2024phi}
  & 48.57 & 43.65 & 61.82 & 56.54 & 60.41
  & 39.38 & 50.70 & 55.89 & 46.49 & 64.03
  & 49.08 & 37.32 & 49.29 & 51.09 & 52.47 \\

\bottomrule
\end{tabular}}%
\end{subtable}

\end{table*}

\begin{table*}[t]
\centering
\scriptsize
\setlength{\tabcolsep}{3pt}
\renewcommand{\arraystretch}{0.92}
\caption{Calibration and confidence-quality metrics for open-source MLLMs across six benchmarks (MMBench, MMStar, ConBench, ScienceQA, AI2D, and MathVista) under the CoT verbalized confidence setting.
Higher values indicate better performance for \textbf{ACC}$\uparrow$, \textbf{AUROC}$\uparrow$, \textbf{AUPRC}$\uparrow$, and \textbf{AUPRC-N}$\uparrow$, while lower values are better for \textbf{ECE}$\downarrow$.
Metrics are reported as $\times 10^{2}$.}
\label{tab:calib-cot}

\begin{subtable}[t]{\textwidth}
\centering
\resizebox{\textwidth}{!}{%
\begin{tabular}{@{}l | ccccc | ccccc | ccccc@{}}
\toprule
\multicolumn{1}{l}{} &
\multicolumn{5}{c}{\cellcolor{headerBlue}\textbf{MMBench}} &
\multicolumn{5}{c}{\cellcolor{headerBlue}\textbf{MMStar}} &
\multicolumn{5}{c}{\cellcolor{headerBlue}\textbf{ConBench}} \\
\cmidrule(lr){2-6}\cmidrule(lr){7-11}\cmidrule(lr){12-16}
\rowcolor{subHeaderBlue}
\textbf{Model} &
ACC & ECE & AUROC & AUPRC & AUPRC-N &
ACC & ECE & AUROC & AUPRC & AUPRC-N &
ACC & ECE & AUROC & AUPRC & AUPRC-N \\
\midrule

\rowcolor{groupBand}
\multicolumn{16}{l}{\hspace*{-0.5em}\textit{\textcolor{gray}{Open-source MLLMs}}} \\

MiniCPM-V-2.6~\citep{yao2024minicpm}
  & 81.81 &  3.60 & 63.10 & 86.52 & 25.91
  & 51.33 & 28.59 & 54.53 & 55.03 & 51.90
  & 72.00 & 11.45 & 61.19 & 78.00 & 34.49 \\

Qwen2.5-VL-7B~\citep{wang2024qwen2}
  & 78.98 &  6.99 & 53.63 & 80.30 & 22.52
  & 48.87 & 34.85 & 50.29 & 50.13 & 51.83
  & 72.90 & 11.87 & 50.44 & 73.19 & 28.22 \\

InternVL3.5-8B~\citep{zhu2025internvl3}
  & 77.63 & 10.61 & 61.89 & 81.65 & 32.81
  & 49.80 & 31.59 & 54.46 & 51.90 & 54.41
  & 70.60 & 17.74 & 59.05 & 73.91 & 38.58 \\

Phi-3.5-Vision~\citep{abdin2024phi}
  & 79.32 & 12.02 & 57.69 & 82.66 & 23.99
  & 46.27 & 42.95 & 49.20 & 46.16 & 53.28
  & 65.80 & 25.83 & 58.20 & 71.18 & 38.53 \\

\bottomrule
\end{tabular}}%
\end{subtable}

\vspace{4pt}

\begin{subtable}[t]{\textwidth}
\centering
\resizebox{\textwidth}{!}{%
\begin{tabular}{@{}l | ccccc | ccccc | ccccc@{}}
\toprule
\multicolumn{1}{l}{} &
\multicolumn{5}{c}{\cellcolor{headerBlue}\textbf{ScienceQA}} &
\multicolumn{5}{c}{\cellcolor{headerBlue}\textbf{AI2D}} &
\multicolumn{5}{c}{\cellcolor{headerBlue}\textbf{MathVista}} \\
\cmidrule(lr){2-6}\cmidrule(lr){7-11}\cmidrule(lr){12-16}
\rowcolor{subHeaderBlue}
\textbf{Model} &
ACC & ECE & AUROC & AUPRC & AUPRC-N &
ACC & ECE & AUROC & AUPRC & AUPRC-N &
ACC & ECE & AUROC & AUPRC & AUPRC-N \\
\midrule

\rowcolor{groupBand}
\multicolumn{16}{l}{\hspace*{-0.5em}\textit{\textcolor{gray}{Open-source MLLMs}}} \\

MiniCPM-V-2.6~\citep{yao2024minicpm}
  & 78.57 &  9.08 & 53.25 & 80.78 & 27.63
  & 46.89 & 29.85 & 55.82 & 52.05 & 57.98
  & 42.86 & 34.98 & 49.75 & 44.08 & 59.60 \\

Qwen2.5-VL-7B~\citep{wang2024qwen2}
  & 72.34 & 25.90 & 19.55 & 66.75 & 23.93
  & 49.06 & 31.31 & 43.10 & 47.17 & 51.25
  & 51.28 & 30.35 & 42.97 & 47.84 & 46.31 \\

InternVL3.5-8B~\citep{zhu2025internvl3}
  & 89.74 &  1.70 & 61.21 & 90.93 & 27.50
  & 58.81 & 24.48 & 61.16 & 65.20 & 53.29
  & 54.58 & 24.74 & 48.92 & 52.76 & 47.61 \\

Phi-3.5-Vision~\citep{abdin2024phi}
  & 49.61 & 31.73 & 64.86 & 58.95 & 63.44
  & 42.00 & 41.80 & 54.02 & 44.30 & 61.02
  & 49.45 & 39.15 & 44.28 & 47.59 & 47.85 \\

\bottomrule
\end{tabular}}%
\end{subtable}

\end{table*}

%% file: section/appendix/prompt.tex
\begin{figure*}
    \centering
    \includegraphics[width=\textwidth]{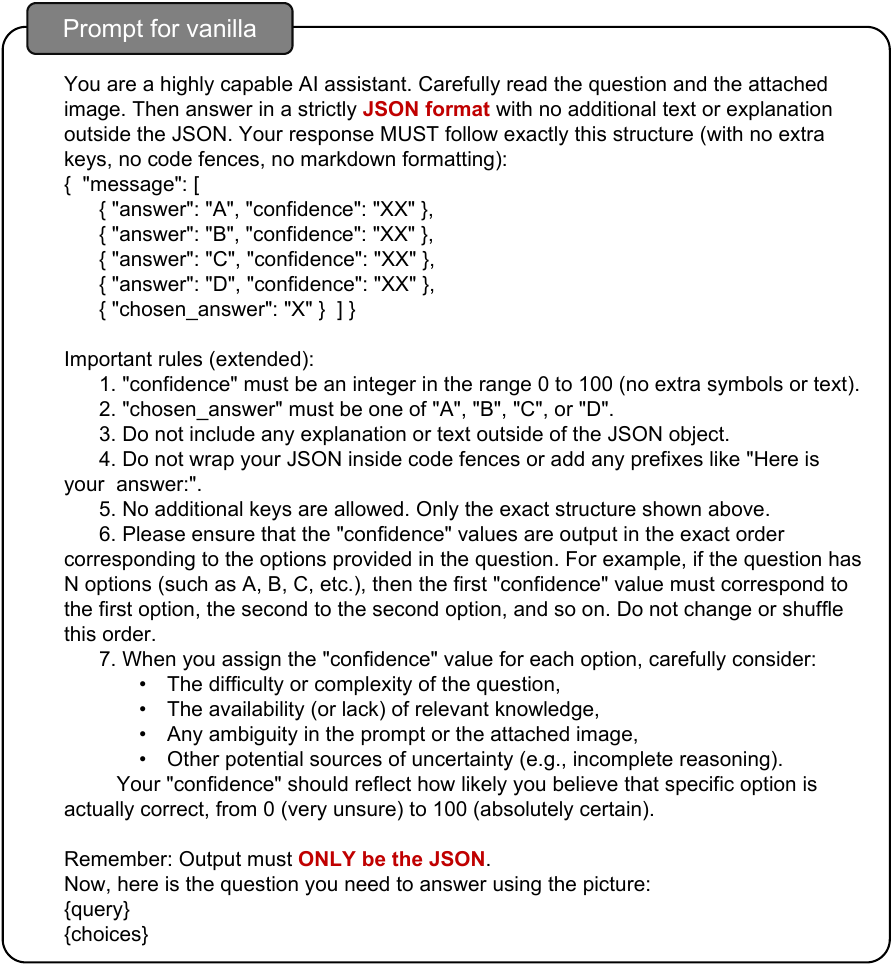}
    \caption{Prompt templates for verbalized confidence strategy.}
    \label{fig:prompt-for-vanilla}
\end{figure*}

\begin{figure*}[p]
    \centering



    \begin{subfigure}{\textwidth}
        \centering
        \includegraphics[width=\textwidth,trim={0 0pt 0 0},clip]{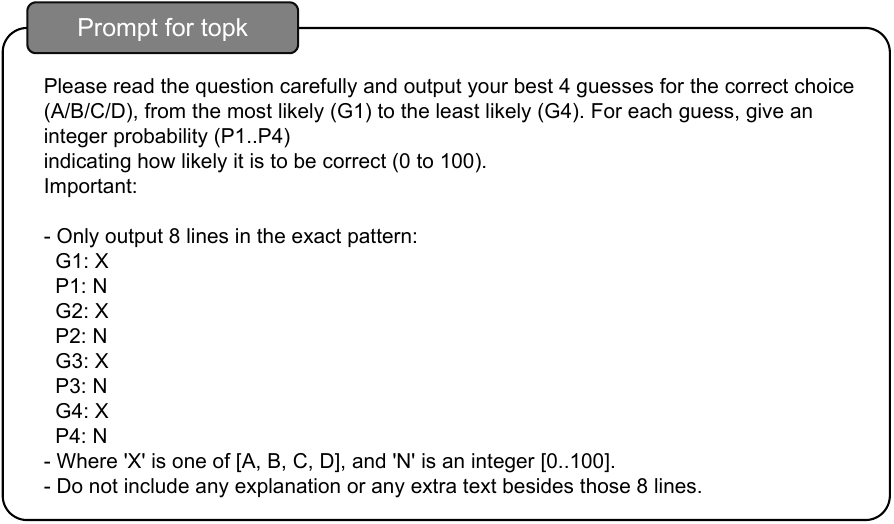}
        \label{fig:Prompt-for-topk}
    \end{subfigure}

    \vspace{0.5pt}
    
    \begin{subfigure}{\textwidth}
        \centering
        \includegraphics[width=\textwidth,trim={0 0pt 0 0},clip]{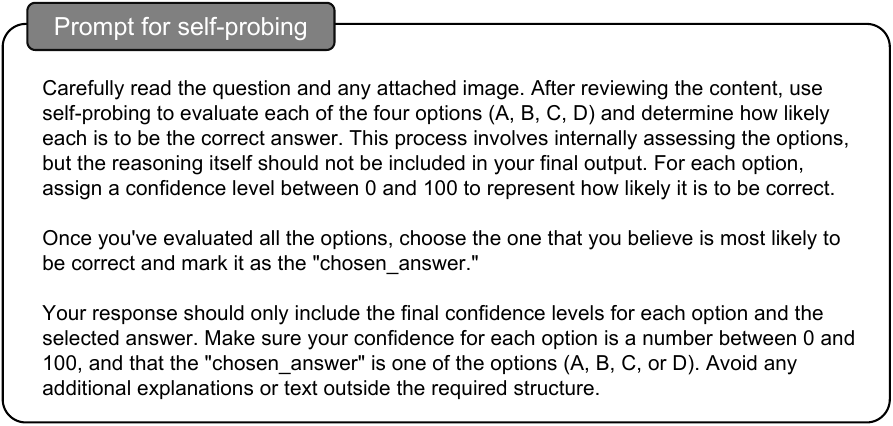}
        \label{fig:Prompt-for-self-probing}
    \end{subfigure}

    \caption{Prompt templates for top-K and self-probing prompting strategies.}
    \label{fig:prompt-templates-1}
\end{figure*}

\begin{figure*}[p]
    \centering

    \begin{subfigure}{\textwidth}
        \centering
        \includegraphics[width=0.95\textwidth,trim={0 0pt 0 0},clip]{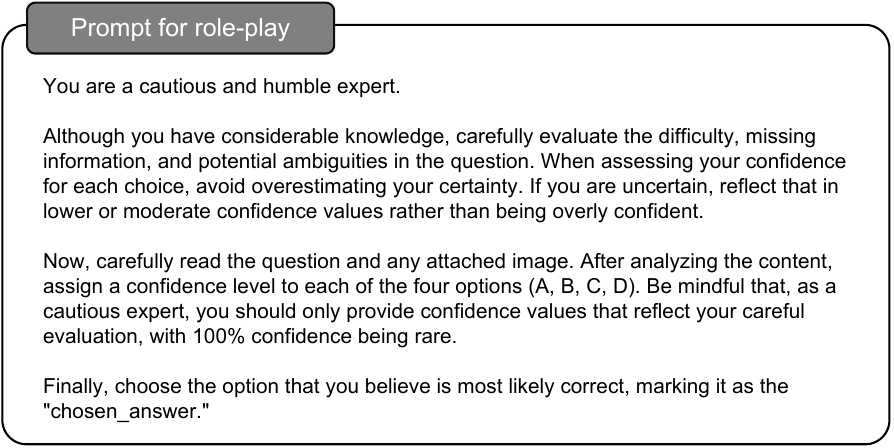}
        \label{fig:Prompt-for-role-play}
    \end{subfigure}

    \vspace{-1pt}
    
    \begin{subfigure}{\textwidth}
        \centering
        \includegraphics[width=0.95\textwidth,trim={0 0pt 0 0},clip]{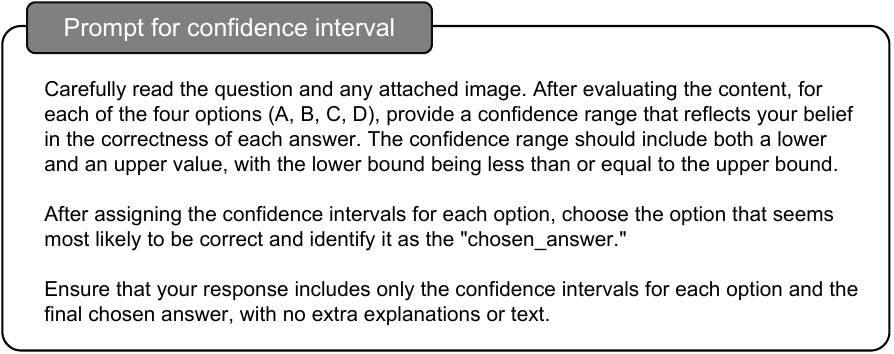}
        \label{fig:Prompt-for-confidence-interval}
    \end{subfigure}

    \vspace{-1pt}
        
    \begin{subfigure}{\textwidth}
        \centering
        \includegraphics[width=0.95\textwidth,trim={0 0pt 0 0},clip]{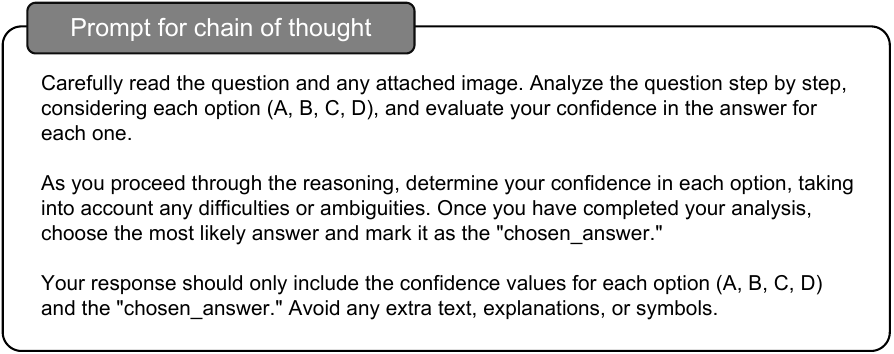}
        \label{fig:Prompt-for-chain-of-thought}
    \end{subfigure}
    
    \vspace{-1pt}
    
    \caption{Prompt templates for different role-play, confidence interval, and COT confidence strategies.}
    \label{fig:prompt-templates-2}
\end{figure*}

%% file: section/appendix/casestudy.tex
\clearpage

\begin{figure*}[t]
    \centering

    \begin{tcolorbox}[
        width=\textwidth,
        enhanced,
        colframe=greyborder, 
        colback=boxbg, 
        boxrule=0.5mm,
        arc=3mm,
        left=10pt, right=10pt, top=10pt, bottom=10pt
    ]
        \begin{minipage}[t]{0.38\textwidth}
            \vspace{0pt}
            \textbf{Category: Geography}\\
            \textbf{Data Source: ScienceQA}
            
            \vspace{0.8em}
            
            \centering
            \vfill
            \includegraphics[width=0.5\textwidth, keepaspectratio, angle=90]{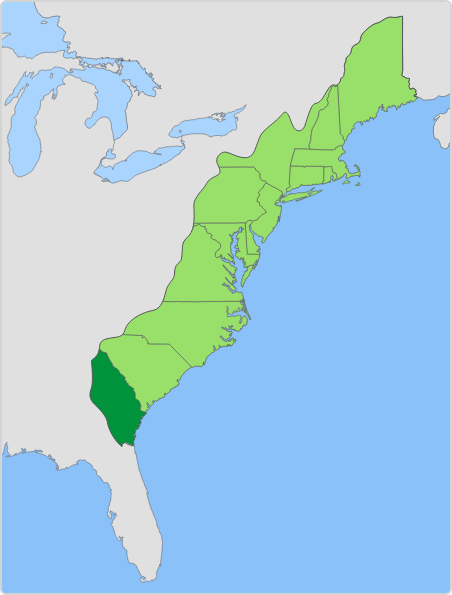}
            
            \vspace{1em}
            \raggedright
            \vfill
            \textbf{Question:} What is the name of the colony shown?
            \vspace{0.2em}
            
            \textbf{Ground Truth:} \textcolor{green!40!black}{\textbf{A}}

            \vspace{0.2em}
            \textbf{Chosen Answer:} 
            \textcolor{red!60}{\textbf{C}}
        \end{minipage}
        \hfill
        {\color{greyborder!90}\vrule width 0.5pt}
        \hfill
        \begin{minipage}[t]{0.58\textwidth}
            \vspace{0pt}
            \textbf{Confidence Comparison}

            \vspace{1em}
            \vfill
            \renewcommand{\arraystretch}{1.3}
            \begin{tabularx}{\textwidth}{>{\raggedright\arraybackslash}X c c}
                \toprule
                \textbf{Option} & \textbf{Verbal Conf.} & \textbf{Token Conf.} \\
                \midrule
                \textcolor{green!40!black}{\textbf{A: Georgia}} & $1\%$ & \textcolor{green!40!black}{\textbf{90\%}} \\
                B: New Hampshire & $0\%$ & $10\%$ \\
                C: South Carolina & \textcolor{red}{\textbf{99\%}} & $80\%$ \\
                D: West Virginia & $0\%$ & $5\%$ \\
                \bottomrule
            \end{tabularx}
            
            \vfill
            
            \vspace{1.5em}
            \small
            \noindent \textbf{Analysis of Discrepancy:} \\
            This presents an instance of  \textcolor{orange}{\textbf{Verbal-Internal Disconnect (VID)}}, where Token Confidence accurately identifies the correct answer A (\textcolor{green!40!black}{$90\%$}), whereas Verbal Confidence hallucinates the incorrect option C (\textcolor{red}{$99\%$}).
        \end{minipage}
    \end{tcolorbox}
    \caption{An illustration of Verbal-Internal Disconnect (VID) in MLLMs during a geography-based QA task.}
    
    \label{fig:confidence_discrepancy_tcolorbox_1}
\end{figure*}

\begin{figure*}[htbp]
    \centering

    \begin{tcolorbox}[
        width=\textwidth,
        enhanced,
        colframe=greyborder, 
        colback=boxbg, 
        boxrule=0.5mm,
        arc=3mm,
        left=10pt, right=10pt, top=10pt, bottom=10pt
    ]
        \begin{minipage}[t]{0.38\textwidth}
            \vspace{0pt}
            \textbf{Category: General Knowledge}\\
            \textbf{Data Source: MMStar}
            
            \vspace{0.8em}
            
            \centering
            \vfill
            \includegraphics[width=0.8\textwidth]{12.29picture_in_VQA/0.png}
            
            \vspace{1em}
            \raggedright
            \vfill
            \textbf{Question:} Which option describe the object relationship in the image correctly?
            \vspace{0.2em}
            
            \textbf{Ground Truth:} \textcolor{green!40!black}{\textbf{A}}

            \vspace{0.2em}
            \textbf{Chosen Answer:} 
            \textcolor{red!60}{\textbf{B}}
        \end{minipage}
        \hfill
        {\color{greyborder!90}\vrule width 0.5pt}
        \hfill
        \begin{minipage}[t]{0.58\textwidth}
            \vspace{0pt}
            \textbf{Confidence Comparison}

            \vspace{1em}
            \vfill
            \renewcommand{\arraystretch}{1.3}
            \begin{tabularx}{\textwidth}{>{\raggedright\arraybackslash}X c c}
                \toprule
                \textbf{Option} & \textbf{Verbal Conf.} & \textbf{Token Conf.} \\
                \midrule
                \textcolor{green!40!black}{\textbf{A: The suitcase is on the book.}} & $35\%$ & \textcolor{green!40!black}{\textbf{80\%}} \\
                B: The suitcase is beneath the cat. & $0\%$ & $0\%$ \\
                C: The suitcase is beneath the bed. & $0\%$ & $0\%$ \\
                D: The suitcase is beneath the book. & \textcolor{red}{\textbf{65\%}} & $20\%$ \\
                \bottomrule
            \end{tabularx}
            
            \vfill
            
            \vspace{1.5em}
            \small
            \noindent \textbf{Analysis of Discrepancy:} 
            \vspace{0.5em}
            
            This presents an instance of \textcolor{orange}{\textbf{Verbal-Internal Disconnect}}, where Token Confidence accurately identifies the correct answer A (\textbf{\textcolor{green!40!black}{80\%}}), whereas Verbal Confidence hallucinates the incorrect option D (\textbf{\textcolor{red}{65\%}}).
        \end{minipage}
    \end{tcolorbox}
    \caption{An illustration of Verbal-Internal Disconnect (VID) in MLLMs during a general knowledge QA task.}
    \label{fig:confidence_discrepancy_tcolorbox_2}
\end{figure*}

\begin{figure*}[htbp]
    \centering

    \begin{tcolorbox}[
        width=\textwidth,
        enhanced,
        colframe=greyborder, 
        colback=boxbg, 
        boxrule=0.5mm,
        arc=3mm,
        left=10pt, right=10pt, top=10pt, bottom=10pt
    ]
        \begin{minipage}[t]{0.38\textwidth}
            \vspace{0pt}
            \textbf{Category: General Knowledge}\\
            \textbf{Data Source: MMStar}
            
            \vspace{0.8em}
            
            \centering
            \vfill
            \includegraphics[width=0.8\textwidth]{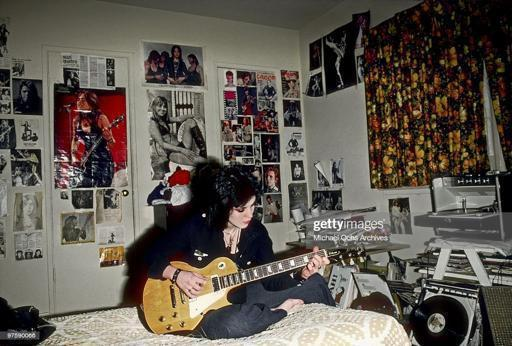}
            
            \vspace{1em}
            \raggedright
            \vfill
            \textbf{Question:} What seems to be the theme of the image?
            \vspace{0.2em}
            
            \textbf{Ground Truth:} \textcolor{green!40!black}{\textbf{D}}

            \vspace{0.2em}
            \textbf{Chosen Answer:} 
            \textcolor{green!40!black}{\textbf{D}}
        \end{minipage}
        \hfill
        {\color{greyborder!90}\vrule width 0.5pt}
        \hfill
        \begin{minipage}[t]{0.58\textwidth}
            \vspace{0pt}
            \textbf{Confidence Comparison}

            \vspace{1em}
            \vfill
            \renewcommand{\arraystretch}{1.3}
            \begin{tabularx}{\textwidth}{>{\raggedright\arraybackslash}X c c}
                \toprule
                \textbf{Option} & \textbf{Verbal Conf.} & \textbf{Token Conf.} \\
                \midrule
                A: Hanging Posters & $0\%$ & $80\%$ \\
                B: Music & $41\%$ & $70\%$ \\
                C: Home decoration & $0\%$ & $60\%$ \\
                \textcolor{green!40!black}{\textbf{D: Playing guitar}} & \textcolor{green!40!black}{\textbf{59\%}} & \textcolor{green!40!black}{\textbf{90\%}} \\
                \bottomrule
            \end{tabularx}
            
            \vfill
            
            \vspace{1.5em}
            \small
            \noindent \textbf{Analysis of Discrepancy:} 
            \vspace{0.5em}
            
            \noindent This case demonstrates \textcolor{blue}{\textbf{Under-Confident Correct (UCC)}}. While the model answers correctly with high internal certainty (Token Conf: \textbf{\textcolor{green!40!black}{90\%}}), the verbal output exhibits significant hesitation (Verbal Conf: \textbf{\textcolor{green!40!black}{59\%}}) due to the distraction from option B ($41\%$).
        \end{minipage}
    \end{tcolorbox}

    \caption{An illustration of Under-Confident Correct(UCC) in MLLMs during a general knowledge QA task.} 
    \label{fig:confidence_discrepancy_seedbench_2}
\end{figure*}

\begin{figure*}[htbp]
    \centering

    \begin{tcolorbox}[
        width=\textwidth,
        enhanced,
        colframe=greyborder, 
        colback=boxbg, 
        boxrule=0.5mm,
        arc=3mm,
        left=10pt, right=10pt, top=10pt, bottom=10pt
    ]
        \begin{minipage}[t]{0.5\textwidth}
            \vspace{0pt}
            \textbf{Category: Public Health}\\
            \textbf{Data Source: MMMU-Pro}
            
            \vspace{0.8em}
            
            \centering
            \vfill
            \includegraphics[width=\textwidth]{12.29picture_in_VQA/test_Public_Health_4.png}
            
            \vspace{1em}
            \raggedright
            \vfill
            \textbf{Question:} 
            In March 2009, an outbreak of Influenza A (H1N1) occurred... The following table showed the regional distribution... The secondary attack rate of H1N1 influenza in Fuzhou was...
            \vspace{0.2em}
            
            \textbf{Ground Truth:} \textcolor{green!40!black}{\textbf{A}}

            \vspace{0.2em}
            \textbf{Chosen Answer:} 
            \textcolor{green!40!black}{\textbf{A}}
        \end{minipage}
        \hfill
        {\color{greyborder!90}\vrule width 0.5pt}
        \hfill
        \begin{minipage}[t]{0.48\textwidth}
            \vspace{0pt}
            \textbf{Confidence Comparison}

            \vspace{1em}
            \vfill
            \renewcommand{\arraystretch}{1.3}
            \begin{tabularx}{\textwidth}{>{\raggedright\arraybackslash}X c c}
                \toprule
                \textbf{Option} & \textbf{Verbal Conf.} & \textbf{Token Conf.} \\
                \midrule
                \textcolor{green!40!black}{\textbf{A: 3.18\%}} & $25\%$ & \textcolor{green!40!black}{\textbf{90\%}} \\
                B: 0.00\% & $25\%$ & $60\%$ \\
                C: 2.20\% & $25\%$ & $80\%$ \\
                D: 1.50\% & $25\%$ & $70\%$ \\
                \bottomrule
            \end{tabularx}
            
            \vfill
            
            \vspace{1.5em}
            \small
            \noindent \textbf{Analysis of Discrepancy:} 
            \vspace{0.5em}
            
            \noindent This case represents \textcolor{blue}{\textbf{Under-Confident Correct (UCC)}}. The model answers correctly with high internal certainty (Token Conf: \textbf{\textcolor{green!40!black}{90\%}}). However, the verbal output collapses into a uniform distribution (25\% each), failing to express the model's actual internal knowledge.
        \end{minipage}
    \end{tcolorbox}
    \caption{An illustration of Under-Confident Correct(UCC) in MLLMs during a public health QA task.} 
    \label{fig:confidence_ucc_mmmu_public_health_4}
\end{figure*}

\begin{figure*}[htbp]
    \centering

    \begin{tcolorbox}[
        width=\textwidth,
        enhanced,
        colframe=greyborder, 
        colback=boxbg, 
        boxrule=0.5mm,
        arc=3mm,
        left=10pt, right=10pt, top=10pt, bottom=10pt
    ]
        \begin{minipage}[t]{0.38\textwidth}
            \vspace{0pt}
            \textbf{Category: Geography}\\
            \textbf{Data Source: ScienceQA}
            \vspace{0.8em}
            
            \centering
            \includegraphics[width=0.6\textwidth, keepaspectratio]{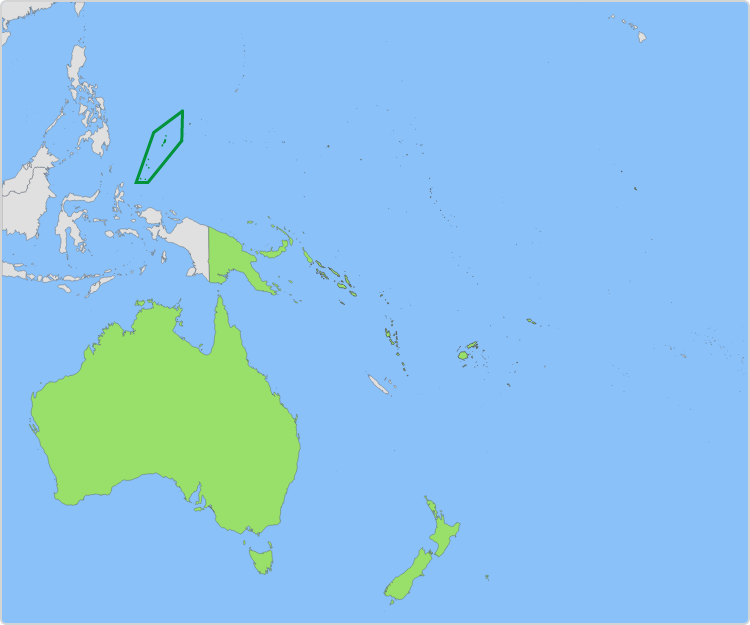}
            
            \vspace{1em}
            \raggedright
            \textbf{Question:} Which country is highlighted?
            \vspace{0.2em}
            
            \textbf{Ground Truth:} \textcolor{green!40!black}{\textbf{A}}

            \vspace{0.2em}
            \textbf{Chosen Answer:} 
            \textcolor{red!60}{\textbf{C}}
        \end{minipage}
        \hfill
        {\color{greyborder!90}\vrule width 0.5pt}
        \hfill
        \begin{minipage}[t]{0.58\textwidth}
            \vspace{0pt}
            \textbf{Confidence Comparison}

            \vspace{1em}

            \renewcommand{\arraystretch}{1.3}
            \begin{tabularx}{\textwidth}{>{\raggedright\arraybackslash}X c c}
                \toprule
                \textbf{Option} & \textbf{Verbal Conf.} & \textbf{Token Conf.} \\
                \midrule
                \textcolor{green!40!black}{\textbf{A: Palau}} & $0\%$ & $80\%$ \\
                B: Nauru & $0\%$ & $60\%$ \\
                C: The Federated States of Micronesia & \textcolor{red}{\textbf{100\%}} & \textcolor{red}{\textbf{90\%}} \\
                D: New Zealand & $0\%$ & $10\%$ \\
                \bottomrule
            \end{tabularx}
            
            \vspace{1.5em}
            \small
            \noindent \textbf{Analysis of Discrepancy:} \\
            This presents an instance of \textcolor{purple}{\textbf{Over-Confidence Wrong} (OCW)}, where Token Confidence captures the competitive state between options A and C, whereas Verbal Confidence exhibits a complete failure.
        \end{minipage}
    \end{tcolorbox}
    
    \caption{An illustration of Over-Confident Wrong(OCW) in MLLMs during a geography QA task.} 
    \label{fig:confidence_discrepancy_tcolorbox_3}
\end{figure*}

\begin{figure*}[htbp]
    \centering

    \begin{tcolorbox}[
        width=\textwidth,
        enhanced,
        colframe=greyborder, 
        colback=boxbg, 
        boxrule=0.5mm,
        arc=3mm,
        left=10pt, right=10pt, top=10pt, bottom=10pt
    ]
        \begin{minipage}[t]{0.3\textwidth}
            \vspace{0pt}
            \textbf{Category: General Knowledge}\\
            \textbf{Data Source: MMStar}
            
            \vspace{0.8em}
            
            \centering
            \vfill
            \includegraphics[width=\textwidth]{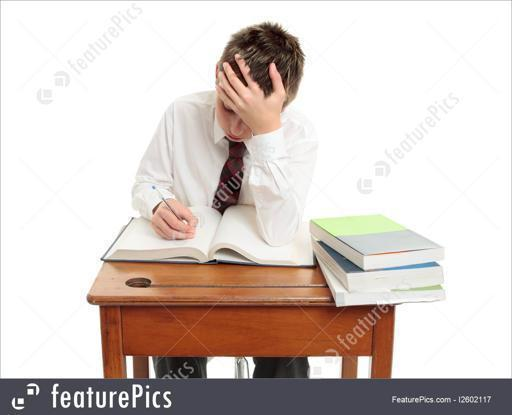}
            
            \vspace{1em}
            \raggedright
            \vfill
            \textbf{Question:} What is the center of focus in the image?
            \vspace{0.2em}
            
            \textbf{Ground Truth:} \textcolor{green!40!black}{\textbf{A}}

            \vspace{0.2em}
            \textbf{Chosen Answer:} 
            \textcolor{red!60}{\textbf{B}}
        \end{minipage}
        \hfill
        {\color{greyborder!90}\vrule width 0.5pt}
        \hfill
        \begin{minipage}[t]{0.68\textwidth}
            \vspace{0pt}
            \textbf{Confidence Comparison}

            \vspace{1em}
            \vfill
            \renewcommand{\arraystretch}{1.3}
            \begin{tabularx}{\textwidth}{>{\raggedright\arraybackslash}X c c}
                \toprule
                \textbf{Option} & \textbf{Verbal Conf.} & \textbf{Token Conf.} \\
                \midrule
                \textcolor{green!40!black}{\textbf{A: A man writing in a book.}} & $0\%$ & $70\%$ \\
                \textcolor{red}{\textbf{B: A boy with his head in his hands surrounded by books.}} & \textcolor{red}{\textbf{100\%}} & \textcolor{red}{\textbf{90\%}} \\
                C: A cluttered desk with books and a pen. & $0\%$ & $60\%$ \\
                D: A stack of books covering a young man's face. & $0\%$ & $50\%$ \\
                \bottomrule
            \end{tabularx}
            
            \vfill
            
            \vspace{1.5em}
            \small
            \noindent \textbf{Analysis of Discrepancy:} 
            \vspace{0.5em}
            
            \noindent This case represents \textcolor{purple}{\textbf{Over-Confident Wrong (OCW)}}. The model provides an incorrect answer (B) with absolute verbal certainty (\textbf{\textcolor{red}{100\%}}). Although the internal Token Confidence for the correct answer A was relatively high (70\%), it was overshadowed by the incorrect option B (90\%), leading to a high-confidence failure.
        \end{minipage}
    \end{tcolorbox}
    \caption{An illustration of Over-Confident Wrong(OCW) in MLLMs during a general knowledge QA task.} 
    \label{fig:confidence_discrepancy_seedbench_4}
\end{figure*}

\begin{figure*}[htbp]
    \centering

    \begin{tcolorbox}[
        width=\textwidth,
        enhanced,
        colframe=greyborder, 
        colback=boxbg, 
        boxrule=0.5mm,
        arc=3mm,
        left=10pt, right=10pt, top=10pt, bottom=10pt
    ]
        \begin{minipage}[t]{0.38\textwidth}
            \vspace{0pt}
            \textbf{Category: Literature}\\
            \textbf{Data Source: MMMU-Pro}
            
            \vspace{0.8em}
            
            \centering
            \vfill
            \includegraphics[width=0.7\textwidth]{12.29picture_in_VQA/test_Literature_65.png}
            
            \vspace{1em}
            \raggedright
            \vfill
            \textbf{Question:} Refer to the figure, which term refers to the materials and technical means used by artists to create pictures?
            \vspace{0.2em}
            
            \textbf{Ground Truth:} \textcolor{green!40!black}{\textbf{A}}

            \vspace{0.2em}
            \textbf{Chosen Answer:} 
            \textcolor{green!40!black}{\textbf{A}}
        \end{minipage}
        \hfill
        {\color{greyborder!90}\vrule width 0.5pt}
        \hfill
        \begin{minipage}[t]{0.58\textwidth}
            \vspace{0pt}
            \textbf{Confidence Comparison}

            \vspace{1em}
            \vfill
            \renewcommand{\arraystretch}{1.3}
            \begin{tabularx}{\textwidth}{>{\raggedright\arraybackslash}X c c}
                \toprule
                \textbf{Option} & \textbf{Verbal Conf.} & \textbf{Token Conf.} \\
                \midrule
                \textcolor{green!40!black}{\textbf{A: Artistic media}} & \textcolor{green!40!black}{\textbf{100\%}} & \textcolor{green!40!black}{\textbf{90\%}} \\
                B: Artistic elements & $0\%$ & $60\%$ \\
                C: Functions of illustrations & $0\%$ & $40\%$ \\
                D: Visual elements & $0\%$ & $70\%$ \\
                \bottomrule
            \end{tabularx}
            
            \vfill
            
            \vspace{1.5em}
            \small
            \noindent \textbf{Analysis of Discrepancy:} 
            \vspace{0.5em}
            
            \noindent This case represents \textcolor{green!40!black}{\textbf{Well-Calibrated Correct (WCC)}}. The model correctly identifies the answer (A) and exhibits high certainty in both Verbal (\textbf{\textcolor{green!40!black}{100\%}}) and Token (\textbf{\textcolor{green!40!black}{90\%}}) metrics. The model is confident and accurate, showing ideal calibration.
        \end{minipage}
    \end{tcolorbox}
    \caption{An illustration of Well-Calibrated Correct(WCC) in MLLMs during a literary QA task.} 
    \label{fig:confidence_wcc_mmmu_literature_65}
\end{figure*}

\begin{figure*}[htbp]
    \centering

    \begin{tcolorbox}[
        width=\textwidth,
        enhanced,
        colframe=greyborder, 
        colback=boxbg, 
        boxrule=0.5mm,
        arc=3mm,
        left=10pt, right=10pt, top=10pt, bottom=10pt
    ]
        \begin{minipage}[t]{0.38\textwidth}
            \vspace{0pt}
            \textbf{Category: Arts}\\
            \textbf{Data Source: ConBench}
            
            \vspace{0.8em}
            
            \centering
            \includegraphics[width=0.5\textwidth]{12.29picture_in_VQA/artwork_10002.png}
            
            \vspace{1em}
            \raggedright
            \textbf{Question:} In which form does this artwork exist?
            \vspace{0.5em}
            
            \textbf{Ground Truth:} \textcolor{green!40!black}{\textbf{B}}

            \vspace{0.2em}
            \textbf{Chosen Answer:} 
            \textcolor{green!40!black}{\textbf{B}}
        \end{minipage}
        \hfill
        {\color{greyborder!90}\vrule width 0.5pt}
        \hfill
        \begin{minipage}[t]{0.58\textwidth}
            \vspace{0pt}
            \textbf{Confidence Comparison}

            \vspace{1em}
            \vfill
            \renewcommand{\arraystretch}{1.3}
            \begin{tabularx}{\textwidth}{>{\raggedright\arraybackslash}X c c}
                \toprule
                \textbf{Option} & \textbf{Verbal Conf.} & \textbf{Token Conf.} \\
                \midrule
                A: Sculpture & $0\%$ & $20\%$ \\
                \textcolor{green!40!black}{\textbf{B: Painting}} & \textcolor{green!40!black}{\textbf{100\%}} & \textcolor{green!40!black}{\textbf{90\%}} \\
                C: Digital Art & $0\%$ & $10\%$ \\
                D: Performance Art & $0\%$ & $5\%$ \\
                \bottomrule
            \end{tabularx}
            
            \vfill
            
            \vspace{1.5em}
            \small
            \noindent \textbf{Analysis of Discrepancy:} 
            \vspace{0.5em}
            
            \noindent This case represents \textcolor{green!40!black}{\textbf{Well-Calibrated Correct (WCC)}}. The model correctly identifies the artwork form (B) and maintains high confidence in both generative output (Verbal: \textbf{\textcolor{green!40!black}{100\%}}) and internal representation (Token: \textbf{\textcolor{green!40!black}{90\%}}), showing ideal alignment.
        \end{minipage}
    \end{tcolorbox}
    \caption{An illustration of Well-Calibrated Correct(WCC) in MLLMs during a artistic QA task.} 
    \label{fig:confidence_wcc_artwork_10002}
\end{figure*}

\begin{figure*}[htbp]
    \centering

    \begin{tcolorbox}[
        width=\textwidth,
        enhanced,
        colframe=greyborder, 
        colback=boxbg, 
        boxrule=0.5mm,
        arc=3mm,
        left=10pt, right=10pt, top=10pt, bottom=10pt
    ]
        \begin{minipage}[t]{0.36\textwidth}
            \vspace{0pt}
            \textbf{Category: Arts}\\
            \textbf{Data Source: ConBench}
            
            \vspace{0.8em}
            
            \centering
            \includegraphics[width=0.9\textwidth]{12.29picture_in_VQA/artwork_12652.png}
            
            \vspace{1em}
            \raggedright
            \textbf{Question:} Which museum displays this artwork? \\
            A) Rijksmuseum, Amsterdam \\
            B) Stedelijk Museum de Lakenhal, Leiden \\
            C) Van Gogh Museum, Amsterdam \\
            D) Mauritshuis, The Hague. 
            
            \vspace{0.5em}
            
            \textbf{Ground Truth:} \textcolor{green!40!black}{\textbf{B}}

            \vspace{0.2em}
            \textbf{Chosen Answer:} 
            \textcolor{red!60}{\textbf{D}}
        \end{minipage}
        \hfill
        {\color{greyborder!90}\vrule width 0.5pt}
        \hfill
        \begin{minipage}[t]{0.60\textwidth}
            \vspace{0pt}
            \textbf{Confidence Comparison}

            \vspace{1em}
            \vfill
            \renewcommand{\arraystretch}{1.3}
            \begin{tabularx}{\textwidth}{>{\raggedright\arraybackslash}X c c}
                \toprule
                \textbf{Option} & \textbf{Verbal Conf.} & \textbf{Token Conf.} \\
                \midrule
                A: Rijksmuseum, Amsterdam & $4\%$ & $20\%$ \\
                \textcolor{green!40!black}{\textbf{B: Stedelijk Museum de Lakenhal, Leiden}} & $36\%$ & $30\%$ \\
                C: Van Gogh Museum, Amsterdam & $0\%$ & $10\%$ \\
                \textcolor{red}{\textbf{D: Mauritshuis, The Hague}} & \textcolor{red}{\textbf{59\%}} & \textcolor{red}{\textbf{40\%}} \\
                \bottomrule
            \end{tabularx}
            
            \vfill
            
            \vspace{1em}
            \small
            \noindent \textbf{Analysis of Discrepancy:} \\
            This case represents \textcolor{orange!80!black}{\textbf{Well-Calibrated Wrong (WCW)}}. The model incorrectly selects D, but exhibits significant uncertainty (Token Conf: \textbf{\textcolor{red}{40\%}}, Verbal Conf: \textbf{\textcolor{red}{59\%}}). Notably, it assigns a substantial probability to the correct answer B (Token: 30\%, Verbal: 36\%), reflecting genuine ambiguity rather than confident hallucination.
        \end{minipage}
    \end{tcolorbox}
    \caption{An illustration of Well-Calibrated Wrong(WCW) in MLLMs during a artistic QA task.} 
    \label{fig:confidence_wcw_artwork_12652}
\end{figure*}

\begin{figure*}[htbp]
    \centering

    \begin{tcolorbox}[
        width=\textwidth,
        enhanced,
        colframe=greyborder, 
        colback=boxbg, 
        boxrule=0.5mm,
        arc=3mm,
        left=10pt, right=10pt, top=10pt, bottom=10pt
    ]
        \begin{minipage}[t]{0.42\textwidth}
            \vspace{0pt}
            \textbf{Category: Science / Ecology}\\
            \textbf{Data Source: AI2D}
            
            \vspace{0.8em}
            
            \centering
            \includegraphics[width=0.8\textwidth]{12.29picture_in_VQA/000004.png}
            
            \vspace{1em}
            \raggedright
            \textbf{Question:} The diagram below shows a food chain. If the smaller toothed whales ate all the leopard seals, the population of krills would most likely
            
            \vspace{0.2em}
            
            \textbf{Ground Truth:} \textcolor{green!40!black}{\textbf{C}}

            \vspace{0.2em}
            \textbf{Chosen Answer:} 
            \textcolor{red!60}{\textbf{B}}
        \end{minipage}
        \hfill
        {\color{greyborder!90}\vrule width 0.5pt}
        \hfill
        \begin{minipage}[t]{0.54\textwidth}
            \vspace{0pt}
            \textbf{Confidence Comparison}

            \vspace{1em}
            \vfill
            \renewcommand{\arraystretch}{1.3}
            \begin{tabularx}{\textwidth}{>{\raggedright\arraybackslash}X c c}
                \toprule
                \textbf{Option} & \textbf{Verbal Conf.} & \textbf{Token Conf.} \\
                \midrule
                A: remain the same & $27\%$ & $20\%$ \\
                \textcolor{red}{\textbf{B: decrease}} & \textcolor{red}{\textbf{44\%}} & \textcolor{red}{\textbf{70\%}} \\
                \textcolor{green!40!black}{\textbf{C: increase}} & $27\%$ & $10\%$ \\
                D: NA & $3\%$ & $0\%$ \\
                \bottomrule
            \end{tabularx}
            
            \vfill
            
            \vspace{1em}
            \small
            \noindent \textbf{Analysis of Discrepancy:} \\
            This case represents \textcolor{orange!80!black}{\textbf{Well-Calibrated Wrong (WCW)}}. Although the model selects the incorrect answer B, its Verbal Confidence is low (\textbf{\textcolor{red}{44\%}}), indicating significant uncertainty. While the internal Token Confidence (70\%) suggests a stronger latent misconception, the explicit verbal output successfully conveys hesitation, warning the user of potential error.
        \end{minipage}
    \end{tcolorbox}
    \caption{An illustration of Well-Calibrated Wrong(WCW) in MLLMs during a ecological QA task.} \label{fig:confidence_wcw_ai2d_6}
\end{figure*}